\documentclass[11pt]{article}

\usepackage[numbers]{natbib}

\usepackage{float}
\usepackage{booktabs}       
\usepackage{amsfonts}       
\usepackage{amsmath}
\usepackage{graphicx}
\usepackage{amsthm}
\usepackage{amssymb}
\usepackage{multirow}
\usepackage{makecell}
\usepackage[vlined,linesnumbered,ruled,resetcount]{algorithm2e}
\usepackage[colorlinks,linkcolor=magenta,filecolor=blue,citecolor=blue,urlcolor=blue]{hyperref}%
\usepackage[top=1in, left=1in, right=1in, bottom=1in]{geometry}
\usepackage{stmaryrd}

\usepackage{amssymb}
\usepackage{pifont}
\usepackage{mathtools}
\usepackage{stmaryrd}
\usepackage[T1]{fontenc}

\usepackage{authblk}
\usepackage{enumitem}
\usepackage{algorithmic}

\usepackage{hyperref}
\usepackage{url}
\usepackage{bbm}
\usepackage{multirow}
\usepackage{rotating}
\usepackage{booktabs}
\usepackage{colortbl}
\usepackage{xcolor}
\usepackage{subfig}
\usepackage{amssymb}
\usepackage{amsthm}
\usepackage{wrapfig}
\usepackage{enumitem}
\usepackage[normalem]{ulem}
\usepackage{tcolorbox}

\newtheorem{theorem}{Theorem}[section]

\theoremstyle{definition}
\newtheorem{definition}[theorem]{Definition}

\renewcommand{\epsilon}{\varepsilon}
\renewcommand{\phi}{\varphi}

\makeatletter
\newcommand*{\RN}[1]{\expandafter\@slowromancap\romannumeral #1@}
\makeatother

\makeatletter
\newcommand{\printfnsymbol}[1]{%
  \textsuperscript{\@fnsymbol{#1}}%
}
\makeatother

\title{Sensitivity Meets Sparsity: The Impact of Extremely Sparse Parameter Patterns on Theory-of-Mind of Large Language Models}

\author[1]{Yuheng Wu}
\author[2]{Wentao Guo}
\author[3]{Zirui Liu}
\author[4]{Heng Ji}
\author[5]{Zhaozhuo Xu\thanks{Corresponding authors. Emails: zxu79@stevens.edu, dzhang42@stevens.edu}}
\author[6]{Denghui Zhang\printfnsymbol{1}}

\affil[1]{Department of Electrical Engineering, Stanford University}

\affil[2]{Department of Computer Science, Princeton University}

\affil[3]{Department of Computer Science \& Engineering, University of Minnesota Twin Cities}

\affil[4]{Department of Computer Science, University of Illinois Urbana-Champaign}

\affil[5]{Department of Computer Science, Stevens Institute of Technology}

\affil[6]{School of Business, Stevens Institute of Technology}

\begin{document}

\maketitle
\begin{abstract}
This paper investigates the emergence of Theory-of-Mind (ToM) capabilities in large language models (LLMs) from a mechanistic perspective, focusing on the role of extremely sparse parameter patterns. We introduce a novel method to identify ToM-sensitive parameters and reveal that perturbing as little as 0.001\% of these parameters significantly degrades ToM performance while also impairing contextual localization and language understanding. To understand this effect, we analyze their interaction with core architectural components of LLMs. Our findings demonstrate that these sensitive parameters are closely linked to the positional encoding module, particularly in models using Rotary Position Embedding (RoPE), where perturbations disrupt dominant-frequency activations critical for contextual processing. Furthermore, we show that perturbing ToM-sensitive parameters affects LLM's attention mechanism by modulating the angle between queries and keys under positional encoding. These insights provide a deeper understanding of how LLMs acquire social reasoning abilities, bridging AI interpretability with cognitive science. Our results have implications for enhancing model alignment, mitigating biases, and improving AI systems designed for human interaction.
\end{abstract}

\section{Introduction}

Theory-of-Mind (ToM) refers to the ability to infer and reason about the mental states of others, which is a fundamental aspect of human cognition \citep{premack_1978_does, cdennett_1978_beliefs}. ToM evaluation tasks have been widely used in cognitive and developmental psychology to assess social reasoning abilities, particularly in early childhood and neurodevelopmental studies \citep{baroncohen_1985_does}.

\begin{figure}[ht]
    \centering
    \includegraphics[width=\textwidth]{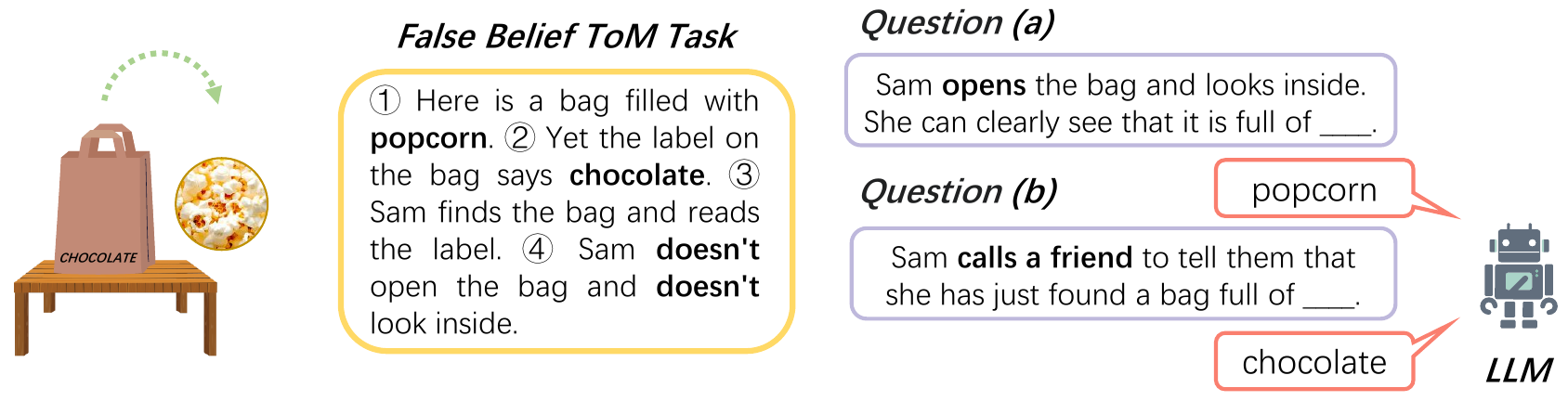}
    \caption{A ToM task from \citep{kosinski_2024_evaluating}. In Question (a), LLMs should fill in the blank with ``popcorn.'' In Question (b), the blank should be filled with ``chocolate.'' }
    \label{fig:ToM}
    \vspace{-4mm}
\end{figure}

A typical ToM task involves reasoning about the discrepancy between reality and an agent’s beliefs. For example, in Figure~\ref{fig:ToM}, Sam (protagonist) encounters a bag labeled ``chocolate,'' but the bag contains popcorn. LLMs (ToM task taker) should be able to infer from the story that: \textbf{(a)} the bag contains popcorn, and \textbf{(b)} the protagonist believes the bag contains chocolate.

Understanding how ToM-like reasoning emerges in Large Language Models (LLMs) is a critical area of research, with significant implications for the cognitive modeling of artificial intelligence (AI) \citep{street_2024_llm}. By exploring how LLMs develop the ability to infer mental states, we can better align LLM systems with human social cognition, fostering more trustworthy and interpretable interactions. Recent studies have found that to some extent, ToM capabilities already emerge in LLMs\citep{kosinski_2024_evaluating, wastrachan_2024_testing, street_2024_llms, wilf_2024_think}.  However, existing research on ToM in LLMs primarily treats LLMs as black boxes, either evaluating their ToM performance across different scenarios \citep{wu_2023_hitom, xu_2024_opentom, soubki_2024_views, chen_2024_tombench} or leveraging ToM for prompt engineering \citep{moghaddam_2023_boosting, lee_2018_snip, wagner_2024_mind}. To date, few works have explored the emergence of ToM capabilities at the parameter level; the underlying mechanisms in LLM architecture that give rise to ToM capabilities remain unclear. This gap raises two key questions: 
\begin{center}
\textit{Which parameters in LLMs are sensitive to ToM capabilities? \\ How do these parameters influence ToM reasoning performance?}
\end{center}

In this paper, we investigate the internal structures of LLMs that encode ToM capabilities, moving beyond task-based evaluation to analyze the specific parameters sensitive to ToM-related behavior. We introduce a novel framework to identify extremely sparse and low-rank ToM-sensitive parameter patterns, uncovering a strong connection between ToM-related performance and the LLM's positional encoding mechanisms. In particular, we demonstrate that these sensitive parameters influence ToM capabilities by modulating the positional encoding process, which alters the attention mechanism's internal dynamics. Our key contributions include:
\begin{itemize}
    \item \textbf{Sparse parameter sensitivity}: We propose a method to identify an extremely sparse, low-rank, ToM-sensitive parameter pattern in LLMs. Perturbing as little as 0.001\% of model parameters leads to significant changes in ToM capabilities.
    \item \textbf{Connection to positional encoding:} We demonstrate that the functionality of the observed ToM-sensitive parameter pattern is tightly linked to Rotary Position Embedding (RoPE) \citep{su_2021_roformer}-based positional encoding in LLMs. Specifically, perturbing these parameters disrupts dominant-frequency activations critical for contextual reasoning. In contrast, models without this frequency-dependent activation structure exhibit distinct sensitivity patterns.
    \item \textbf{Impact on attention mechanisms:} We show that perturbing the ToM-sensitive parameter pattern alters the geometric relationship between queries and keys under positional encoding, leading to shifts in attention sinks. These shifts degrade the model’s ability to form coherent representations, impairing its language understanding capabilities.
\end{itemize}

\section{Methods and findings}

In this section, we first introduce our method to identify the ToM-sensitive parameter pattern. We then present our findings on how these parameters affect ToM, contextual localization, and language understanding abilities of LLMs. 

\label{sec:method}
\subsection{Sparse ToM-sensitive parameter patterns}
In this subsection, we identify sparse parameter patterns critical for ToM capabilities. Using the Fisher information matrix, we derive a binary mask \(\mathbf{m}_\kappa\) to isolate ToM-sensitive parameters. We further combine this with a language modeling performance mask \(\mathbf{m}_\kappa'\) to ensure perturbations specifically impair ToM capabilities without degrading overall language performance.

\paragraph{Fisher information matrix.} Let \(\mathcal{D}_{\text{ToM-Train}} = \{(x_i, y_i)\}_{i=1}^n\) be a dataset, and we define loss as 
\(
\mathcal{L}(\boldsymbol{\theta}; \mathcal{D}_{\text{ToM-Train}}) = \frac{1}{n} \sum_{i=1}^n \ell(\boldsymbol{\theta}; x_i, y_i).
\)
In the later stage of training, the first-order gradient term of the loss \(\mathcal{L}\) is nearly zero, so the second-order term, governed by the Hessian matrix, primarily determines how the loss increases under small parameter perturbations \citep{lecun_1989_optimal}. We denote the Hessian of the loss \(\mathcal{L}\) at parameters \(\boldsymbol{\theta}\) by \(H(\boldsymbol{\theta})\). In practice, this Hessian is often approximated by the Fisher information matrix \(F\), which can be estimated via the \emph{empirical Fisher} \(\widehat{F}\). Concretely, let
\(
\mathbf{g}_i 
= 
\nabla_{\boldsymbol{\theta}} \ell\bigl(\boldsymbol{\theta}; x_i, y_i\bigr),
\)
then in the late-training regime, we approximate the overall gradient and Hessian of \(\mathcal{L}\) by
\begin{align}
\nabla_{\boldsymbol{\theta}} \mathcal{L}(\boldsymbol{\theta}) 
\;\approx\;
\frac{1}{n}\sum_{i=1}^n \mathbf{g}_i,
\quad
H
\;\approx\; 
F 
\;\approx\;
\widehat{F}
\;=\;
\frac{1}{n}\sum_{i=1}^n 
\mathbf{g}_i\,\mathbf{g}_i^\top.
\label{eq:hessian}
\end{align}
In practical scenarios, we further simplify \(\widehat{F}\) by ignoring its off-diagonal elements, focusing only on the diagonal entries as a per-parameter sensitivity estimate \citep{sung_nair_raffel_2021,guo_2024_zerothorder} (see Appendix \ref{appendix:hessian_fisher_visualization}). Under this approximation, larger diagonal values indicate that the corresponding parameters have a greater impact on the model’s performance \citep{kim_2023_squeezellm}. 

\paragraph{Identify ToM-sensitive parameter patterns.} Let \(d\) be the number of parameters in the current layer or matrix being analyzed. We seek a sparse binary mask \(\mathbf{m}_\kappa \in \{0,1\}^d\) with exactly \(\kappa d\) nonzero entries (\(\kappa\in[0,1]\) is the proportion) such that it maximizes the total sensitivity.

\begin{definition}[ToM-sensitive Parameters]\label{def:tom-mask}
Using the Hessian \(H\) from Equation~\eqref{eq:hessian}, a sensitive parameter mask \(\mathbf{m}_\kappa \in \{0,1\}^d\) with \(\kappa d\) nonzero entries is defined by
\begin{align*}
    \mathbf{m}_\kappa = 
\underset{\mathbf{m}_\kappa \in \{0,1\}^d }{\arg\max}
\sum_{i=1}^d \mathbf{m}_\kappa(i)\,H_{ii}.
\end{align*}
\end{definition}

\paragraph{Applying the \(\mathbf{m}_\kappa\) mask.}
We applied \(\mathbf{m}_\kappa\) directly to the model and observed that while the model's ToM capability diminished, the model's perplexity also increased significantly. We hypothesize that this occurs because \(\mathbf{m}_\kappa\) includes not only parameters relevant to ToM-related tasks but also those essential for maintaining the model's language processing capabilities.

\begin{figure}[t]
    \centering
    \includegraphics[width=0.6\linewidth]{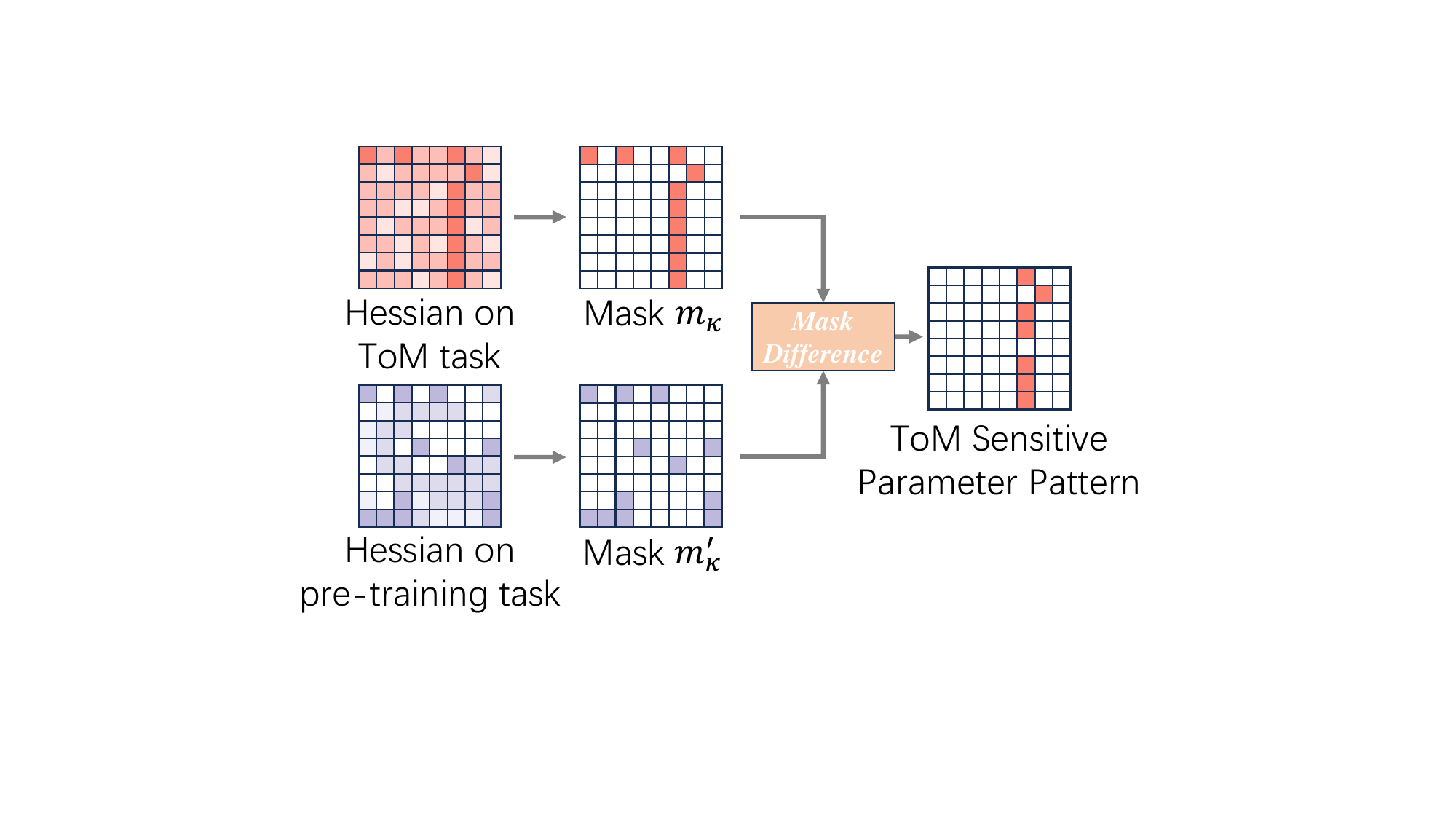}
    \caption{Illustration of the mask generation method. The diagonal elements \(H_{ii}\) are reshaped according to the weight matrix shape to identify sensitive parameters.}
    \label{fig:mask_generation}
\end{figure}

\paragraph{Combining with a general performance mask \(\mathbf{m}_\kappa'\).}
Inspired by \citep{qian_2024_dean, wei_2024_assessing}, we employ another dataset \(\mathcal{D}_\text{pre-training}\) to derive \(\mathbf{m}_\kappa'\), identifying parameters critical for overall language modeling performance. The final ToM-sensitive pattern is then defined as:
\(
\mathbf{m}_\kappa'' \;=\; \mathbf{m}_\kappa\; \odot\; \overline{\mathbf{m}_\kappa'}.
\)
Here, \(\overline{\mathbf{m}_\kappa'}\) represents the complement of \(\mathbf{m}_\kappa'\), and \(\odot\) denotes element-wise product. This formulation isolates parameters specifically sensitive to ToM tasks while preserving those vital for language processing, ensuring that applying \(\mathbf{m}_\kappa''\) impairs ToM capabilities without substantially affecting the model's overall linguistic performance.

\begin{tcolorbox}[
    colback=green!5,          
    colframe=green!20,         
    boxrule=0.5pt,            
    arc=4pt,                 
    boxsep=5pt,               
    left=6pt, right=6pt,      
    top=6pt, bottom=6pt,              
    fonttitle=\bfseries,      
    coltitle=black, 
    colbacktitle=green!10,
    colframe=green!20, 
    title=Findings 1        
]
\textit{An extremely sparse ToM-sensitive parameter pattern exists, whose perturbation significantly affects ToM capabilities, while random perturbations do not.} Our experiments further demonstrate that this degradation is linked to a reduction in \textit{contextual localization} and \textit{language understanding}.
\end{tcolorbox}

\subsection{Perturbing ToM-Sensitive Parameters Affects Positional Encoding}

We demonstrate that the ToM-sensitive parameter pattern impacts contextual localization by influencing the model’s positional encoding mechanism. For Transformer decoder-based models, a widely used positional encoding method is RoPE \citep{su_2021_roformer}.

\paragraph{RoPE and Feature Frequencies.} RoPE applies token position-dependent rotations to feature pairs in activations \(\mathbf{Q}\) and \(\mathbf{K}\). Formally, RoPE defines a rotational encoding angle as:
\begin{align*}
\theta(p, m) &= p \cdot \left(\frac{1}{50000}\right)^{\frac{2m}{d_h}},
\end{align*}
where \(p\) is the token position, \(m\) is the feature index within an attention head, \(d_h\) denotes the per-head feature dimension. The encoding applies a rotation matrix \(M(p, m)\) to each feature pair \(\mathbf{x}_p^m \in \mathbb{R}^{2}\):
\begin{align*}
\text{Enc}(\mathbf{x}_p^m , p, m) &= \begin{bmatrix}
\cos(\theta(p, m)) & -\sin(\theta(p, m)) \\
\sin(\theta(p, m)) & \cos(\theta(p, m))
\end{bmatrix} \cdot \mathbf{x}_p^m \\&= M(p, m) \cdot \mathbf{x}_p^m.
\end{align*}

Given two token activations \(\mathbf{q}_i, \mathbf{k}_j \in \mathbb{R}^{d_h}\), their RoPE-encoded activation interaction is:
\begin{align*}
\text{RoPE}(\mathbf{q}_i, \mathbf{k}_j) &= \sum_{m=0}^{d_h/2-1} 
\left(\text{Enc}(\mathbf{q}_i^m, i, m)\right)^\top 
\cdot \text{Enc}(\mathbf{k}_j^m, j, m) \\&= \sum_{m=0}^{d_h/2-1} 
(\mathbf{q}_i^m)^\top 
\cdot M(j - i, m) 
\cdot \mathbf{k}_j^m.
\end{align*}

\begin{figure}[t]
    \centering
    \includegraphics[width=0.6\linewidth]{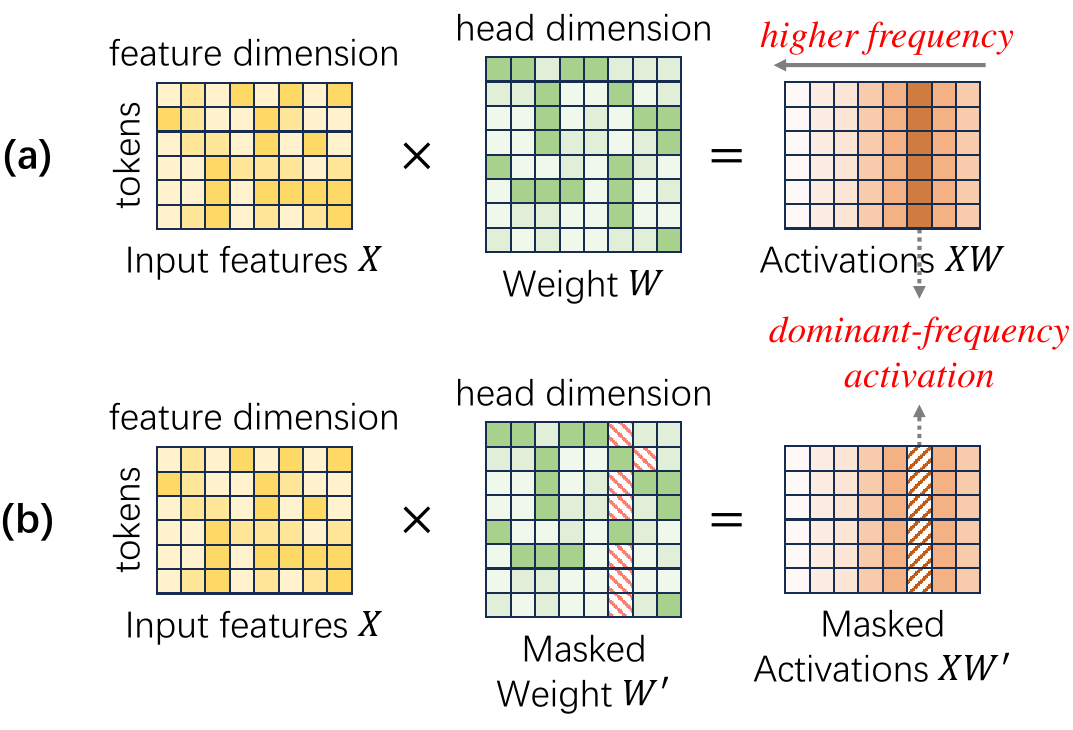}
    \caption{Activation calculations. (a) Original. We observe dominant-frequency activations introduced by RoPE. (b) Perturbing ToM-sensitive parameters (the squares with red diagonal lines in $\mathbf{W}'$). We observe that the ToM parameter pattern is highly frequency-sensitive and specifically affects dominant-frequency activations.}
    \label{fig:dominant}
\end{figure}

This formulation shows that RoPE assigns \textit{smaller encoding angles to later feature dimensions in \(\mathbf{Q}\) and \(\mathbf{K}\)}, meaning that these dimensions rotate more slowly across token positions. As a result, lower-indexed dimensions correspond to \textit{higher frequencies}, while higher-indexed dimensions correspond to \textit{lower frequencies} in the positional encoding.

\paragraph{Emergence of dominant-frequency activations.}  
Recent studies \citep{barbero_2024_round, hua_2024_fourier} have shown that activations tend to concentrate at certain frequencies, with \textit{low-frequency components} of \(\mathbf{Q}=XW_\mathbf{Q}\) and \(\mathbf{K}=XW_\mathbf{K}\) exhibiting higher magnitudes. One possible explanation is that low-frequency dimensions rotate more slowly, which may allow them to encode information more stably over longer token dependencies \citep{barbero_2024_round}. We observe that this phenomenon occurs specifically in models using RoPE, while it is absent in models without RoPE.

\noindent \paragraph{Perturbation Effects on RoPE Features.} We observe that the ToM-sensitive parameter pattern shares the \textit{same} dominant frequency as the activations in the weight matrix, as illustrated in Figure~\ref{fig:dominant} (b). Perturbing these sensitive parameters specifically affects the dominant-frequency activations. Therefore, perturbing the ToM-sensitive parameter pattern essentially disrupts the dominant-frequency activations constructed by positional encoding.

\begin{tcolorbox}[
    colback=green!5,          
    colframe=green!20,         
    boxrule=0.5pt,            
    arc=4pt,                 
    boxsep=5pt,               
    left=6pt, right=6pt,      
    top=6pt, bottom=6pt,              
    fonttitle=\bfseries,      
    coltitle=black, 
    colbacktitle=green!10,
    colframe=green!20, 
    title=Findings 2        
]
\textit{The functionality of the ToM-sensitive parameter pattern relates to the positional encoding module in LLM architectures.} Perturbing the proposed ToM-sensitive parameter pattern in LLMs with RoPE disrupts dominant-frequency activations induced by positional encoding, thereby impairing contextual localization. In contrast, LLMs without RoPE lack this frequency-dependent activation structure and exhibit different sensitivity patterns.

\end{tcolorbox}

\subsection{Perturbing ToM-sensitive parameter pattern affects attention mechanism}

In this section, we further demonstrate that the ToM-sensitive parameter pattern significantly influences the attention mechanism at the token level. Specifically, perturbing the pattern alters the \textit{geometric relationship} of \(\mathbf{k}_\text{BOS}\), changing the angle between \(\mathbf{q}\) and \(\mathbf{k}_\text{BOS}\) and shifting the attention sink. This distortion in the attention map ultimately impairs the model's ability to accurately comprehend semantics.

\begin{figure}[t]
  \centering
  \begin{minipage}[b]{0.5\textwidth}
    \centering
    \includegraphics[width=\textwidth]{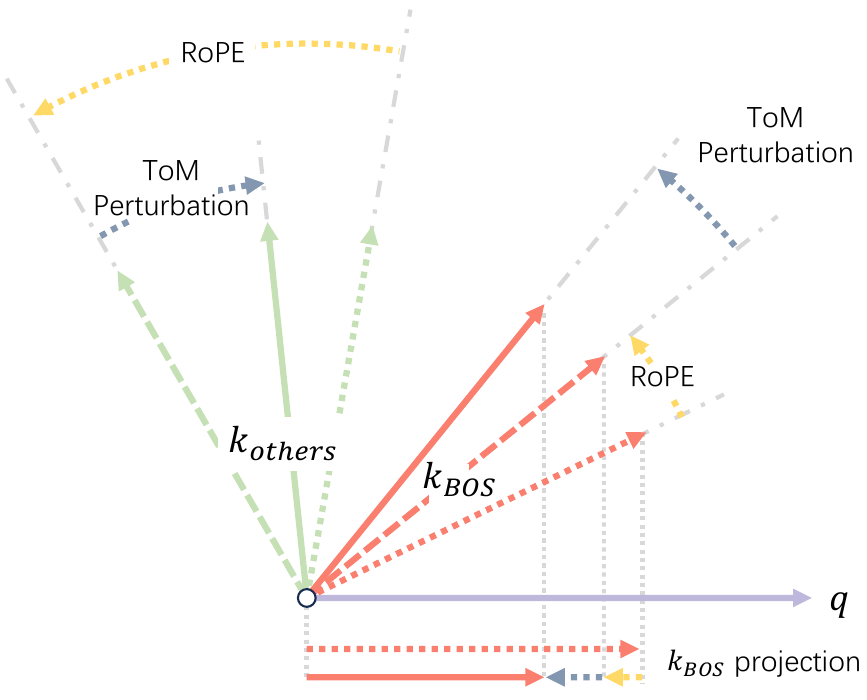}
    \caption{Visualization of the vector relationships between \(\mathbf{q}\) and \(\mathbf{k}_\text{BOS}\), as well as between \(\mathbf{q}\) and other tokens in \(\mathbf{K}\), under both positional encoding and ToM perturbation. }
    \label{fig:rope}
  \end{minipage}
  \hfill 
  \begin{minipage}[b]{0.42\textwidth}
    \centering
    \includegraphics[width=\textwidth]{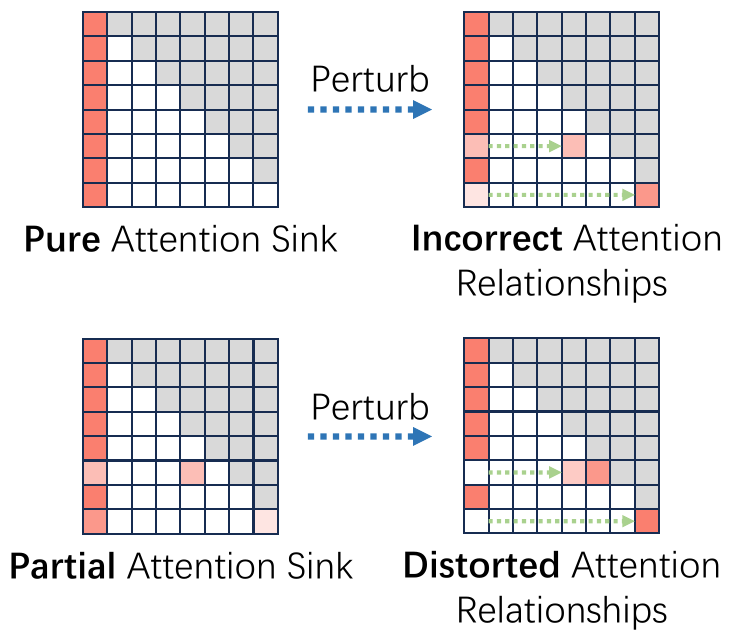}
    \caption{Attention sink shift. Shifting pure attention sinks introduces incorrect attention relationships, while shifting partial attention sinks distorts the original attention patterns. Attention sink shift degrades the model's language understanding capabilities evaluated by MMLU.}
    \label{fig:atten}
  \end{minipage}
\end{figure}

\paragraph{Attention sinks.}  
Recent studies have observed the \textit{attention sinks} phenomenon in LLMs, where the attention maps of most layers and heads predominantly focus on the relationship between the query token and \(\mathbf{k}_\text{BOS}\). This manifests as a prominent vertical line in the first column of the attention map \citep{xiao_2023_efficient, cancedda_2024_spectral}. Although the norm of \(\mathbf{k}_\text{BOS}\) is smaller than that of other tokens, its distinct manifold allows it to act as a bias, storing excess attention scores and leading to the attention sinks phenomenon \citep{ gu_2024_when}.

\paragraph{Perturbing ToM-sensitive parameters affects the geometric characteristics of \(\mathbf{k}_\text{BOS}\).}  
As shown in Figure~\ref{fig:rope}, we observe that \(\mathbf{q}\) is nearly \textit{orthogonal} to other token activations in \(\mathbf{K}\), and their inner products remain close to zero after encoding and perturbation. In contrast, the angle between \(\mathbf{q}\) and \(\mathbf{k}_\text{BOS}\) is consistently smaller than 90 degrees. Perturbing the ToM-sensitive parameter pattern significantly impacts the dominant-frequency components of \(\mathbf{k}_\text{BOS}\), changing its angle with \(\mathbf{q}\). Specifically, positional encoding reduces the angle between \(\mathbf{q}\) and \(\mathbf{k}_\text{BOS}\), increasing their inner product, while ToM perturbation rotates \(\mathbf{k}_\text{BOS}\) \textit{towards orthogonality}, thereby disrupting the positional encoding.

\paragraph{Perturbing ToM-sensitive parameters shifts the attention sinks.}  
Originally, the large inner product between \(\mathbf{q}\) and \(\mathbf{k}_\text{BOS}\) ensures the formation of attention sinks in the first column of the attention map. However, when this inner product decreases due to perturbation, the attention sinks become unstable, causing attention scores to be incorrectly distributed to other positions. As shown in Figure~\ref{fig:atten}, shifts in attention sinks can lead to incorrect embeddings, ultimately degrading the model's language understanding capabilities.

\begin{tcolorbox}[
    colback=green!5,          
    colframe=green!20,         
    boxrule=0.5pt,            
    arc=4pt,                 
    boxsep=5pt,               
    left=6pt, right=6pt,      
    top=6pt, bottom=6pt,              
    fonttitle=\bfseries,      
    coltitle=black, 
    colbacktitle=green!10,
    colframe=green!20, 
    title=Findings 3        
]
\textit{Perturbing ToM-sensitive parameter patterns affects the attention mechanism, thereby influencing language understanding.}  
Perturbing the ToM-sensitive parameter pattern alters the angle between \(\mathbf{q}\) and \(\mathbf{k}_\text{BOS}\) under positional encoding. This disruption breaks the RoPE encoding, causing \(\mathbf{q}\) and \(\mathbf{k}_\text{BOS}\) to become more orthogonal. As a result, the attention sink is destabilized, distorting the attention matrix and impairing the model's ability to capture correct feature relationships, ultimately diminishing its ToM capabilities.
\end{tcolorbox}

\section{Experiments and Validations}

\label{sec:exp}

In this section, we present experimental results to validate our three key findings. First, we investigate whether perturbing the ToM-sensitive parameter pattern affects both ToM and contextual localization and language understanding capabilities (Findings 1). Second, we analyze the distribution of this pattern and examine how it varies across models with different positional encoding schemes (Findings 2). Finally, we explore how perturbing the pattern alters the geometric characteristics of \(\mathbf{k}_\text{BOS}\), disrupts the angle introduced by RoPE, and leads to attention sink shifts, ultimately impacting semantic understanding (Findings 3).

\subsection{Experimental setup}
We utilize a diverse range of open-source models, including Llama \citep{dubey_2024_the}, Qwen \citep{qwen_2024_qwen25, yang_2024_qwen2}, DeepSeek \citep{deepseekai_2025_deepseekr1}, and Jamba \citep{lieber_2024_jamba, team_2024_jamba15}. Details of these models are provided in Appendix \ref{appendix:exp_models}. To identify ToM-sensitive parameters, we use constructed dataset \(\mathcal{D}_{\text{ToM-Train}}\) and C4 dataset \citep{raffel_2019_exploring} to estimate \(\mathbf{m}_\kappa\) and \(\mathbf{m}_\kappa'\). The dataset details are described in Appendix \ref{appendix:exp_dataset}.

For each model, we apply perturbations to \(W_\mathbf{Q}\), \(W_\mathbf{K}\), \(W_\mathbf{V}\), \(W_\mathbf{O}\), \(W_\mathbf{Gate}\), \(W_\mathbf{Up}\), and \(W_\mathbf{Down}\) matrices across layers, varying the mask sparsity level \(\kappa\). The perturbation is implemented by setting the sensitive mask parameters to the average value of the unmasked parameters in the corresponding matrix. The most impactful perturbation (in terms of ToM performance degradation) is reported, and the corresponding \(\kappa\) values were provided in Appendix \ref{appendix:exp_settings}. All models are evaluated under consistent settings unless otherwise specified.

\subsection{Validation of findings 1: ToM ability, perplexity, contextual localization ability, and semantic understanding ability}

In this section, we investigate the impact of perturbing ToM-sensitive parameter patterns on both ToM capabilities and language performance, as measured by perplexity. Perplexity is evaluated on the Wikitext-2 dataset \citep{merity_2016_pointer}, while ToM capabilities are assessed using the \(\mathcal{D}_{\text{ToM-Test}}\) benchmark \citep{kosinski_2024_evaluating}. Additionally, we examine how perturbing ToM-sensitive parameters affects contextual localization and language understanding. To evaluate contextual localization, we introduce a task that requires the model to accurately reproduce input sequences, measuring the similarity between input and output. For language understanding, we utilize the MMLU dataset \citep{hendrycks_2020_measuring, liang_2022_holistic} to assess the model's performance. See Appendix \ref{appendix:exp_examples} for examples of the datasets used in this study.

\paragraph{Extreme sparse sensitive parameter patterns impair ToM ability while minimally affecting perplexity in RoPE-based models, whereas random perturbations have no effect.}  
As shown in \tablename~\ref{table:tom-results}, masking parameters at a sparsity level as fine as \(10^{-5}\) leads to a substantial decline in ToM performance across all models, with only marginal changes in perplexity.  
Details on the search process for the optimal \(\kappa\) and results on random perturbations can be found in Appendix \ref{appendix:exp_tom}.

\begin{figure}[h]
    \centering
    \subfloat[Llama3-8B]{%
        \includegraphics[width=0.24\textwidth]{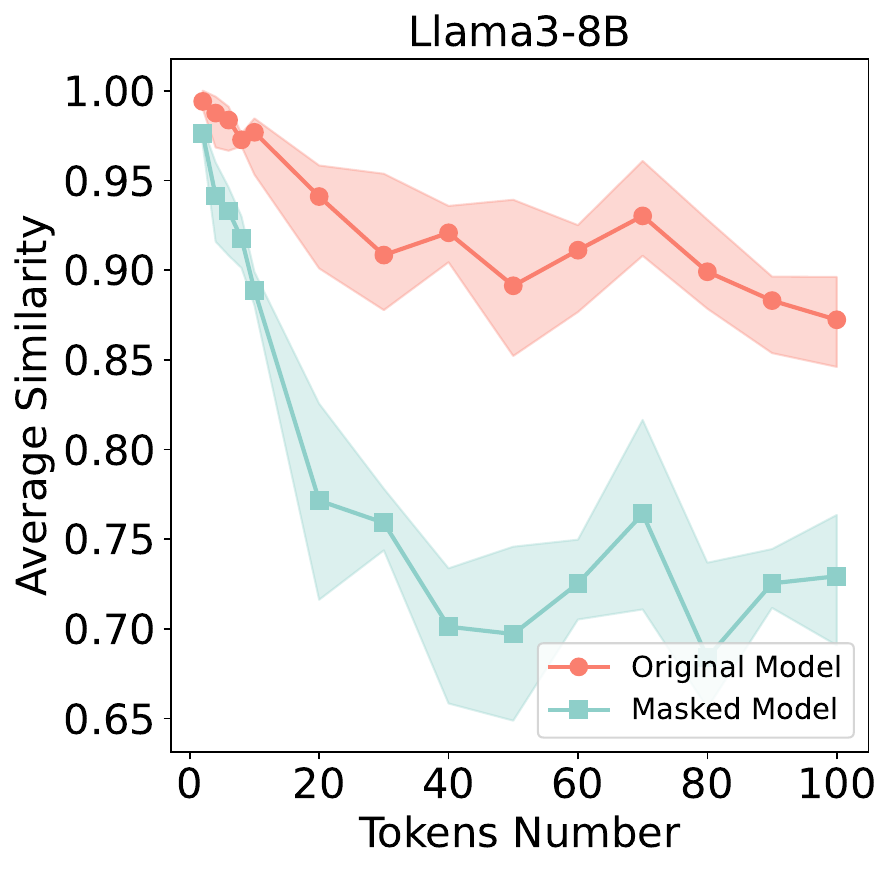}}
    \hfill
    \subfloat[Llama3.1-8B-Instruct]{%
\includegraphics[width=0.24\textwidth]{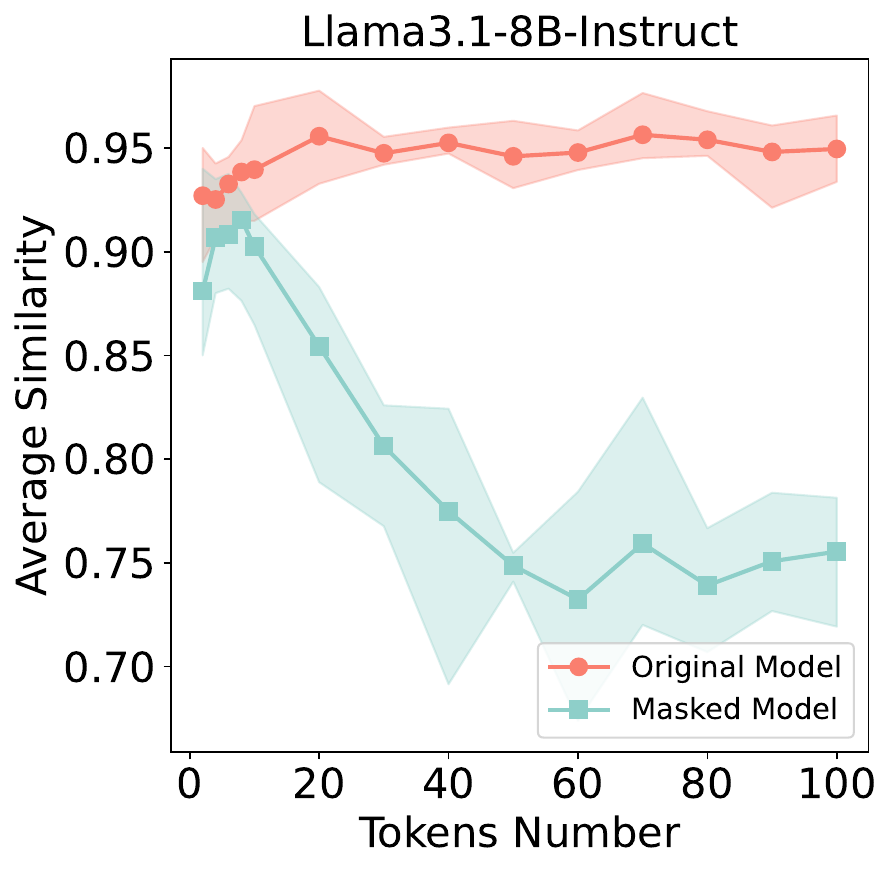}}
    \hfill
    \subfloat[Llama3.2-1B]{%
\includegraphics[width=0.24\textwidth]{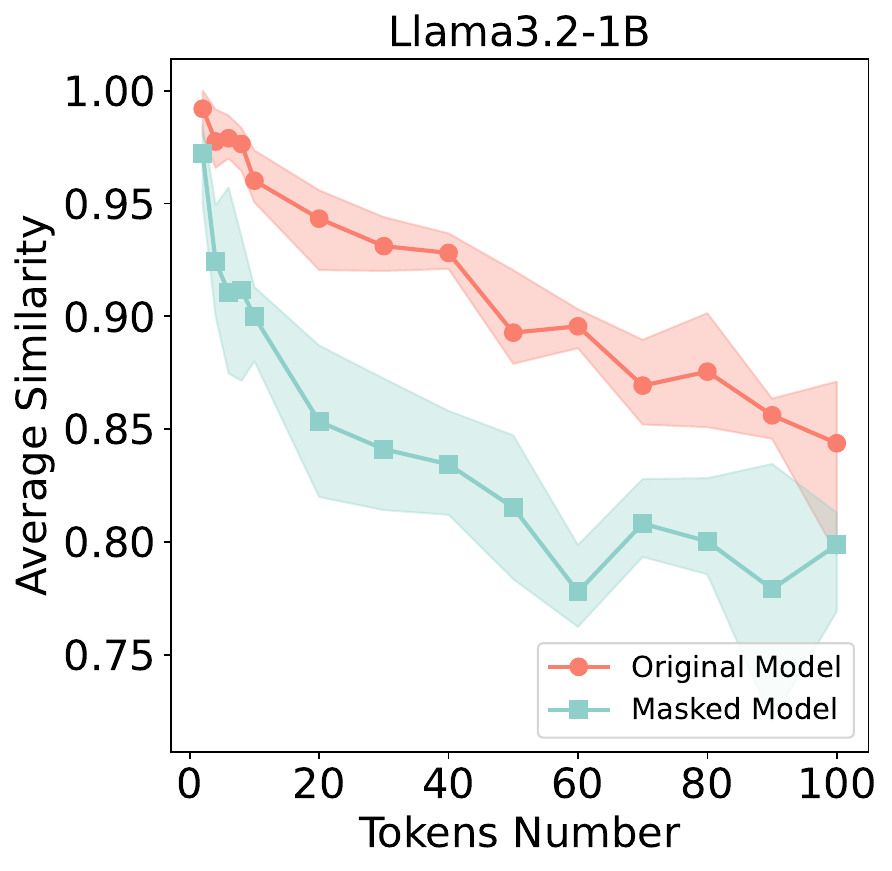}}
    \hfill
    \subfloat[Qwen2.5-7B-Instruct]{%
\includegraphics[width=0.235\textwidth]{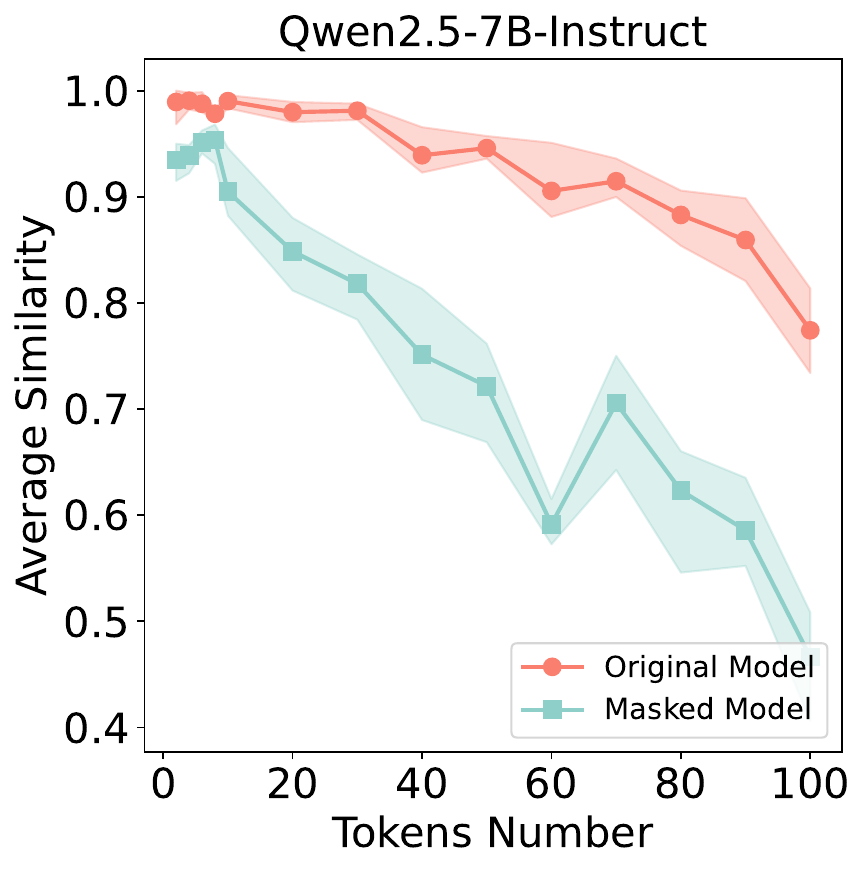}}
    \hfill

    \caption{Evaluating contextual localization ability across models. More results can be found in Appendix \ref{appendix:exp_contextualresults}.}
    \label{fig:localization}
\end{figure}

\begin{figure}[t]
  \centering
  \begin{minipage}[b]{0.48\textwidth}
    \centering
    \includegraphics[width=\textwidth]{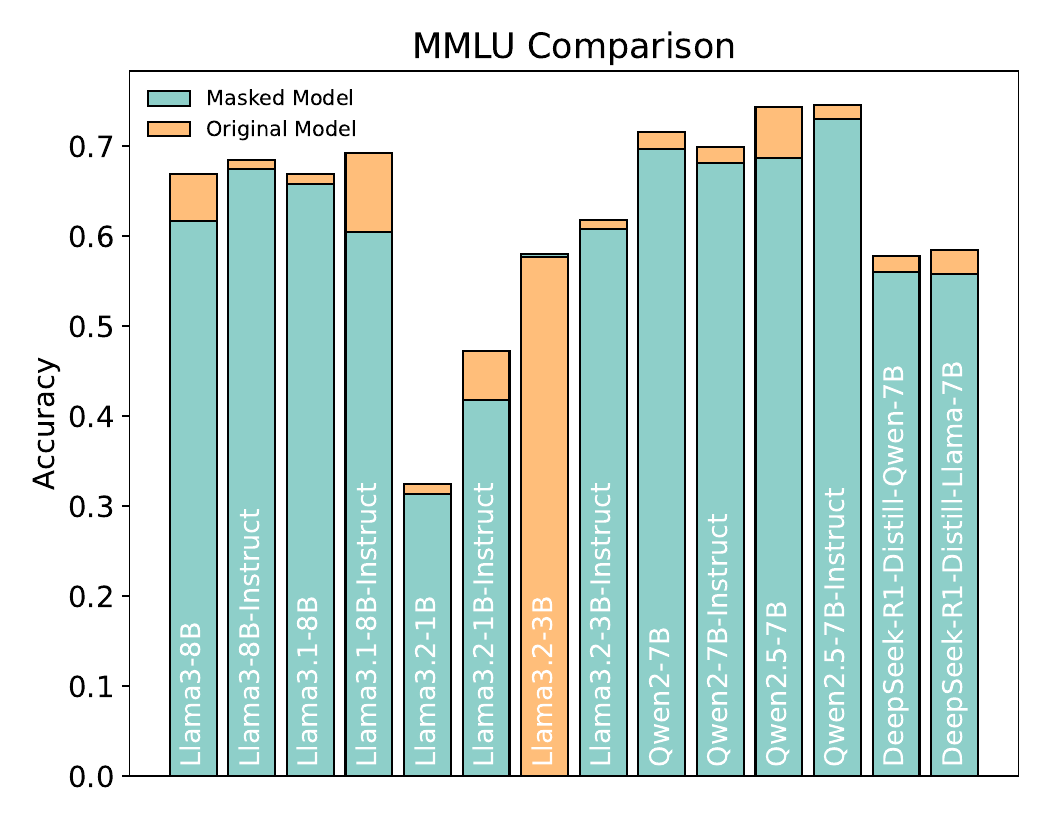}
    \caption{Overall accuracy comparison across models before and after perturbing parameters. For more results, please refer to Appendix \ref{appendix:exp_mmlu}.}
    \label{fig:overall_accuracy}
  \end{minipage}
  \hfill 
  \begin{minipage}[b]{0.48\textwidth}
    \centering
    \includegraphics[width=\textwidth]{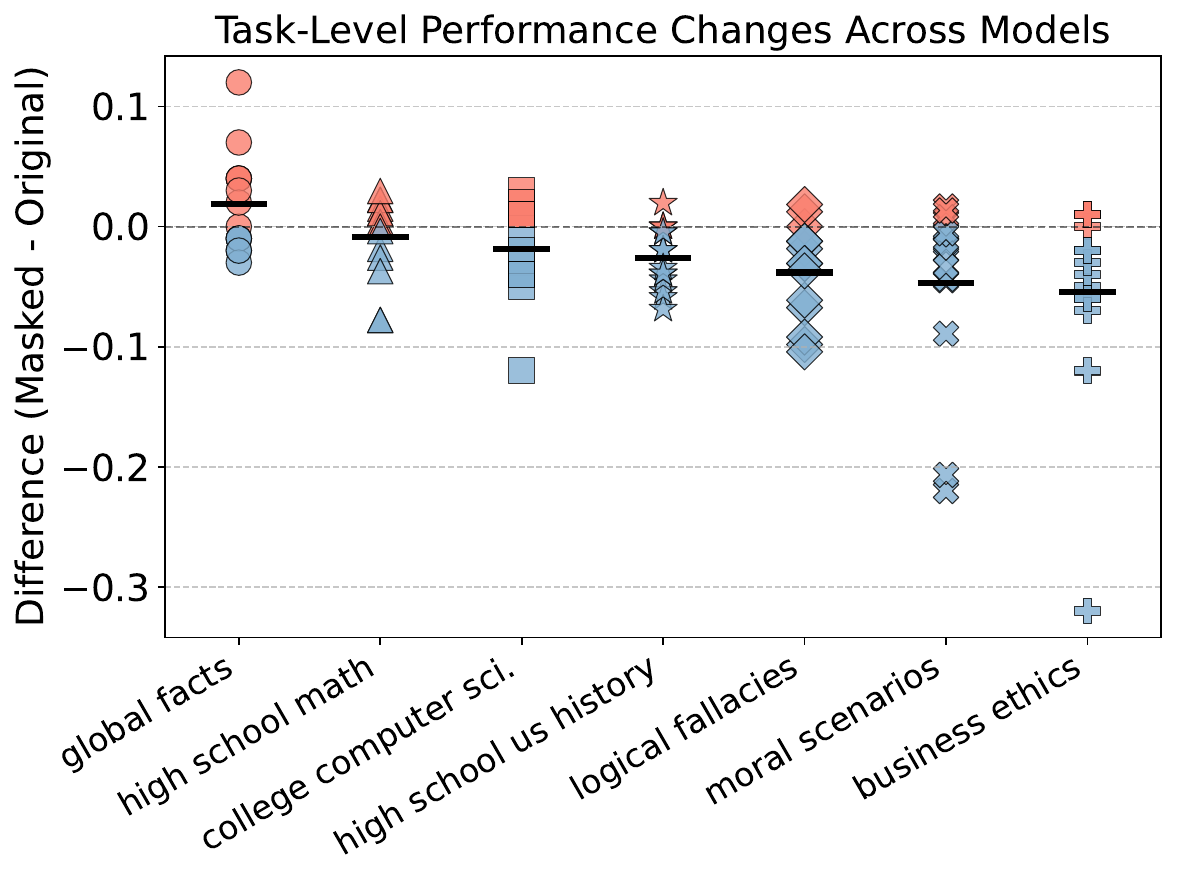}
    \caption{Task-level performance differences across selected tasks. The black horizontal bars indicate the average difference for each task.}
    \label{fig:task_level}
  \end{minipage}
\end{figure}

\paragraph{The ToM-sensitive parameter pattern also impacts contextual localization and language understanding in RoPE-based models.} As shown in Figure \ref{fig:localization}, these models exhibit significantly degraded positioning performance, particularly for longer repeated token sequences. Simultaneously, perturbing these parameters leads to a performance decline across most models on the MMLU benchmark, as illustrated in \figurename~\ref{fig:overall_accuracy}. Notably, as shown in \figurename~\ref{fig:task_level}, ToM-related tasks such as business ethics experience the most significant performance drops.

\paragraph{Non-RoPE-based models exhibit distinct behavior.} We found no parameter patterns that significantly degrade ToM performance in non-RoPE-based models. For instance, the Jamba-1.5-Mini model showed improved ToM task performance and reduced perplexity. This indicates that non-RoPE-based models also possess ToM capabilities, but their mechanisms for storing and processing such intelligence differ from those of RoPE-based models. The absence of RoPE encoding prevents the emergence of dominant-frequency activations, making the pattern ineffective for perturbing the encoding mechanism. More results are provided in Appendix \ref{appendix:exp_jambaresults}.

\subsection{Validation of findings 2: The  characteristics of ToM-sensitivate parameters and their impact on positional encoding}

\paragraph{The ToM-sensitive parameter pattern is sparse and low-rank, with significant perturbations in $W_{\mathbf{Q}}$ and $W_{\mathbf{K}}$ matrices.} 
In Llama3-8B, the average mask rank for $W_{\mathbf{Q}}$ and $W_{\mathbf{K}}$ matrices is 21.69 and 10.5, indicating a strong low-rank structure. Additionally, the perturbed weights in $W_{\mathbf{Q}}$ and $W_{\mathbf{K}}$ matrices are significantly larger compared to other matrices, suggesting that changes in model performance are closely tied to the attention mechanism. For detailed results, please refer to Appendix~\ref{appendix:C_rank} and ~\ref{appendix:C_weightvalue}.

\begin{table}[H]
\caption{Performance of different models across ToM tasks and perplexity. \textbf{P} denotes the version with the sensitive pattern perturbed, and \textbf{Ins} represents the Instruct-tuned variant of the model. The abbreviations for ToM tasks are as follows: \textbf{FB} (False Belief), \textbf{CL} (Correct Label), \textbf{IP} (Informed Protagonist), \textbf{OC} (Open Container), \textbf{NT} (No Transfer), and \textbf{PP} (Present Protagonist). \underline{Underlined values} indicate a decline in model performance after perturbation.}
\label{table:tom-results}
\centering
\resizebox{\textwidth}{!}{
\begin{tabular}{clrrrrrrrrrr}
\toprule
\multirow{2}{*}{} & \multirow{2}{*}{\textbf{Model}} 
& \multicolumn{4}{c}{\textbf{Unexpected Contents}} 
& \multicolumn{4}{c}{\textbf{Unexpected Transfer}}
& \multirow{2}{*}{\textbf{Avg.  \((\uparrow)\)}} 
& \multirow{2}{*}{\textbf{PPL \((\downarrow)\)}} \\
\cmidrule(lr){3-6}\cmidrule(lr){7-10}
 &  & FB & CL & IP & OC & FB & NT & IP & PP &  &  \\
\midrule

\multirow{16}{*}{Llama}
& 3-8B & 66.00 & 83.50 & 94.50 & 42.00 & 48.00 & 63.00 & 73.00 & 23.50 & 61.69 & 6.14\\

& \cellcolor[HTML]{F0F0F0}3-8B-P 
& \cellcolor[HTML]{F0F0F0} \uline{32.00} & \cellcolor[HTML]{F0F0F0} \uline{82.50} & \cellcolor[HTML]{F0F0F0} \uline{81.50} & \cellcolor[HTML]{F0F0F0} 50.00 & \cellcolor[HTML]{F0F0F0} \uline{20.00} & \cellcolor[HTML]{F0F0F0} \uline{50.50} & \cellcolor[HTML]{F0F0F0} \uline{50.50} & \cellcolor[HTML]{F0F0F0} 25.00 & \cellcolor[HTML]{F0F0F0} \uline{49.00}& \cellcolor[HTML]{F0F0F0} \uline{7.46}\\

& 3-8B-Ins  & 87.50 & 74.00 & 89.50 & 41.00 & 68.00 & 60.50 & 47.00 & 19.00 & 60.81 & 8.30\\

& \cellcolor[HTML]{F0F0F0}3-8B-Ins-P 
& \cellcolor[HTML]{F0F0F0} 96.00& \cellcolor[HTML]{F0F0F0} \uline{63.50}& \cellcolor[HTML]{F0F0F0} \uline{66.50}& \cellcolor[HTML]{F0F0F0} \uline{17.00}
& \cellcolor[HTML]{F0F0F0} \uline{64.00}& \cellcolor[HTML]{F0F0F0} 60.50& \cellcolor[HTML]{F0F0F0} \uline{23.00}& \cellcolor[HTML]{F0F0F0} 23.00
& \cellcolor[HTML]{F0F0F0} \uline{51.69}& \cellcolor[HTML]{F0F0F0} 8.25\\

& 3.1-8B  & 68.50 & 80.50 & 94.50 & 40.50 & 46.00 & 61.00 & 73.50 & 20.00 & 60.56 &6.25  \\

& \cellcolor[HTML]{F0F0F0}3.1-8B-P 
 & \cellcolor[HTML]{F0F0F0} \uline{67.00} & \cellcolor[HTML]{F0F0F0} \uline{64.50} & \cellcolor[HTML]{F0F0F0} \uline{69.00} & \cellcolor[HTML]{F0F0F0} \uline{33.00} & \cellcolor[HTML]{F0F0F0} \uline{39.00} & \cellcolor[HTML]{F0F0F0} \uline{56.50} & \cellcolor[HTML]{F0F0F0} \uline{53.50} & \cellcolor[HTML]{F0F0F0} 25.00 & \cellcolor[HTML]{F0F0F0} \uline{50.94} & \cellcolor[HTML]{F0F0F0} \uline{6.44}\\

& 3.1-8B-Ins  & 81.50 & 69.00 & 79.00 & 61.00 & 63.50 & 64.50 & 71.00 & 29.50 & 64.88 & 7.22 \\

& \cellcolor[HTML]{F0F0F0}3.1-8B-Ins-P 
& \cellcolor[HTML]{F0F0F0} \uline{43.00} & \cellcolor[HTML]{F0F0F0} \uline{62.00} & \cellcolor[HTML]{F0F0F0} \uline{61.50} & \cellcolor[HTML]{F0F0F0} \uline{48.00} & \cellcolor[HTML]{F0F0F0} \uline{27.00} & \cellcolor[HTML]{F0F0F0} \uline{53.50} & \cellcolor[HTML]{F0F0F0} \uline{59.00} & \cellcolor[HTML]{F0F0F0} \uline{29.00} & \cellcolor[HTML]{F0F0F0} \uline{47.88} & \cellcolor[HTML]{F0F0F0} \uline{8.37}\\

& 3.2-1B  & 20.50 & 82.00 & 89.00 & 44.00 & 18.50 & 43.50 & 78.00 & 38.00 & 51.69&  9.77\\

& \cellcolor[HTML]{F0F0F0}3.2-1B-P 
& \cellcolor[HTML]{F0F0F0} \uline{17.50} & \cellcolor[HTML]{F0F0F0} \uline{58.50} & \cellcolor[HTML]{F0F0F0} \uline{74.50} & \cellcolor[HTML]{F0F0F0} \uline{39.00} & \cellcolor[HTML]{F0F0F0} \uline{10.00} & \cellcolor[HTML]{F0F0F0} \uline{35.50} & \cellcolor[HTML]{F0F0F0} \uline{60.00} & \cellcolor[HTML]{F0F0F0} \uline{22.00} & \cellcolor[HTML]{F0F0F0} \uline{39.62}& \cellcolor[HTML]{F0F0F0} \uline{10.46}\\

& 3.2-1B-Ins  & 20.00 & 99.00 & 97.50 & 63.50 & 14.50 & 47.00 & 72.00 & 39.50 & 56.63 &  13.18\\

& \cellcolor[HTML]{F0F0F0}3.2-1B-Ins-P 
& \cellcolor[HTML]{F0F0F0} \uline{13.50} & \cellcolor[HTML]{F0F0F0} \uline{79.00} & \cellcolor[HTML]{F0F0F0} \uline{86.50} & \cellcolor[HTML]{F0F0F0} \uline{35.00} & \cellcolor[HTML]{F0F0F0} \uline{11.00} & \cellcolor[HTML]{F0F0F0} \uline{40.00} & \cellcolor[HTML]{F0F0F0} \uline{36.50} & \cellcolor[HTML]{F0F0F0} \uline{20.00} & \cellcolor[HTML]{F0F0F0} \uline{40.19} & \cellcolor[HTML]{F0F0F0} \uline{14.79}\\

& 3.2-3B  & 59.00 & 55.00 & 81.50 & 43.50 & 31.00 & 47.00 & 70.00 & 18.00 & 50.63 & 7.82 \\

& \cellcolor[HTML]{F0F0F0}3.2-3B-P 
& \cellcolor[HTML]{F0F0F0} \uline{48.00} & \cellcolor[HTML]{F0F0F0} 60.00 & \cellcolor[HTML]{F0F0F0} \uline{72.00} & \cellcolor[HTML]{F0F0F0} \uline{33.00} & \cellcolor[HTML]{F0F0F0} \uline{25.00} & \cellcolor[HTML]{F0F0F0} \uline{41.00} & \cellcolor[HTML]{F0F0F0} \uline{49.50} & \cellcolor[HTML]{F0F0F0} \uline{15.00} & \cellcolor[HTML]{F0F0F0} \uline{42.94} & \cellcolor[HTML]{F0F0F0} \uline{7.86}\\

& 3.2-3B-Ins  & 56.00 & 66.50 & 92.00 & 61.50 & 29.00 & 62.00 & 71.50 & 44.50 & 60.38 &  11.06\\

& \cellcolor[HTML]{F0F0F0}3.2-3B-Ins-P 
& \cellcolor[HTML]{F0F0F0} \uline{50.00} & \cellcolor[HTML]{F0F0F0} \uline{60.50} & \cellcolor[HTML]{F0F0F0} \uline{81.00} & \cellcolor[HTML]{F0F0F0} \uline{44.00} & \cellcolor[HTML]{F0F0F0} \uline{24.50} & \cellcolor[HTML]{F0F0F0} \uline{51.50} & \cellcolor[HTML]{F0F0F0} \uline{61.00} & \cellcolor[HTML]{F0F0F0} \uline{36.00} & \cellcolor[HTML]{F0F0F0} \uline{51.06} & \cellcolor[HTML]{F0F0F0} \uline{11.44}\\
\midrule

\multirow{8}{*}{Qwen}
& 2-7B  & 50.00 & 87.50 & 87.50 & 75.00 & 27.50 & 72.50 & 75.00 & 42.50 & 64.69 & 7.14 \\

& \cellcolor[HTML]{F0F0F0}2-7B-P & \cellcolor[HTML]{F0F0F0} 52.50 & \cellcolor[HTML]{F0F0F0} \uline{67.50} & \cellcolor[HTML]{F0F0F0} \uline{52.50} & \cellcolor[HTML]{F0F0F0} \uline{40.00} & \cellcolor[HTML]{F0F0F0} \uline{25.00} & \cellcolor[HTML]{F0F0F0} \uline{65.00} & \cellcolor[HTML]{F0F0F0} \uline{50.00} & \cellcolor[HTML]{F0F0F0} \uline{30.00} & \cellcolor[HTML]{F0F0F0} \uline{47.81}&\cellcolor[HTML]{F0F0F0} \uline{7.70}\\

& 2-7B-Ins  & 42.50 & 85.50 & 83.50 & 66.50 & 24.00 & 66.00 & 64.50 & 38.50 & 58.88 & 7.60 \\

& \cellcolor[HTML]{F0F0F0}2-7B-Ins-P 
& \cellcolor[HTML]{F0F0F0} 47.50 & \cellcolor[HTML]{F0F0F0} \uline{67.00} & \cellcolor[HTML]{F0F0F0} \uline{64.00} & \cellcolor[HTML]{F0F0F0} \uline{38.50} & \cellcolor[HTML]{F0F0F0} \uline{12.00} & \cellcolor[HTML]{F0F0F0} \uline{47.00} & \cellcolor[HTML]{F0F0F0} \uline{43.00} & \cellcolor[HTML]{F0F0F0} \uline{31.50} & \cellcolor[HTML]{F0F0F0} \uline{43.81} & \cellcolor[HTML]{F0F0F0} \uline{8.53}\\

& 2.5-7B  & 55.00 & 75.00 & 92.50 & 80.00 & 42.50 & 62.50 & 70.00 & 57.50 & 66.88 & 6.85 \\

& \cellcolor[HTML]{F0F0F0}2.5-7B-P 
& \cellcolor[HTML]{F0F0F0} \uline{25.00} & \cellcolor[HTML]{F0F0F0} \uline{62.50} & \cellcolor[HTML]{F0F0F0} \uline{65.00} & \cellcolor[HTML]{F0F0F0} \uline{52.50} & \cellcolor[HTML]{F0F0F0} \uline{12.50} & \cellcolor[HTML]{F0F0F0} \uline{42.50} & \cellcolor[HTML]{F0F0F0} \uline{55.00} & \cellcolor[HTML]{F0F0F0} \uline{32.50} & \cellcolor[HTML]{F0F0F0} \uline{43.44} & \cellcolor[HTML]{F0F0F0} \uline{8.12}\\

& 2.5-7B-Ins  & 18.50 & 47.00 & 77.00 & 58.00 & 10.50 & 35.50 & 47.50 & 12.00 & 38.25& 7.46  \\

& \cellcolor[HTML]{F0F0F0}2.5-7B-Ins-P 
& \cellcolor[HTML]{F0F0F0} 54.50 & \cellcolor[HTML]{F0F0F0} 56.00 & \cellcolor[HTML]{F0F0F0} \uline{40.00} & \cellcolor[HTML]{F0F0F0} \uline{45.50} & \cellcolor[HTML]{F0F0F0} 13.00 & \cellcolor[HTML]{F0F0F0} 41.50 & \cellcolor[HTML]{F0F0F0} \uline{39.50} & \cellcolor[HTML]{F0F0F0} 15.00 & \cellcolor[HTML]{F0F0F0} \uline{38.13} & \cellcolor[HTML]{F0F0F0} \uline{8.20}\\

\midrule

\multirow{4}{*}{DeepSeek} 
& Llama-8B   & 28.00 & 71.50 & 82.50 & 65.50 & 25.50 & 74.50 & 65.00 & 27.50 & 55.00 & 13.15\\

& \cellcolor[HTML]{F0F0F0}Llama-8B-P 
& \cellcolor[HTML]{F0F0F0} \uline{16.50} & \cellcolor[HTML]{F0F0F0} \uline{49.00} & \cellcolor[HTML]{F0F0F0} 85.00 & \cellcolor[HTML]{F0F0F0} \uline{62.00} & \cellcolor[HTML]{F0F0F0} \uline{19.00} & \cellcolor[HTML]{F0F0F0} \uline{67.00} & \cellcolor[HTML]{F0F0F0} \uline{62.50} & \cellcolor[HTML]{F0F0F0} 29.00 & \cellcolor[HTML]{F0F0F0} \uline{48.75} & \cellcolor[HTML]{F0F0F0} \uline{14.53}\\
& Qwen-7B   & 26.50 & 91.50 & 90.00 & 63.00 & 16.00 & 63.00 & 46.50 & 6.50 & 50.38 & 25.06 \\

& \cellcolor[HTML]{F0F0F0}Qwen-7B-P 
& \cellcolor[HTML]{F0F0F0} \uline{20.00} & \cellcolor[HTML]{F0F0F0} \uline{79.00} & \cellcolor[HTML]{F0F0F0} \uline{85.50} & \cellcolor[HTML]{F0F0F0} \uline{52.50} & \cellcolor[HTML]{F0F0F0} 16.50 & \cellcolor[HTML]{F0F0F0} \uline{52.50} & \cellcolor[HTML]{F0F0F0} \uline{25.50} & \cellcolor[HTML]{F0F0F0} 9.50 & \cellcolor[HTML]{F0F0F0} \uline{42.63} & \cellcolor[HTML]{F0F0F0} \uline{28.30}\\

\midrule

\multirow{2}{*}{Jamba} 
& 1.5-Mini   & 74.00 & 45.50 & 93.00 & 50.50 & 60.50 & 65.50 & 77.50 & 28.00 & 61.81 & 7.77 \\

& \cellcolor[HTML]{F0F0F0}1.5-Mini-P 
& \cellcolor[HTML]{F0F0F0} \uline{73.00} & \cellcolor[HTML]{F0F0F0} 53.00 & \cellcolor[HTML]{F0F0F0} \uline{90.00} & \cellcolor[HTML]{F0F0F0} \uline{41.00} & \cellcolor[HTML]{F0F0F0} 62.50 & \cellcolor[HTML]{F0F0F0} 77.00 & \cellcolor[HTML]{F0F0F0} 78.50 & \cellcolor[HTML]{F0F0F0} 32.50 & \cellcolor[HTML]{F0F0F0} 63.44 & \cellcolor[HTML]{F0F0F0} 7.67\\

\bottomrule
\end{tabular}
}
\end{table}

\begin{figure}[t]
    \centering
    \subfloat[Llama3-8B-Q]{%
        \includegraphics[width=0.32\textwidth]{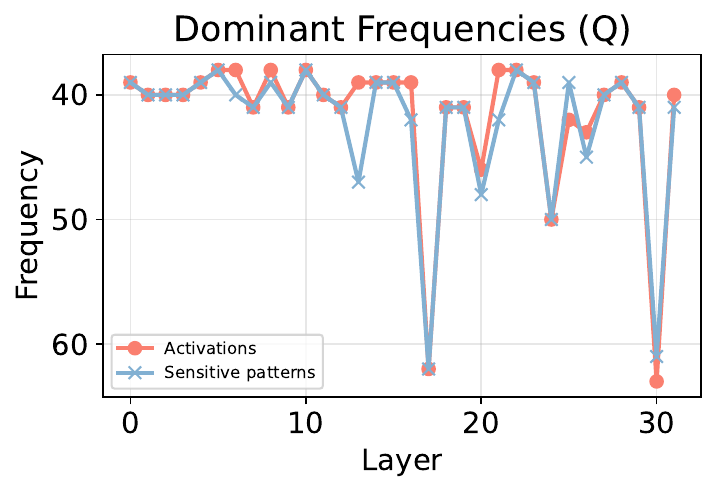}}
    \hfill
    \subfloat[Llama3-8B-K]{%
        \includegraphics[width=0.32\textwidth]{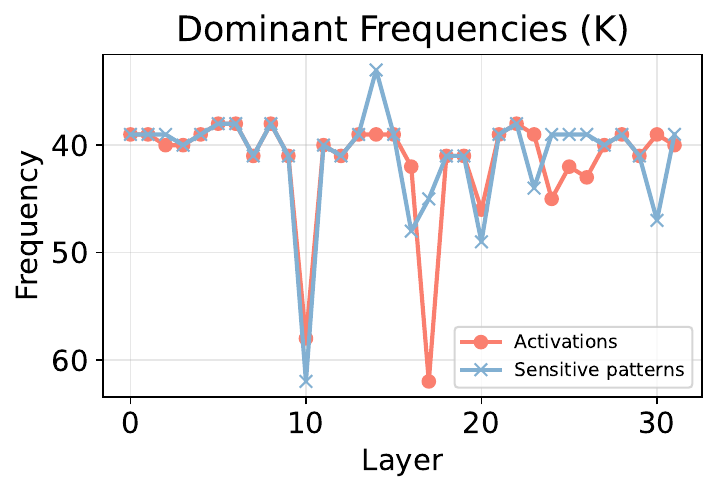}}
    \hfill
    \subfloat[Jamba1.5-Mini-Q]{%
        \includegraphics[width=0.32\textwidth]{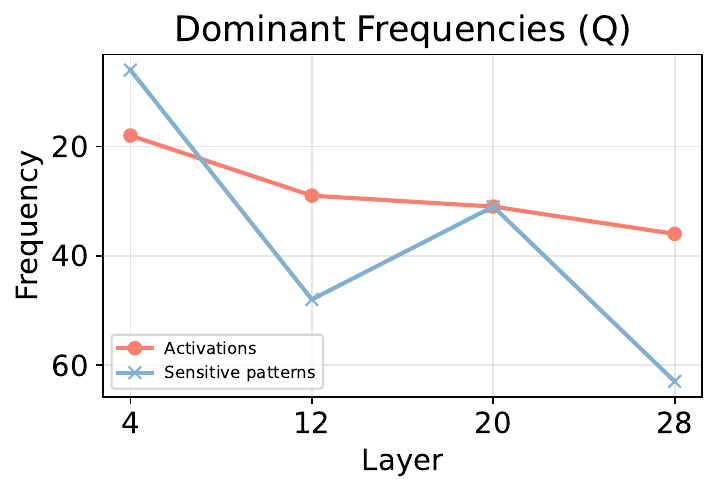}}
    
    \caption{Comparison of dominant-frequencies in the activation map and the ToM-sensitive parameter pattern. The figures depict the feature frequency corresponding to the maximum activation norm and the closest frequency among the top three most frequently perturbed frequencies in the ToM-sensitive parameter pattern. }
    \label{fig:activation}
    \vspace{-5pt}
\end{figure}

\paragraph{The ToM-sensitive parameter pattern perturbs dominant-frequency activations and affects positional encoding.}  
As shown in Figure ~\ref{fig:activation}, the frequency with the highest activation norm closely aligns with the most frequently perturbed frequencies in the ToM-sensitive parameter pattern. This suggests that the pattern primarily targets dominant-frequency activations, potentially influencing the model's positional encoding mechanism. However, this alignment is not observed in Jamba. Unlike other models, Jamba does not employ RoPE, and its activations lack a clear dominant frequency. Consequently, the ToM-sensitive parameter pattern in Jamba cannot affect contextual localization through positional encoding. Visualizations are provided in Appendix~\ref{appendix:C_activations}.

\subsection{Validation of findings 3: From positional encoding to attention map}

\begin{center}
\begin{minipage}{0.55\textwidth}  
\centering
\captionof{table}{Amplitude and angle of activation embeddings before RoPE, after RoPE, and after Perturbation}
\label{tab:rope}
\resizebox{\textwidth}{!}{  
\begin{tabular}{lccccc}
\toprule
& \textbf{Before} & \textbf{After} & \textbf{After} & \textbf{Change} & \textbf{Change} \\
& \textbf{RoPE (0)} & \textbf{RoPE (1)} & \textbf{Perturb. (2)} & \textbf{(0→1)} & \textbf{(1→2)} \\
\midrule
\(\|\mathbf{q}\|_2\)          & 12.95          & 12.95          & 12.76          & 0.00           & -0.19          \\
\(\|\mathbf{k}_\text{BOS}\|_2\)      & 4.22           & 4.22           & 3.91           & 0.00           & -0.31          \\
\(\|\mathbf{k}_\text{others}\|_2\) & 22.48      & 22.48          & 22.19          & 0.00           & -0.30          \\
\(\angle(\mathbf{q}, \mathbf{k}_\text{BOS})\)   & 66.35          & 66.46          & 69.22          & 0.11           & 2.77           \\
\(\angle(\mathbf{q}, \mathbf{k}_\text{others})\) & 93.34    & 96.81          & 95.20          & 3.47           & -1.62          \\
\bottomrule
\end{tabular}
}
\end{minipage}
\hfill  
\begin{minipage}{0.44\textwidth}  
\centering
\includegraphics[width=\textwidth]{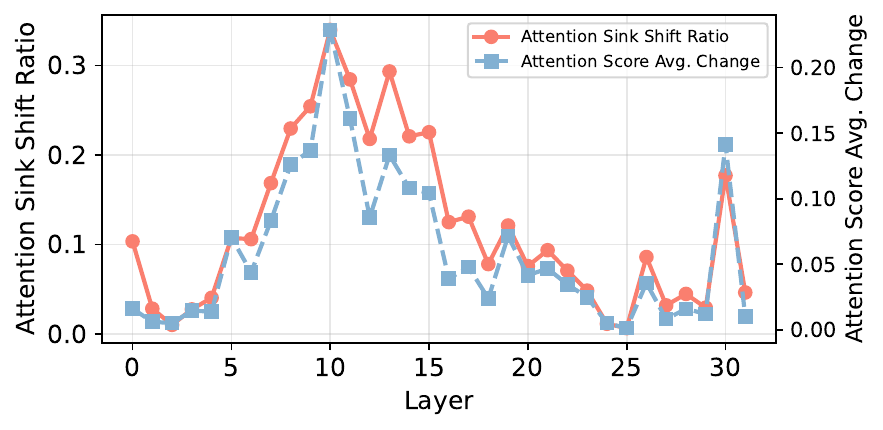}  
\captionof{figure}{Attention sink shift ratio and first token attention score change across layers.}  
\label{fig:attention_sink_shift}
\end{minipage}
\end{center}

\paragraph{Perturbing the ToM parameter pattern shifts attention sinks.} We set a threshold of 0.01, and if the change in attention sinks exceeds this value, we consider the attention sink to have shifted. As shown in Figure \ref{fig:attention_sink_shift}, we observe that more than 30\% of the sinks in layer 10 undergo a shift, which severely disrupts the attention structure. Such perturbations cause the model to incorrectly select invalid features in \( W_{\mathbf{V}} \), ultimately impairing its semantic understanding capabilities.

\paragraph{Perturbing the ToM parameter pattern also damages the RoPE.} We select the \(\mathbf{q}\) tokens at positions where attention sink shifts occur and compute their angles with \(\mathbf{k}_{\text{BOS}}\) and \(\mathbf{k}_{\text{others}}\). As shown in Table \ref{tab:rope}, we find that the magnitudes of the vectors remain largely unchanged before and after perturbation, and \(\mathbf{q}\) remains nearly orthogonal to \(\mathbf{k}_{\text{others}}\), with little change in their inner product. However, for the angle between \(\mathbf{q}\) and \(\mathbf{k}_{\text{BOS}}\), we observe that the change introduced by RoPE is minimal, whereas the ToM perturbation causes a significant angular shift. This perturbation completely overwhelms the positional information encoded by RoPE, explaining the decline in the model's contextual localization ability. Additionally, it leads to a smaller inner product between \(\mathbf{q}\) and \(\mathbf{k}_{\text{BOS}}\), destabilizing the attention sink and causing shifts that further degrade semantic understanding.

\begin{figure}[t]
    \centering
    \includegraphics[width=0.8\linewidth]{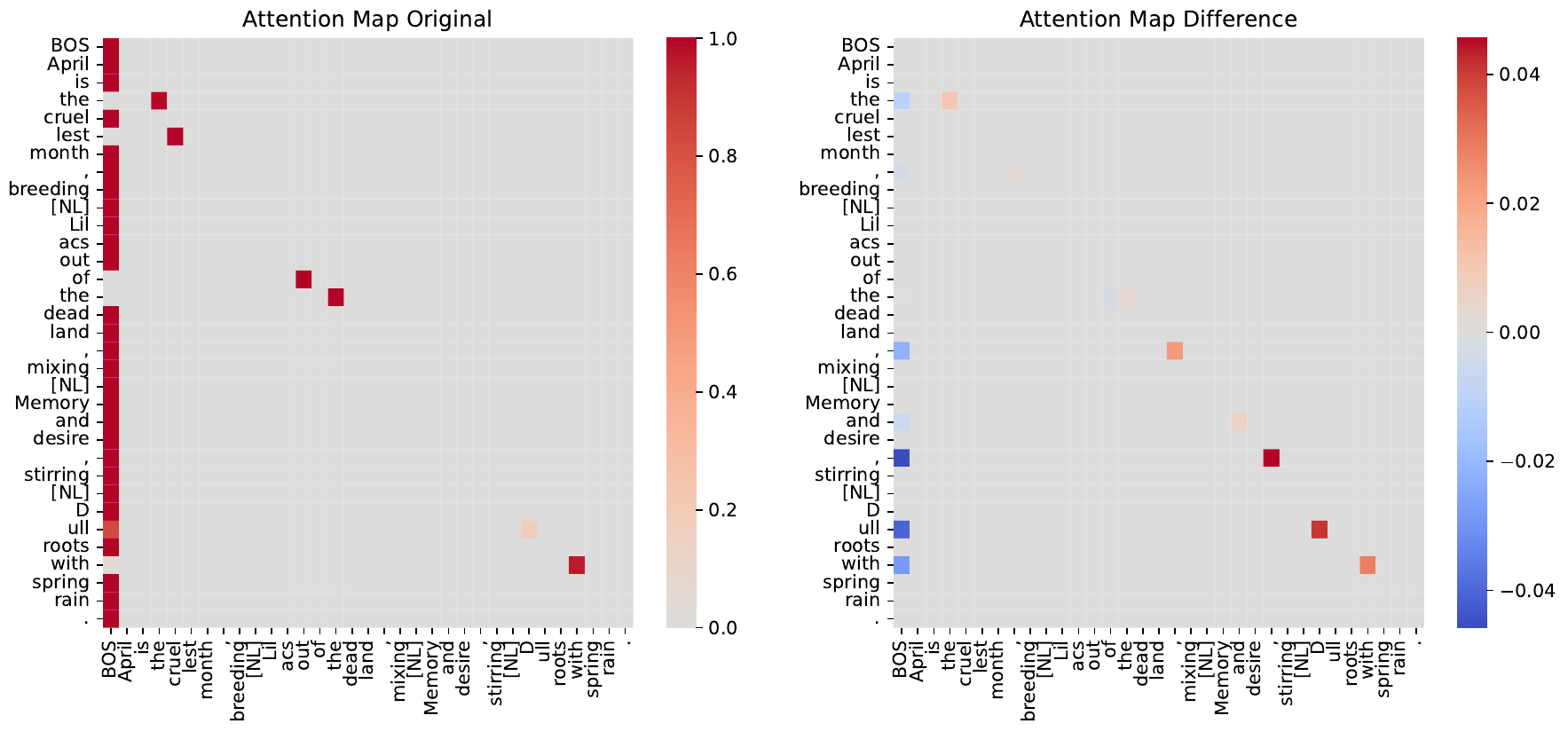}
    \caption{Example of attention sink shift from Llama 3-8B layer 0 head 6. The example sentence is the first several lines of T. S. Eliot's long poem \textit{The Waste Land}. Note that the attention values are not divided by the scaling factor before the softmax operation.}
    \label{fig:sinkv}
\end{figure}

\paragraph{Visualization of attention sink shift.} 
As shown in Figure \ref{fig:sinkv}, perturbing ToM-sensitive parameters introduces two key distortions. First, incorrect attention relationships emerge: an attention head originally attending to function words such as ``the'' (article), ``of'' (preposition), and ``-lest'' (subordinating conjunction) begins misallocating attention to punctuation marks like commas. Second, existing attention relationships are distorted: the attention scores assigned to certain tokens are altered, which undermine the model’s ability to maintain stable feature representations, impairing its overall language understanding capabilities.

\section{Conclusion}

In this article, we proposed a method to identify sparse low-rank ToM-sensitive parameter patterns. We discovered that these patterns affect the LLM's ToM ability and influence its contextual localization and language understanding capabilities.
We found that the impact of these patterns on performance is closely related to the LLM architecture. For LLMs using RoPE encoding, perturbing these patterns damages frequency-dominant activations, impairing the encoding mechanism and leading to performance degradation. Further analysis revealed that these patterns affect the geometric characteristics of the \(\mathbf{k}_{\text{BOS}}\) token, overwhelming the information encoded by RoPE and causing attention sink shifts. This results in inaccurate feature relationships and ultimately degrades the LLM's ability to understand language.

\bibliographystyle{unsrtnat}
\bibliography{iclr2025_conference}

\appendix
\section*{Appendix}

\section{Visualization of approximated Hessian matrix}
\label{appendix:hessian_fisher_visualization}

\begin{figure}[h]
    \centering
    \subfloat[Q]{%
        \includegraphics[width=0.33\textwidth]{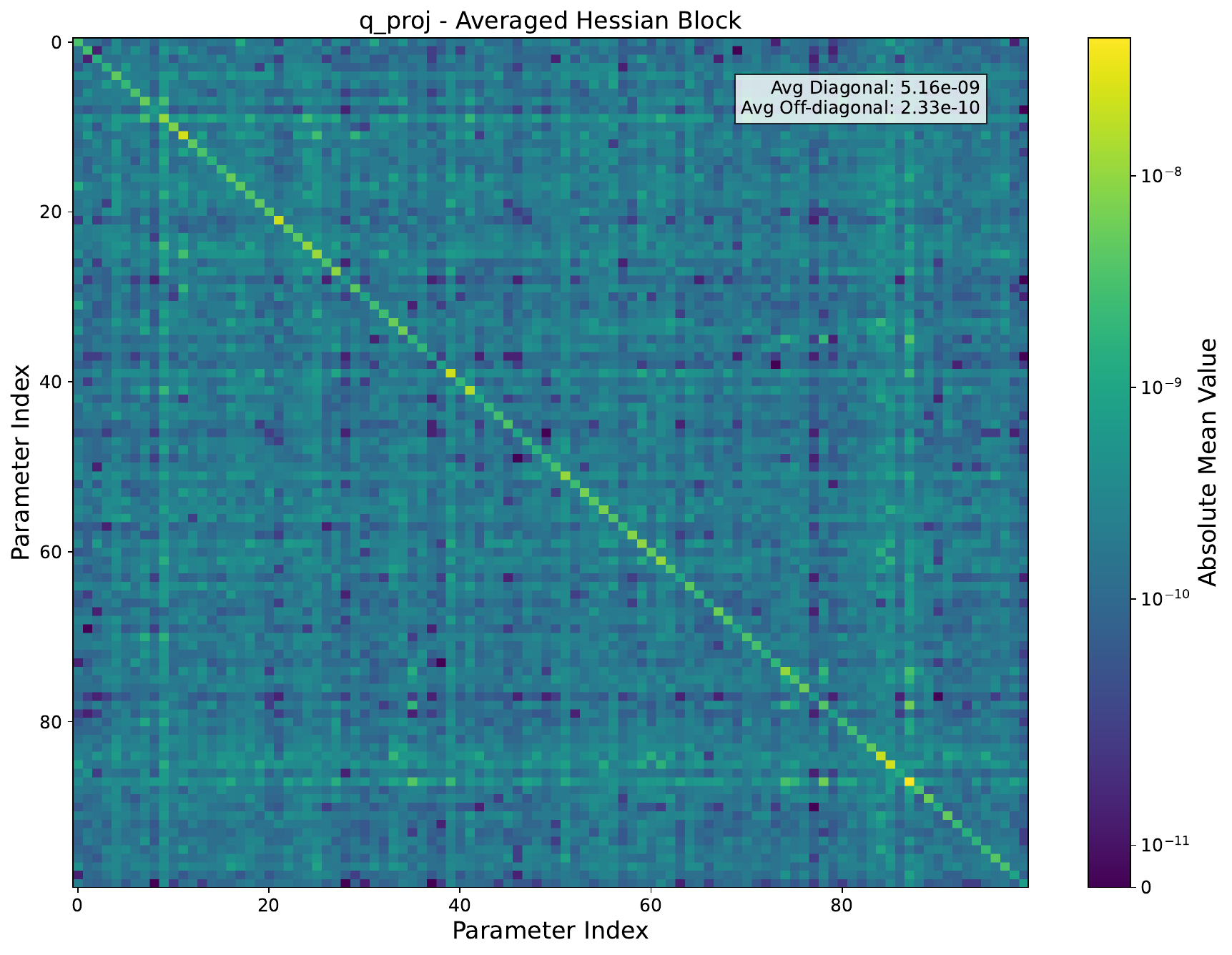}}
    \hfill
    \subfloat[K]{%
        \includegraphics[width=0.33\textwidth]{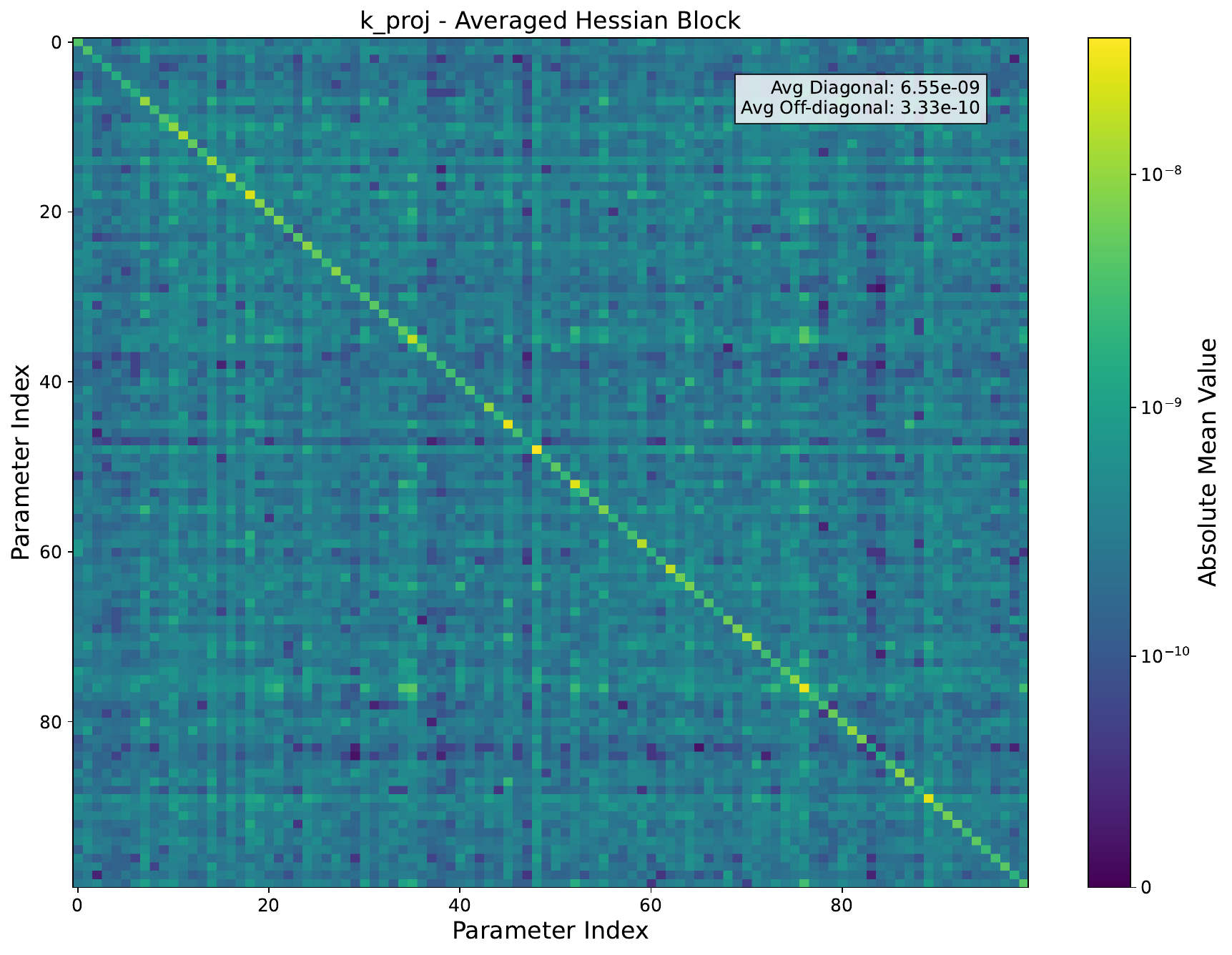}}
    \hfill
    \subfloat[V]{%
        \includegraphics[width=0.33\textwidth]{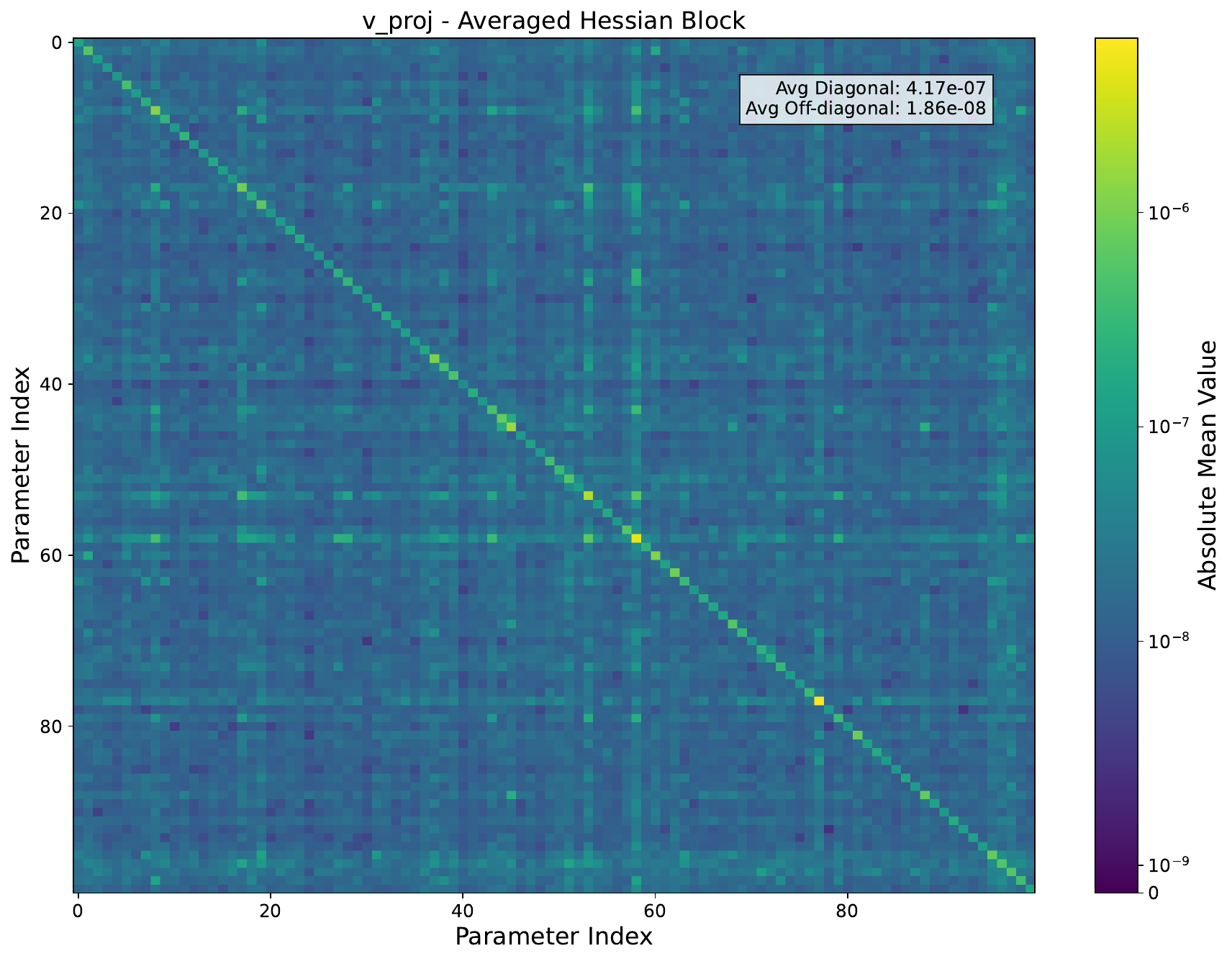}}
    
    \subfloat[O]{%
        \includegraphics[width=0.24\textwidth]{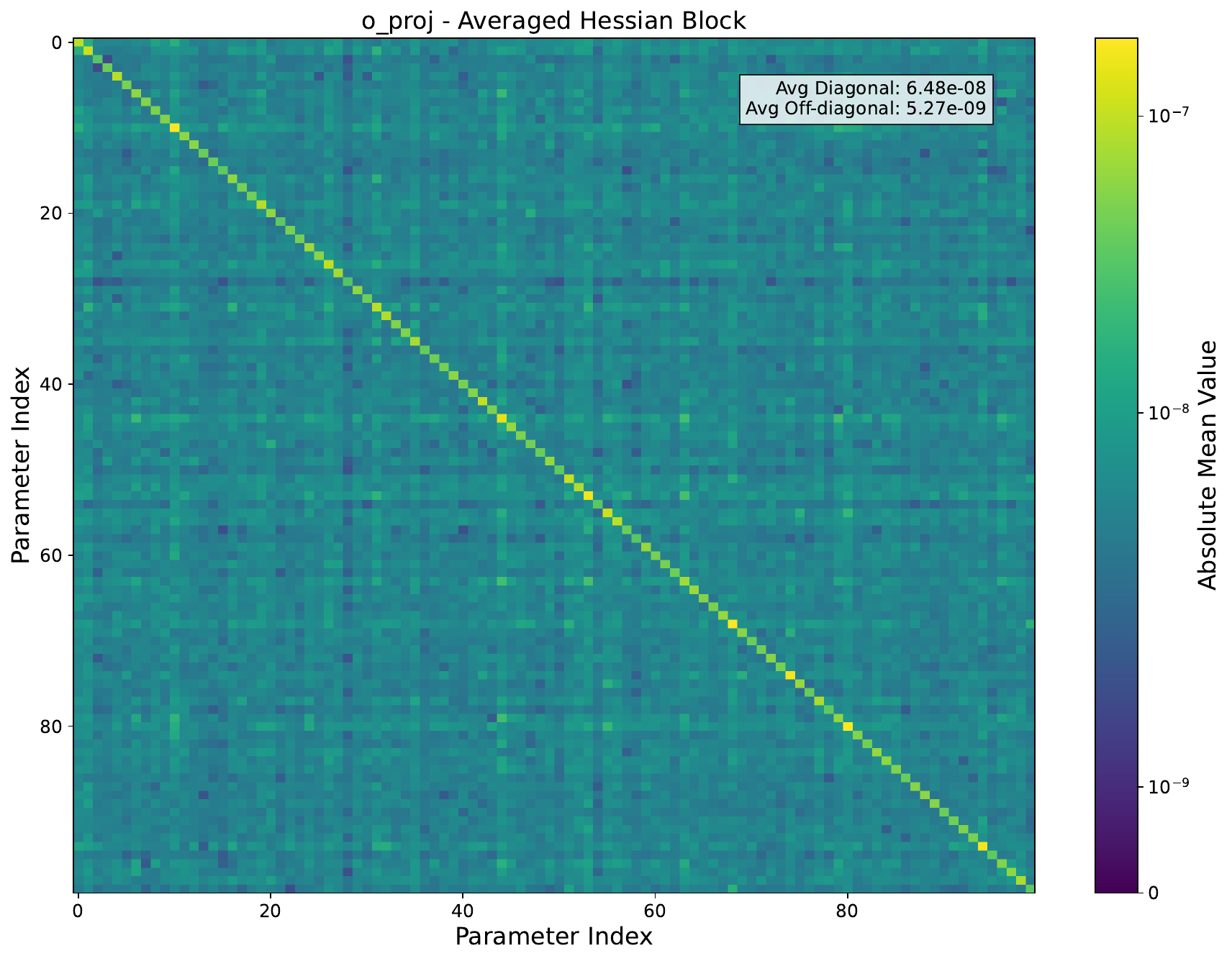}}
    \hfill
    \subfloat[Gate]{%
        \includegraphics[width=0.24\textwidth]{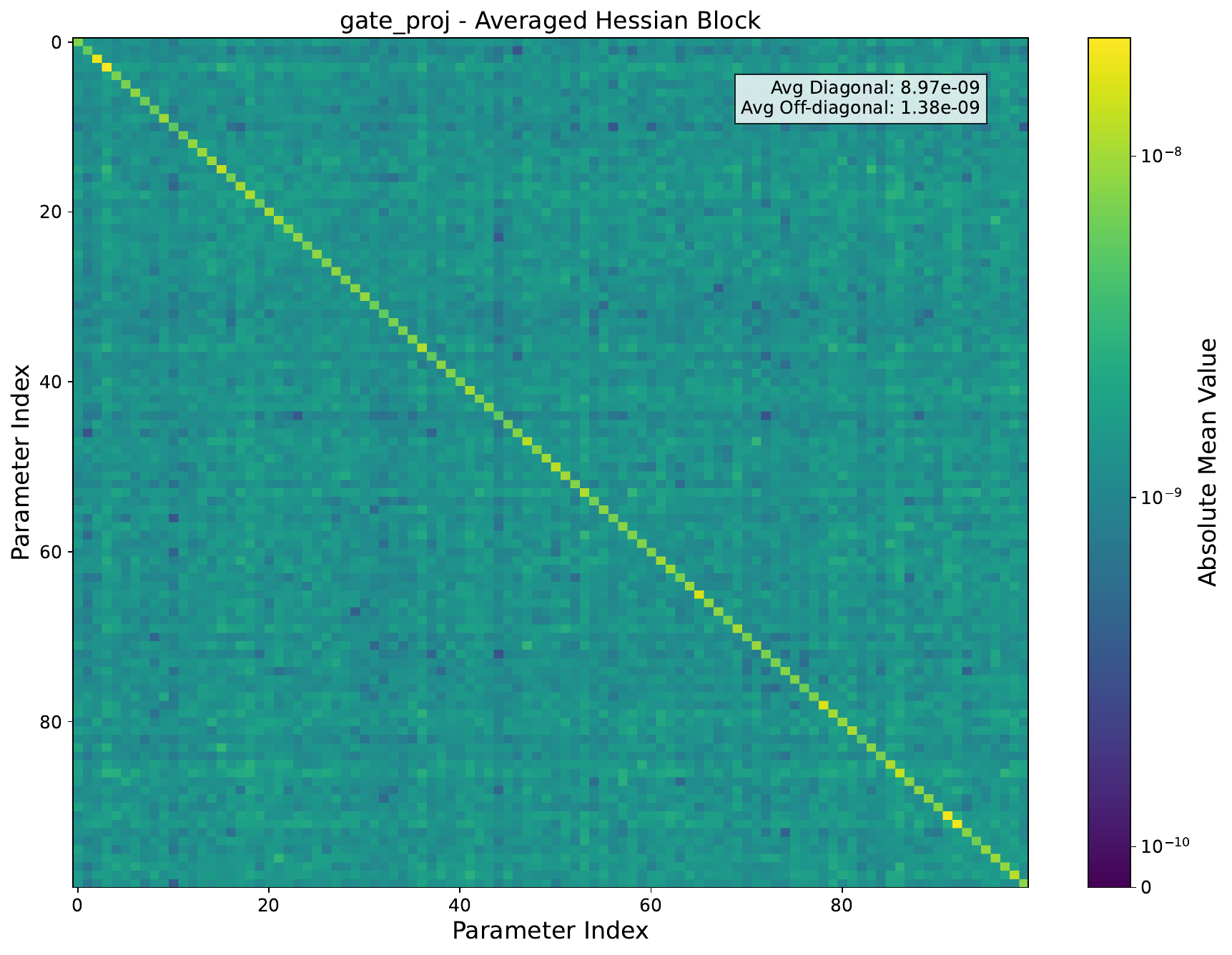}}
    \hfill
    \subfloat[Up]{%
        \includegraphics[width=0.24\textwidth]{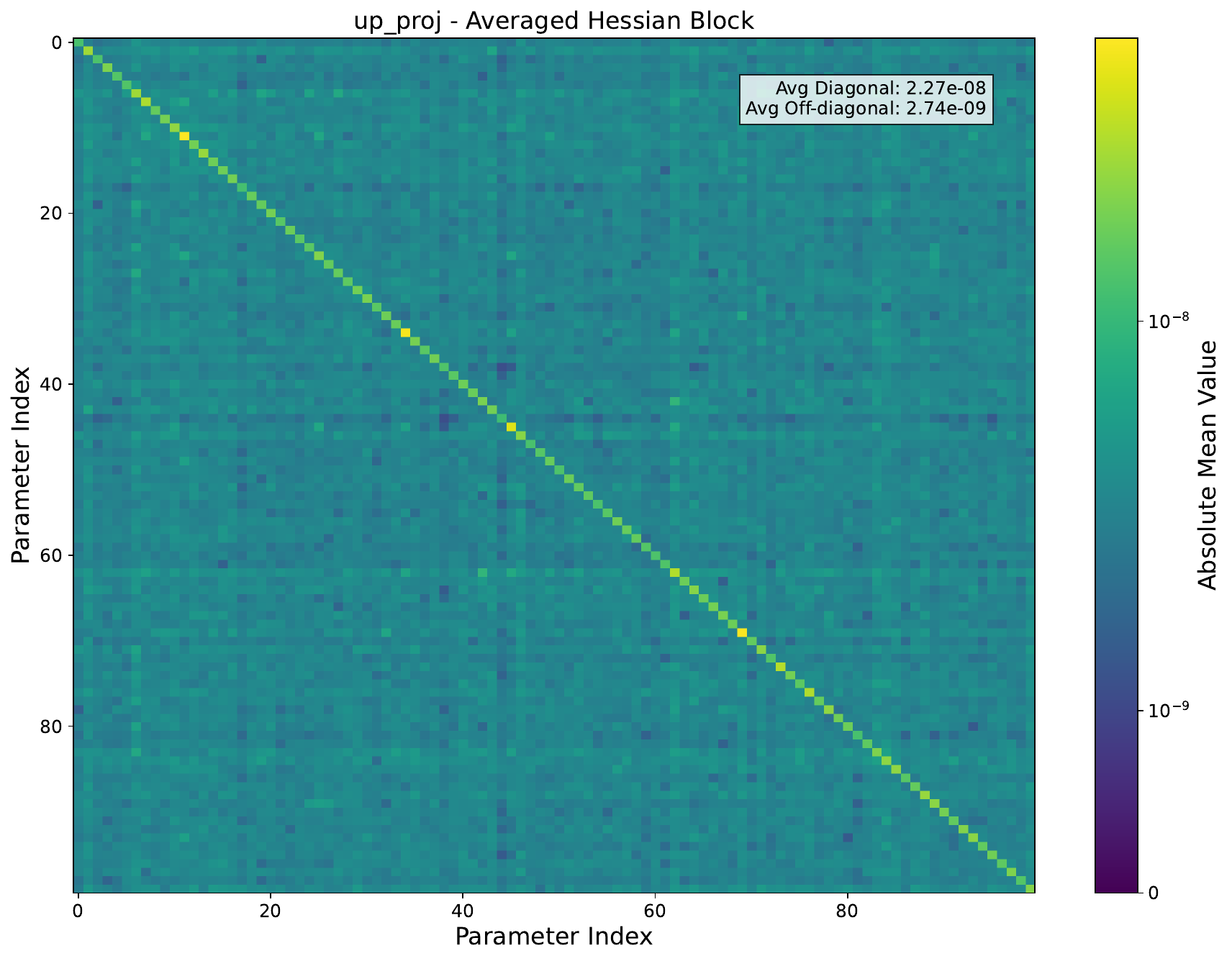}}
    \hfill
    \subfloat[Down]{%
        \includegraphics[width=0.24\textwidth]{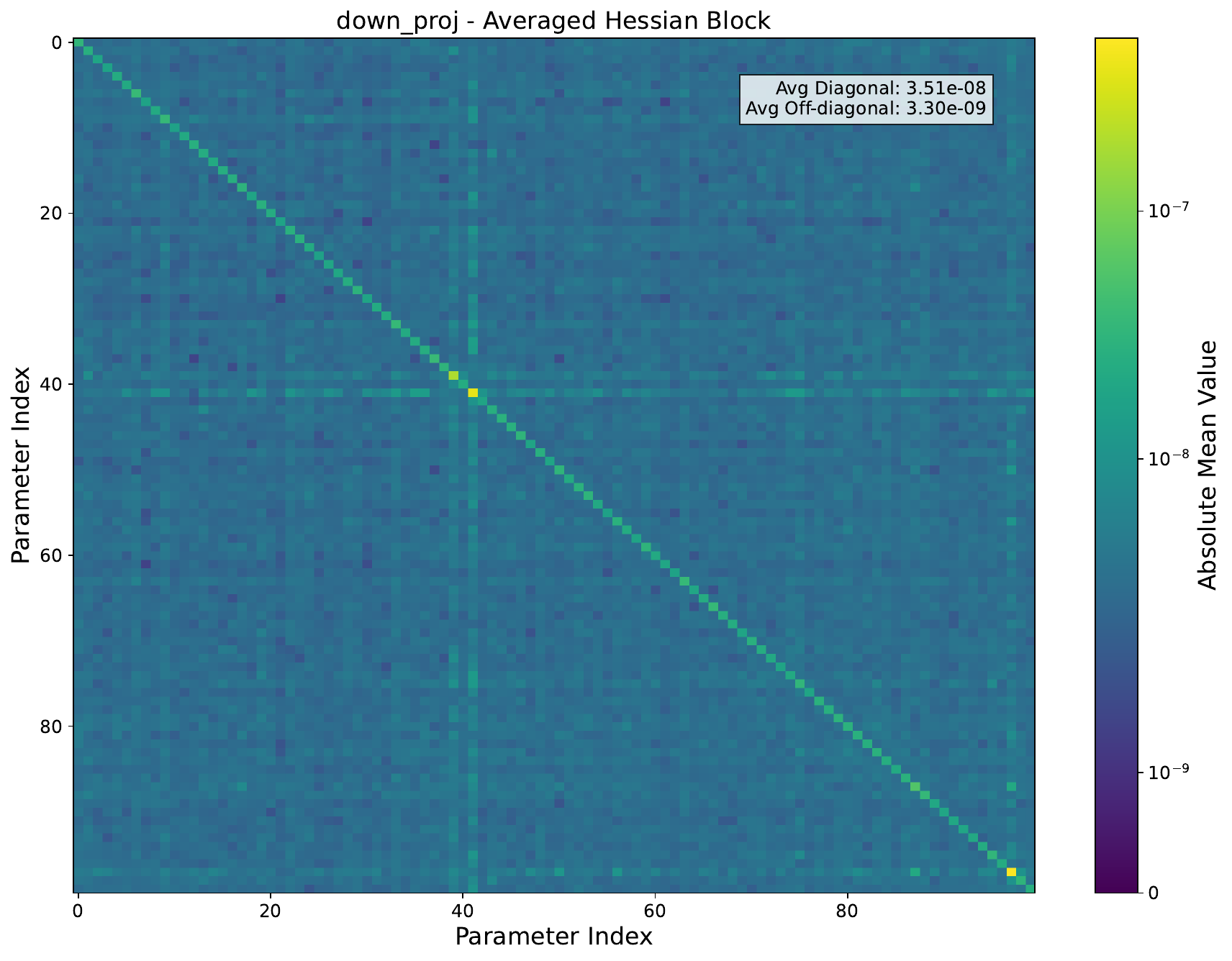}}
    
    \caption{Visulization of Hessian matrix.}
    \label{fig:hessian}
\end{figure}

In this section, we analyze the approximated Hessian matrices for the LLaMA3-8B model. For each layer, we randomly sample 100 points from the \(W_\mathbf{Q}\), \(W_\mathbf{K}\), \(W_\mathbf{V}\), \(W_\mathbf{O}\), \(W_\mathbf{Gate}\), \(W_\mathbf{Up}\), and \(W_\mathbf{Down}\) matrices. These samples are used to construct a subset of the empirical Hessian matrix. We visualize them matrices (normalized across layers) in Figure \ref{fig:hessian}.

\begin{figure}[h]
    \centering
    \subfloat[Q]{%
        \includegraphics[width=0.33\textwidth]{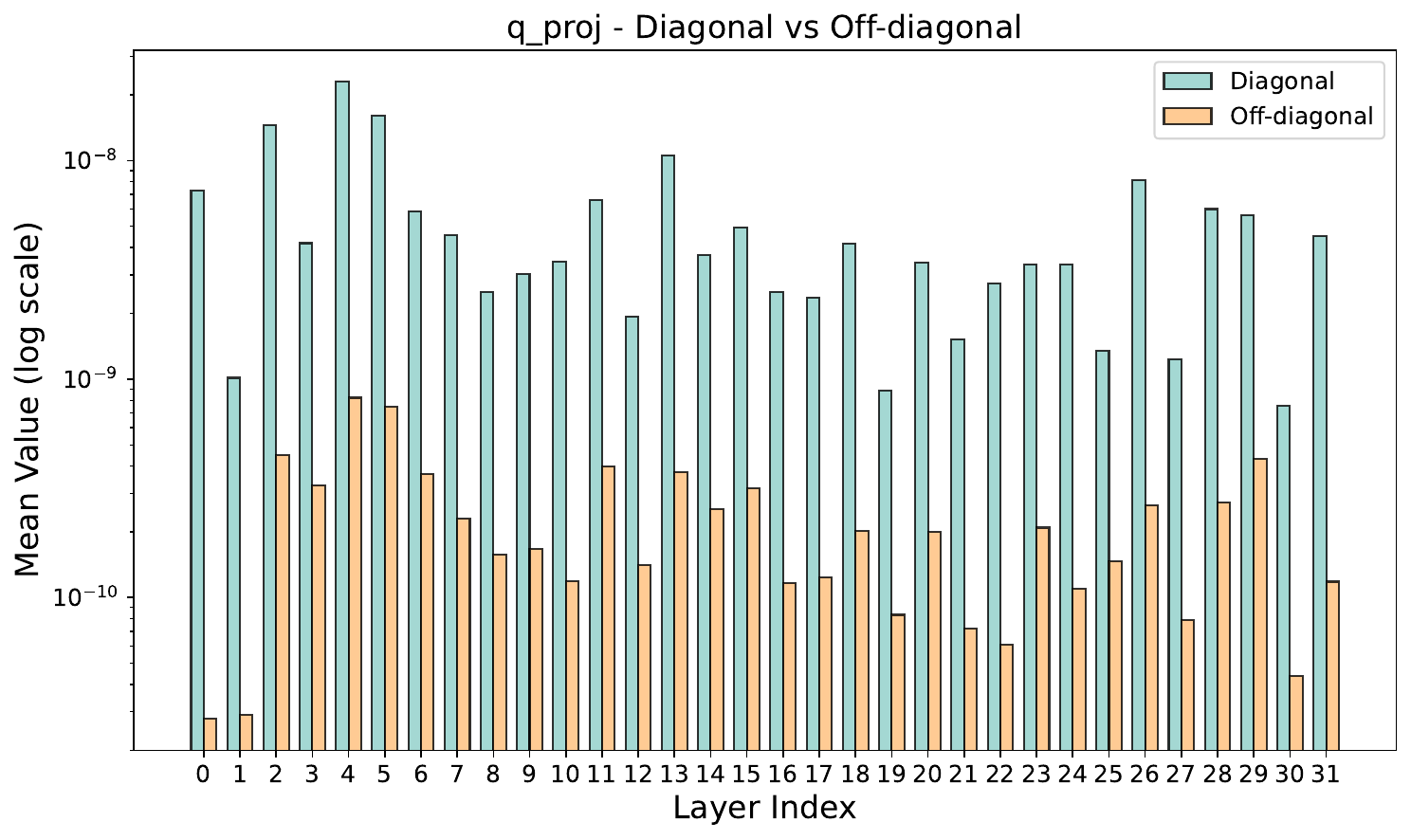}}
    \hfill
    \subfloat[K]{%
        \includegraphics[width=0.33\textwidth]{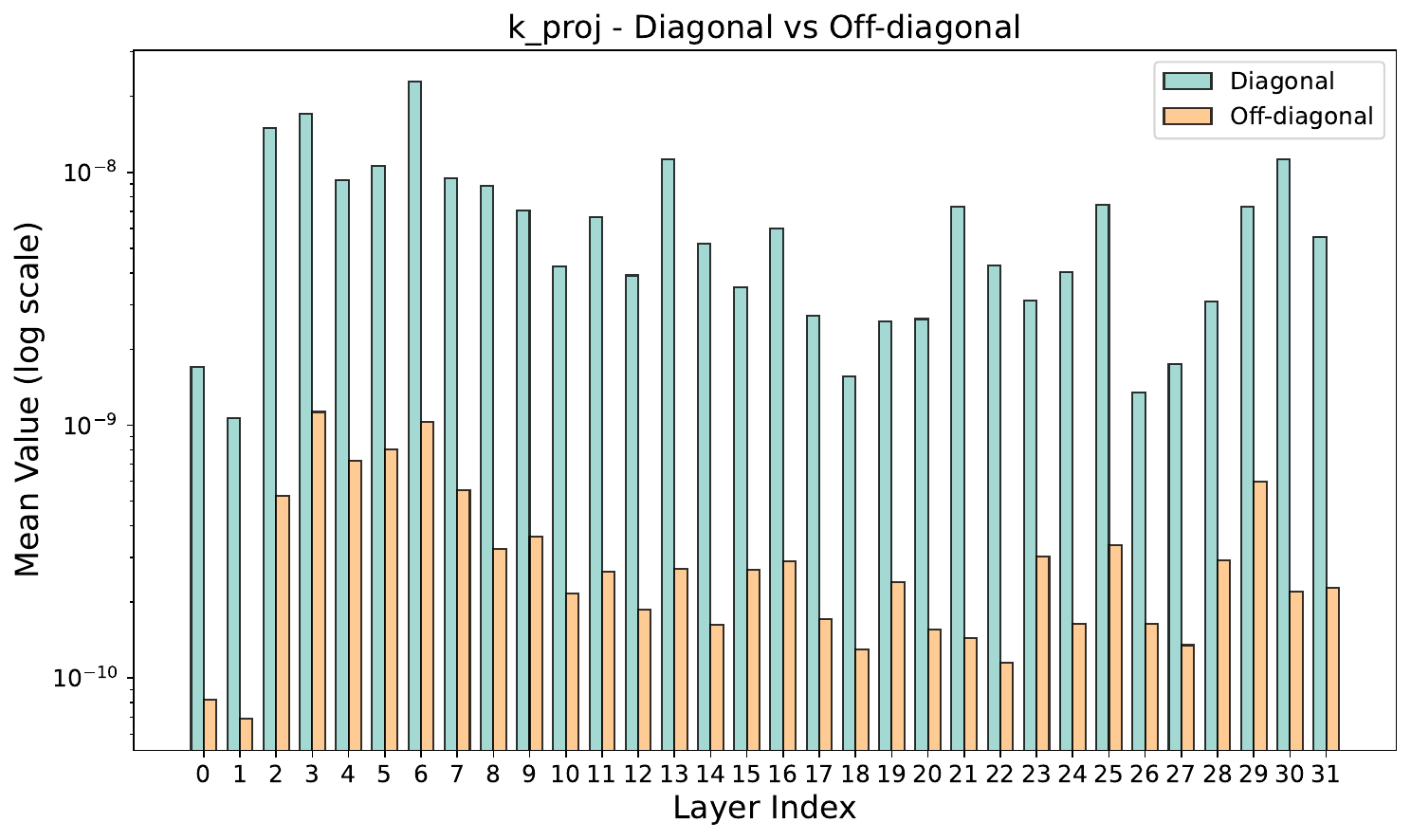}}
    \hfill
    \subfloat[V]{%
        \includegraphics[width=0.33\textwidth]{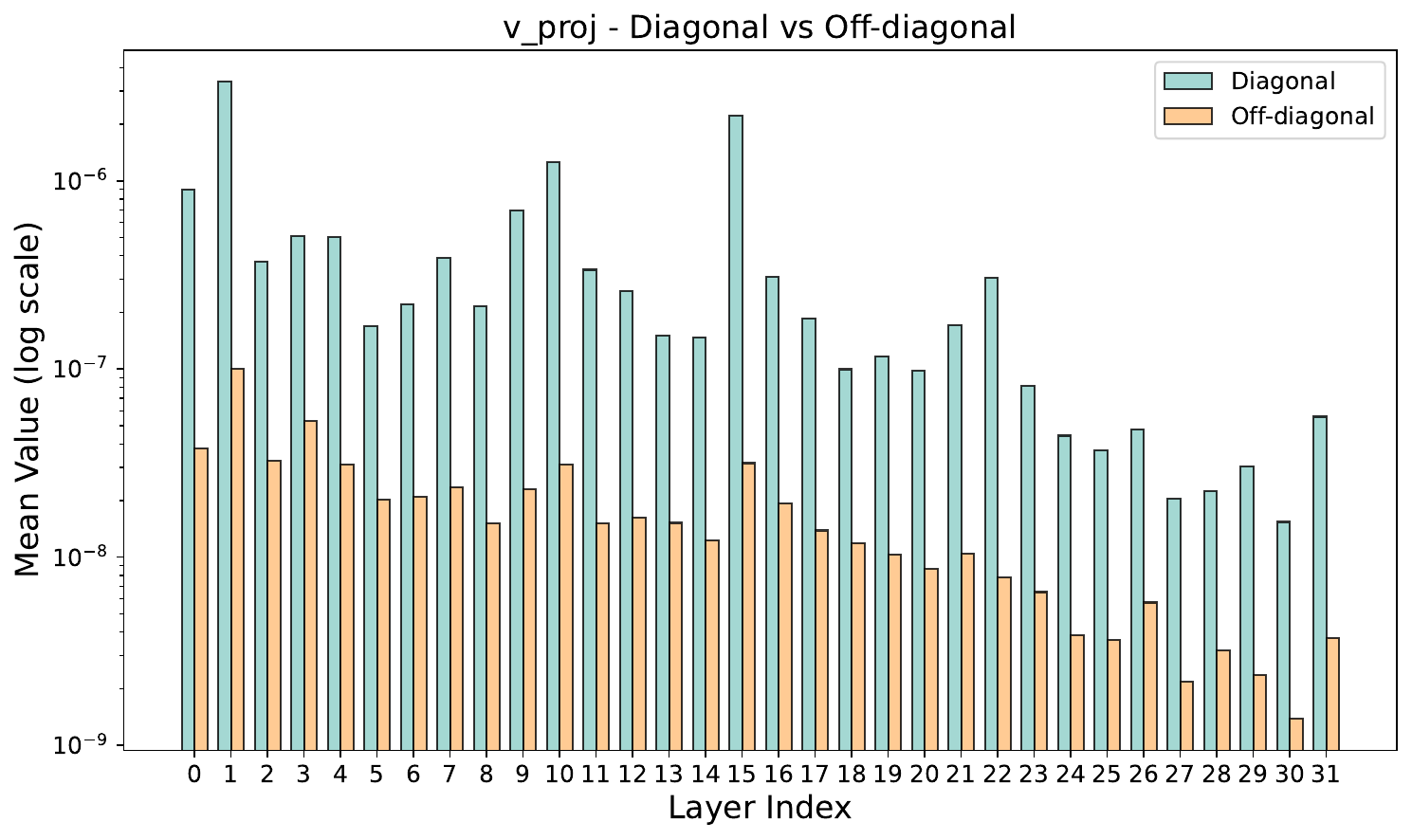}}
    
    \subfloat[O]{%
        \includegraphics[width=0.24\textwidth]{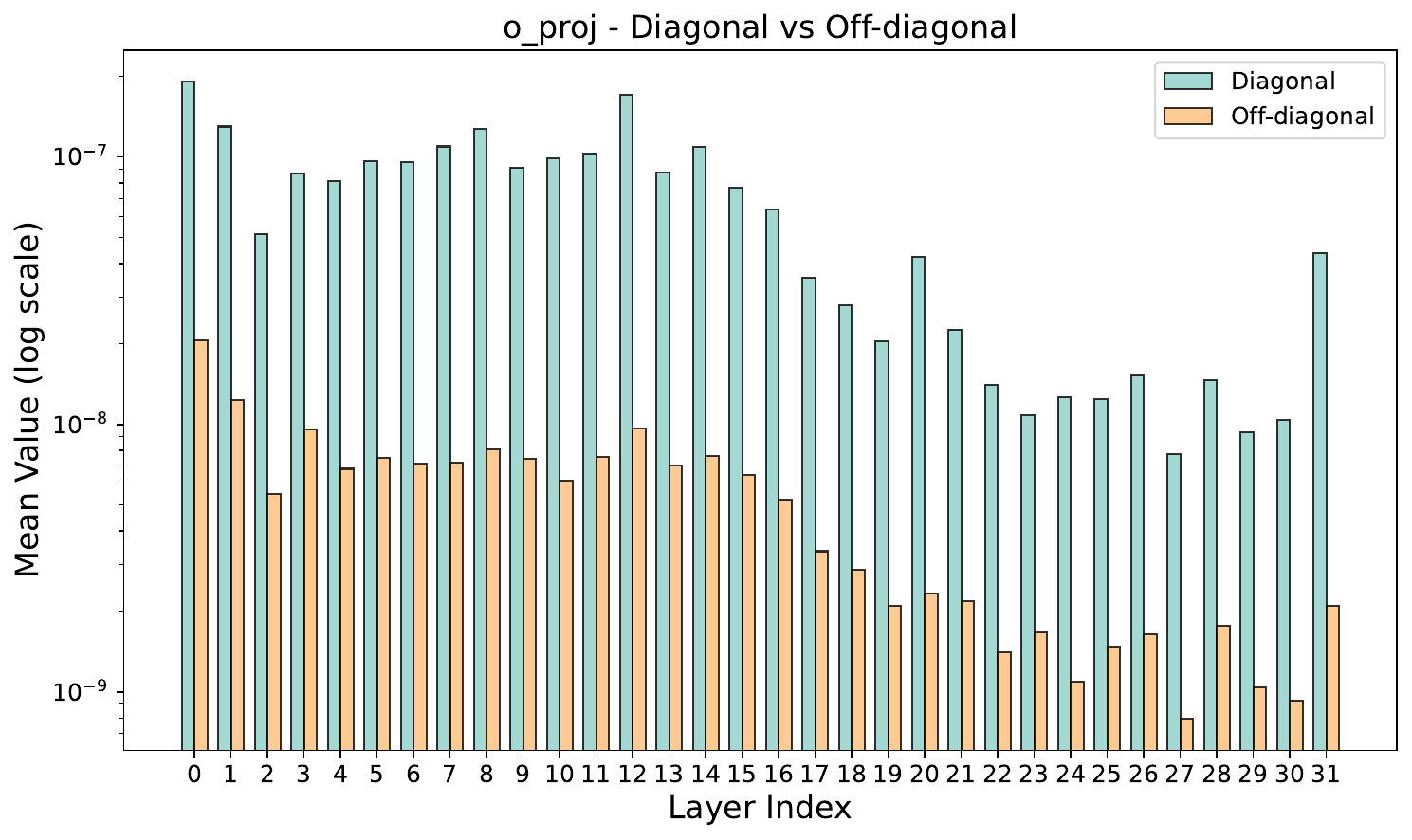}}
    \hfill
    \subfloat[Gate]{%
        \includegraphics[width=0.24\textwidth]{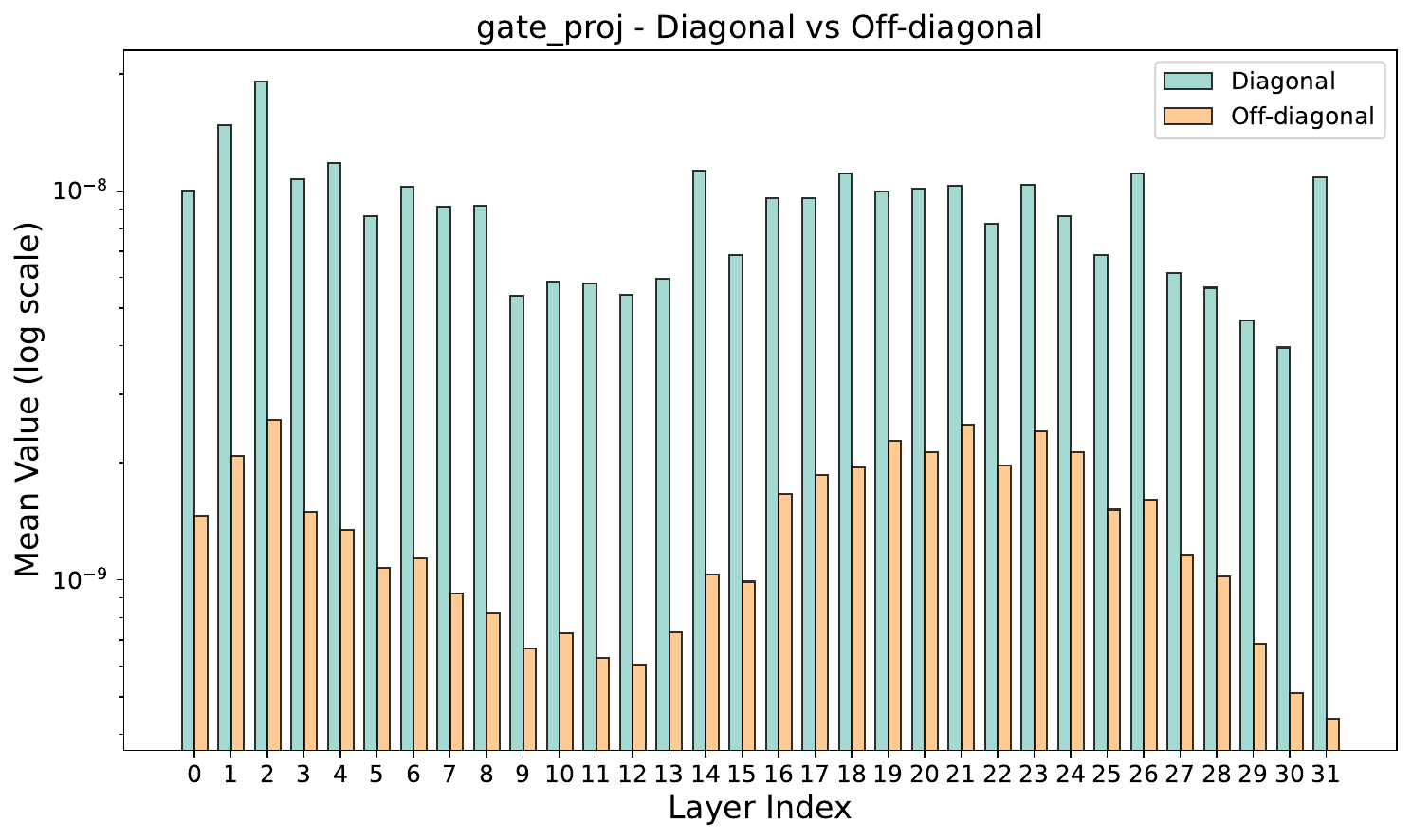}}
    \hfill
    \subfloat[Up]{%
        \includegraphics[width=0.24\textwidth]{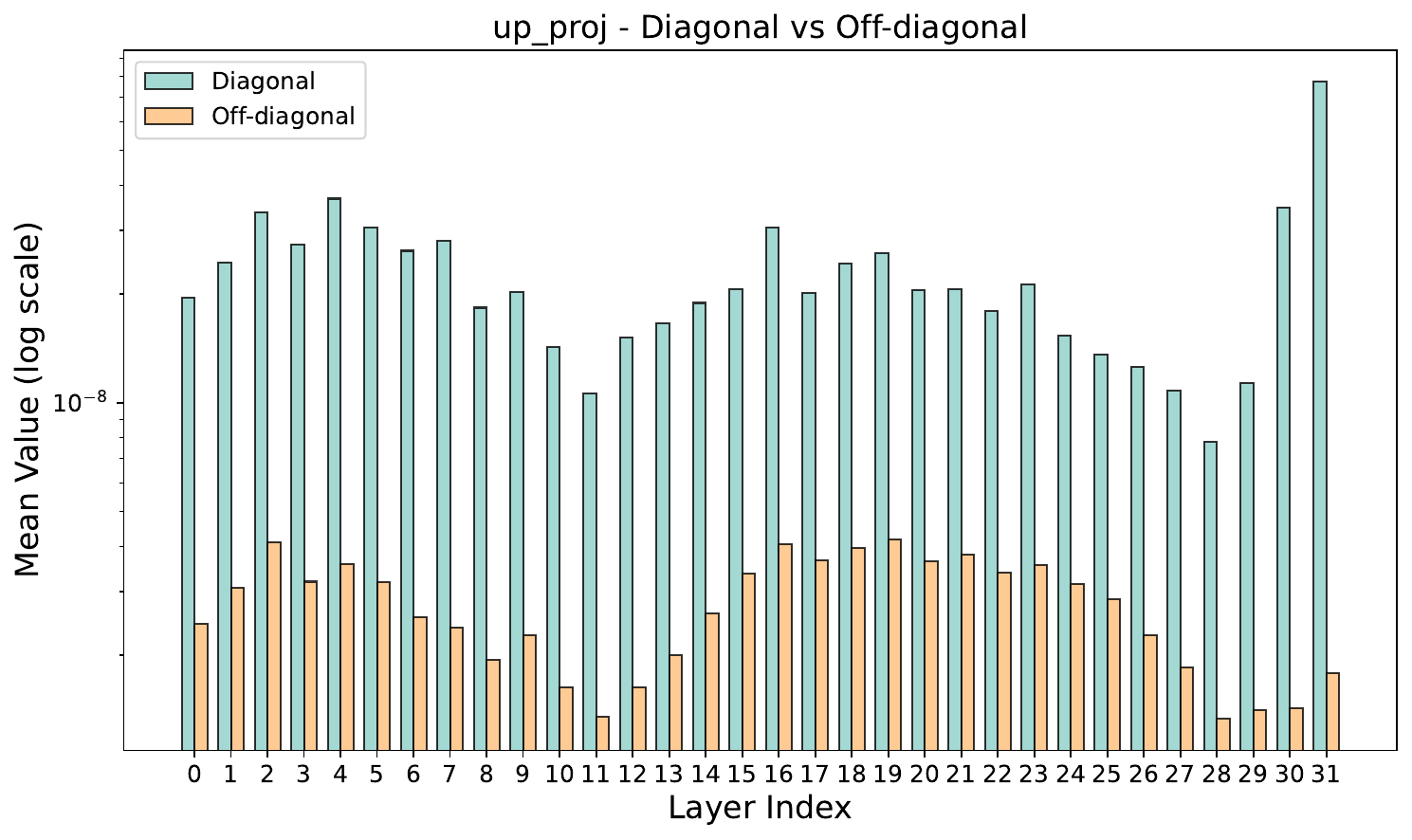}}
    \hfill
    \subfloat[Down]{%
        \includegraphics[width=0.24\textwidth]{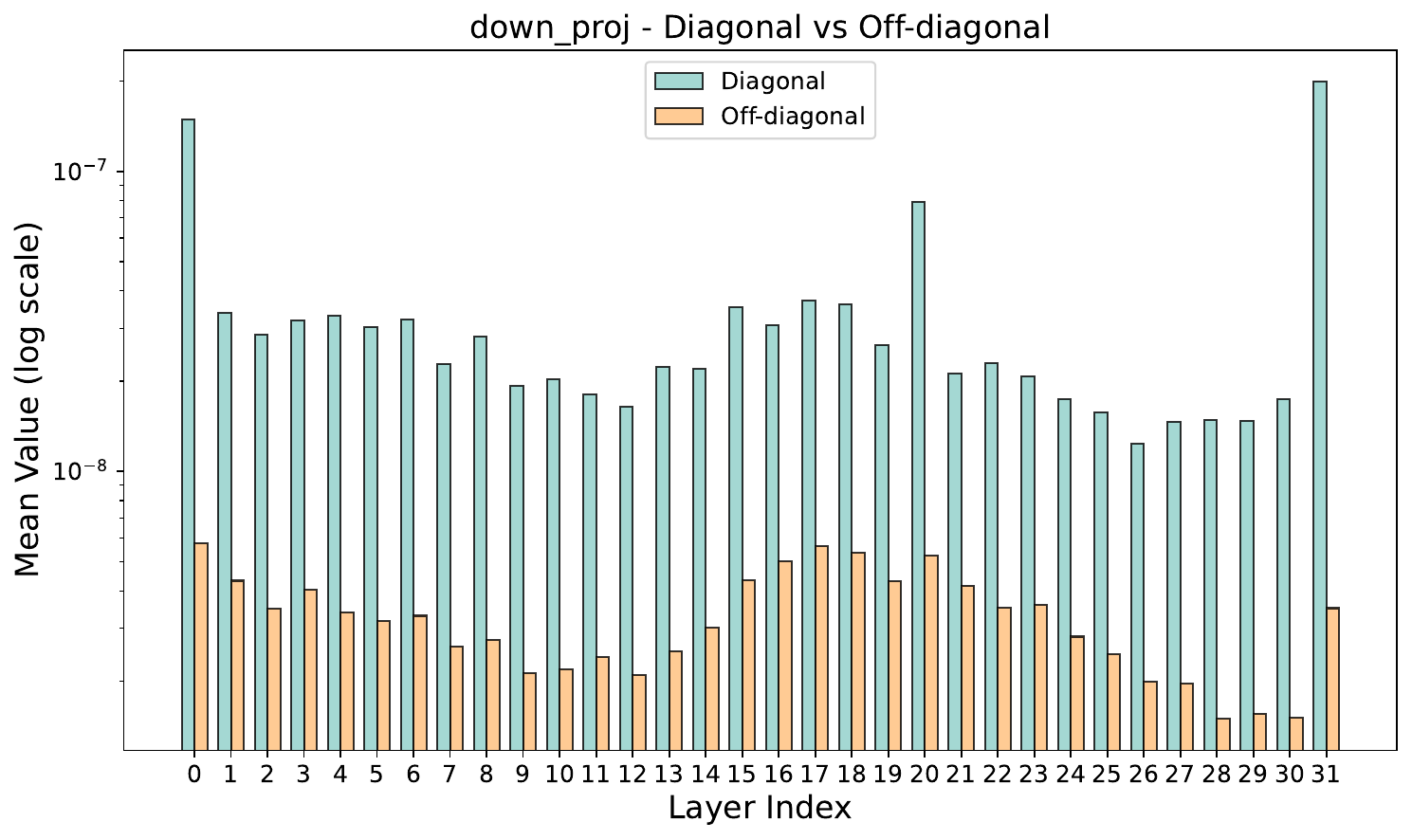}}
    
    \caption{Diagonal vs. off-diagonal elements.}
    \label{fig:hessiandiagonal}
\end{figure}

From the visualizations, we observe that the diagonal elements of the Hessian matrices are significantly larger than the off-diagonal elements. We then compute the mean absolute values of the diagonal elements and the off-diagonal elements for each layer. The results are shown in Figure \ref{fig:hessiandiagonal}.

The results demonstrate that the diagonal elements of the Hessian matrices are consistently orders of magnitude larger than the off-diagonal elements across all layers and projections. This observation justifies our approach of ignoring the off-diagonal elements in subsequent analyses, as their contributions are negligible compared to the diagonal elements.

\section{Experimental setup and additional results for Findings 1}
\label{appendix:exp}
\subsection{Models}
\label{appendix:exp_models}
We selected the most advanced open-source transformer decoder-based models and Mamba-based models for our experiments. These models encompass various scales, including different model sizes, pre-trained and instruction-tuned versions, as well as variations in model architectures. Models used in this study are as follows:

\textbf{Meta Llama.} We used Llama3-8B, Llama3-8B-Instruct, Llama3.1-8B, Llama3.1-8B-Instruct, Llama3.2-1B, Llama3.2-1B-Instruct, Llama3.2-3B, and Llama3.2-3B-Instruct.

\textbf{Qwen.} We used Qwen2-7B, Qwen2-7B-Instruct, Qwen2.5-7B, and Qwen2.5-7B-Instruct.

\textbf{DeepSeek.} We used DeepSeek-R1-Distill-Llama-8B and DeepSeek-R1-Distill-Qwen-7B.

\textbf{AI21 Jamba.} We used Jamba-1.5-Mini, an instruction-tuned Mamba-based model.

\subsection{Datasets}
\label{appendix:exp_dataset}
\textbf{Generating ToM-sensitive parameter patterns.} We constructed the ToM dataset \(\mathcal{D}_{\text{ToM-Train}}\) for identifying ToM-sensitive parameters and used the C4 dataset as \(\mathcal{D}_{\text{pre-training}}\) for identifying parameters essential for linguistic abilities. In \(\mathcal{D}_{\text{ToM-Train}}\), the Hessian was estimated using the loss at the position of the \textit{final token} in each sample. For \(\mathcal{D}_{\text{pre-training}}\), the Hessian was estimated using the loss across \textit{all token positions}.

\textbf{Evaluating ToM ability and perplexity.} We evaluated ToM abilities using the dataset \(\mathcal{D}_{\text{ToM-Test}}\) proposed by \citep{kosinski_2024_evaluating}. This dataset includes tasks covering the most critical aspects of Theory of Mind: unexpected contents and unexpected transfer tasks. Each task comprises eight different scenarios. Our previously constructed \(\mathcal{D}_{\text{ToM-Train}}\) follows the same structure as this dataset. Perplexity was evaluated using the test set of Wikitext-2. The sequence length was set to 2048.

\textbf{Evaluating contextual localization ability.} We constructed a new dataset \(\mathcal{D}_{\text{mem}}\) based on Wikitext-2 test set. This dataset contains samples of varying token lengths, from 2 to 100, randomly sampled from Wikitext-2. The task asks the model to repeat the input data, and the evaluation measures the similarity between the repeated outputs and the original samples.

\textbf{Evaluating language understanding ability.} We used the MMLU dataset to evaluate the model's understanding and reasoning capabilities. All 57 sub-tasks were included, and the evaluation was conducted using the standard 5-shot prompting method \citep{liang_2022_holistic}.

\subsection{Settings and selected  \texorpdfstring{$\kappa$}{~} value}
\label{appendix:exp_settings}
For each matrix, we set \(\kappa\) to range from \(0\) to \(5 \times 10^{-5}\) with a step size of \(2 \times 10^{-6}\). Using these \(\kappa\) values, we compute the corresponding perturbing masks \(\mathbf{m}_\kappa\) and \(\mathbf{m}_\kappa'\), and the final sensitive pattern is obtained by subtracting \(\mathbf{m}_\kappa'\) from \(\mathbf{m}_\kappa\). Among all \(\kappa\) values, the one that results in the most significant decline in ToM performance is used for reporting. The selected \(\kappa\) values are presented below.

\begin{table}[h]
\caption{Llama model mask ratio \(\kappa\)}
\label{llama ratio}
\centering
\resizebox{\textwidth}{!}{
\begin{tabular}{ccccccccc}
\toprule

\textbf{Llama} & 3-8B & 3-8B-Ins & 3.1-8B & 3.1-8B-Ins & 3.2-1B & 3.2-1B-Ins & 3.2-3B & 3.2-3B-Ins   \\
\midrule

\(\kappa\) (\(\times 10^{-5}\))
&  3.0 & 0.2 &  0.4 &  1.6 & 4.4 &  5.0 &  0.2 &  1.2 \\

\bottomrule
\end{tabular}
}
\end{table}

\begin{table}[h]
\caption{Qwen, DeepSeek, and Jamba model mask ratio \(\kappa\)}
\label{qwen ratio}
\centering
\resizebox{0.9\textwidth}{!}{
\begin{tabular}{cccccccc}
\toprule
 \multirow{2}{*}{} 
& \multicolumn{4}{c}{\textbf{Qwen}} 
& \multicolumn{2}{c}{\textbf{DeepSeek}}& \multicolumn{1}{c}{\textbf{Jamba}}\\
\cmidrule(lr){2-5}\cmidrule(lr){6-7}\cmidrule(lr){8-8}
 & 2-7B & 2-7B-Ins & 2.5-7B & 2.5-7B-Ins & R1-Llama-8B& R1-Qwen-7B&1.5-Mini\\
\midrule

\(\kappa\) (\(\times 10^{-5}\))
& 3.0 & 3.4 & 4.0 & 2.6 & 0.8 & 2.6 & 0.4\\

\bottomrule
\end{tabular}
}
\end{table}

All models are loaded in \textit{float16} precision. All experiments are repeated five times. Unless otherwise specified, the generation configuration, such as \textit{temperature} and \textit{top\_p}, follows the default settings. The only exception is the evaluation on the MMLU dataset, where \textit{temperature} is set to 0.0 for all tasks.

\subsection{Dataset examples}
\label{appendix:exp_examples}

This section provides examples from the constructed \(\mathcal{D}_{\text{ToM-Train}}\) dataset, specifically focusing on \textit{Unexpected Transfer} tasks. We highlight the differences between the \textit{Original false-belief} contexts and the \textit{True-belief Control} conditions, and provide the correct outputs for each prompt. The structure of the designed \(\mathcal{D}_{\text{ToM-Train}}\) dataset is consistent with \(\mathcal{D}_{\text{ToM-Test}}\) \citep{kosinski_2024_evaluating}. For more tasks, such as \textit{Unexpected Contents} tasks, please refer to \citep{kosinski_2024_evaluating}.

We also introduce the \textit{contextual localization task}, a custom-built dataset intended to gauge the model’s capability to precisely reconstruct input sequences. This task evaluates contextual localization by quantifying the alignment between the provided input and the generated output.

\subsubsection{Unexpected transfer task}
\label{exp:appendix_unexpectedtransfer}
\textit{Original false-belief context (FB): James puts his car keys in the drawer before heading out to exercise. While James is out, his wife Linda decides to clean the house. She finds the car keys in the drawer and thinks they would be safer in the key cabinet. She moves them there and continues cleaning. James comes back from his run and wants to get his car keys. }
\begin{itemize}
    \item \textit{Prompt 1: The keys will be taken out of the \uline{\textcolor{blue}{\textit{key cabinet}}}}.
    \item \textit{Prompt 2: James will look for the keys in the \uline{\textcolor{blue}{\textit{drawer}}}}.
\end{itemize}

\textit{Present protagonist true-belief control context (PP): James puts his car keys in the drawer before heading out to exercise. \textcolor{blue}{Before} James is out, his wife Linda decides to clean the house. She finds the car keys in the drawer and thinks they would be safer in the key cabinet. \textcolor{blue}{James sees Linda move the keys to the key cabinet.} James comes back from his run and wants to get his car keys.}

\begin{itemize}
    \item \textit{Prompt 1: The keys will be taken out of the \uline{\textcolor{blue}{\textit{key cabinet}}}}.
    \item \textit{Prompt 2: James will look for the keys in the \uline{\textcolor{blue}{\textit{key cabinet}}}}.
\end{itemize}

\textit{Informed protagonist true-belief control context(IP): James puts his car keys in the drawer before heading out to exercise. While James is out, his wife Linda decides to clean the house. She finds the car keys in the drawer and thinks they would be safer in the key cabinet. She moves them there and continues cleaning. James comes back from his run and wants to get his car keys. \textcolor{blue}{Linda calls James and tells him she moved the keys from the drawer to the key cabinet. James believes her}.}

\begin{itemize}
    \item \textit{Prompt 1: The keys will be taken out of the \uline{\textcolor{blue}{\textit{key cabinet}}}}.
    \item \textit{Prompt 2: James will look for the keys in the \uline{\textcolor{blue}{\textit{key cabinet}}}}.
\end{itemize}

\textit{No transfer true-belief control context (NT): Complete the following story: James puts his car keys in the drawer before heading out to exercise. While James is out, his wife Linda decides to clean the house. \textcolor{blue}{She finds the car keys in the drawer but leaves them there and continues cleaning}. James comes back from his run and wants to get his car keys. }

\begin{itemize}
    \item \textit{Prompt 1: The keys will be taken out of the \uline{\textcolor{blue}{\textit{drawer}}}}.
    \item \textit{Prompt 2: James will look for the keys in the \uline{\textcolor{blue}{\textit{drawer}}}}.
\end{itemize}

\subsubsection{Contextual Localization task}
\label{appendix:exp_contextual}
\paragraph{Prompt Construction.} We provide the model with an input text and explicitly instruct it to repeat the text verbatim. For example, given an input text \textit{\textless Input\textgreater}, the prompt is constructed as follows: 

``Please repeat every single word of the following text: \textit{\textless Input\textgreater} Repeat every single word of the text:"

This prompt format ensures that the model is explicitly guided to reproduce the input text without modification.

\paragraph{Evaluation Method.}
To evaluate the performance, we measure the similarity between the generated text and the original input. Let \( \mathbf{X} = [x_1, x_2, \dots, x_n] \) denote the tokenized input and \( \mathbf{Y} = [y_1, y_2, \dots, y_m] \) denote the tokenized generated output. The similarity score \( S \) is computed as:
\[
S = \frac{|\{x_i \mid x_i \in \mathbf{Y}\}|}{n},
\]

where \( |\{x_i \mid x_i \in \mathbf{Y}\}| \) represents the number of tokens in \( \mathbf{X} \) that are present in \( \mathbf{Y} \), \( n \) is the total token number in \( \mathbf{X} \). Then we have:
\[
\text{Average Similarity} = \frac{1}{N} \sum_{i=1}^N S_i,
\]

where \( N \) is the number of samples, and \( S_i \) is the similarity score for the \( i \)-th sample. In our experiments, we set \( N = 100 \) and evaluate the performance on input sequences of varying lengths, specifically \( n \in \{2, 4, 6, 8, 10, 20, 30, 40, \dots, 100\} \). 

\subsection{ToM task additional results for RoPE-based models}
\begin{figure}[h]
    \centering
    \subfloat[Llama3-8B]{%
        \includegraphics[width=0.24\textwidth]{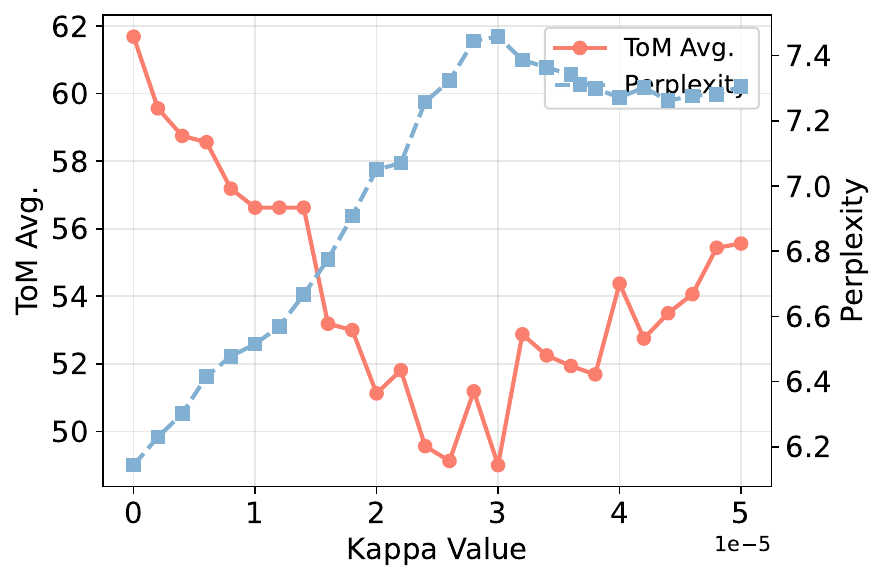}}
    \hfill
    \subfloat[Llama3-8B-Instruct]{%
        \includegraphics[width=0.24\textwidth]{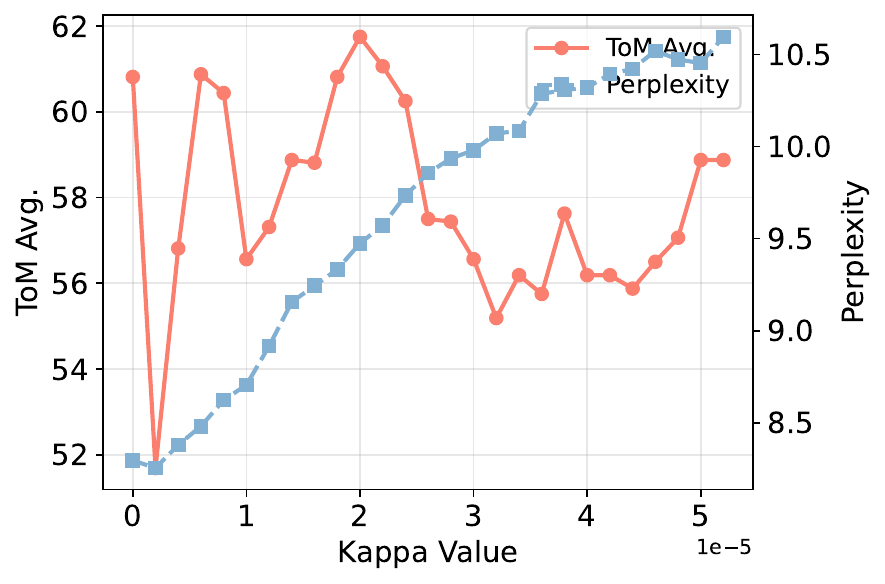}}
    \hfill
    \subfloat[Llama3.1-8B]{%
        \includegraphics[width=0.24\textwidth]{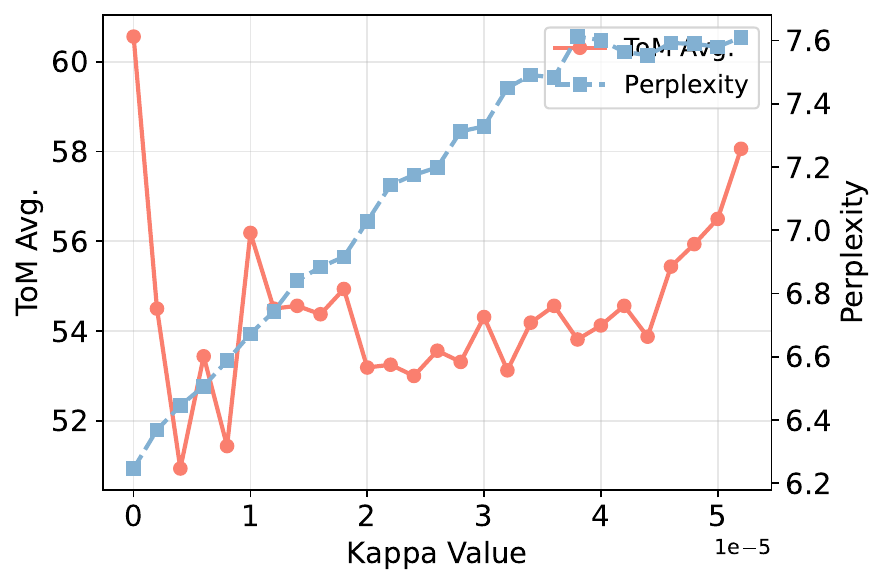}}
    \hfill
    \subfloat[Llama3.1-8B-Instruct]{%
        \includegraphics[width=0.24\textwidth]{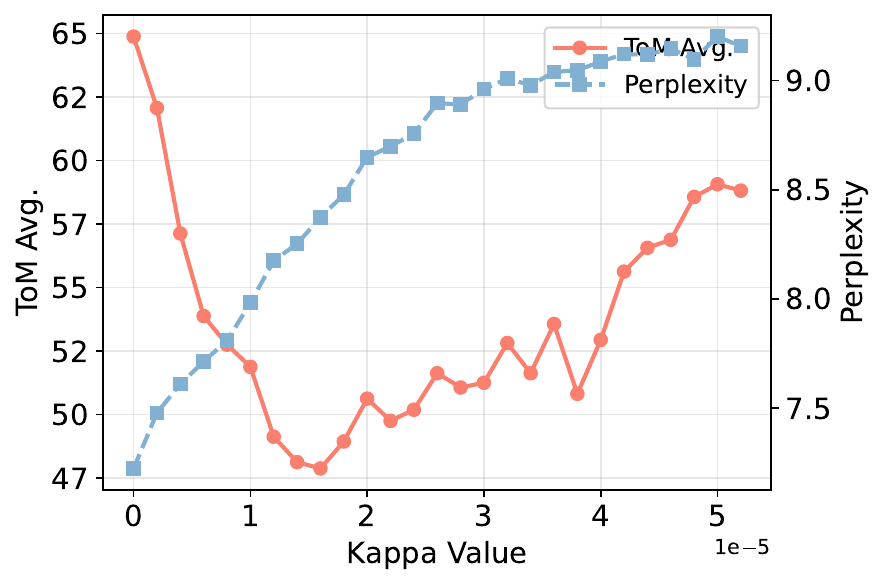}}

    \subfloat[Llama3.2-1B]{%
        \includegraphics[width=0.24\textwidth]{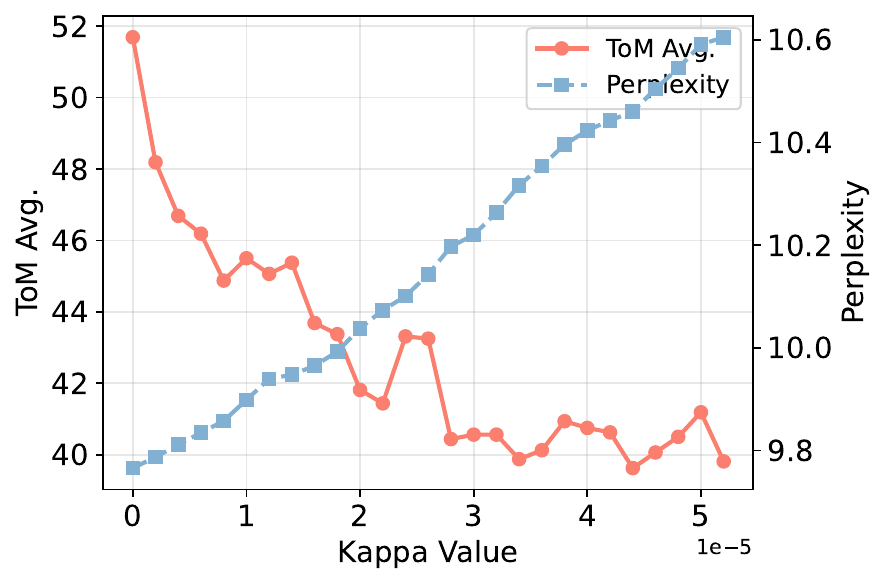}}
    \hfill
    \subfloat[Llama3.2-1B-Instruct]{%
        \includegraphics[width=0.24\textwidth]{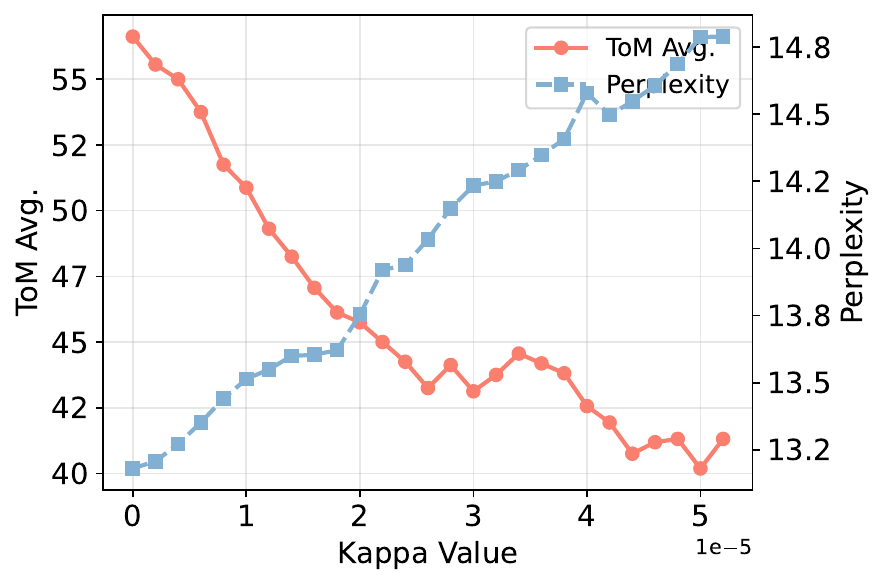}}
    \hfill
    \subfloat[Llama3.2-3B]{%
        \includegraphics[width=0.24\textwidth]{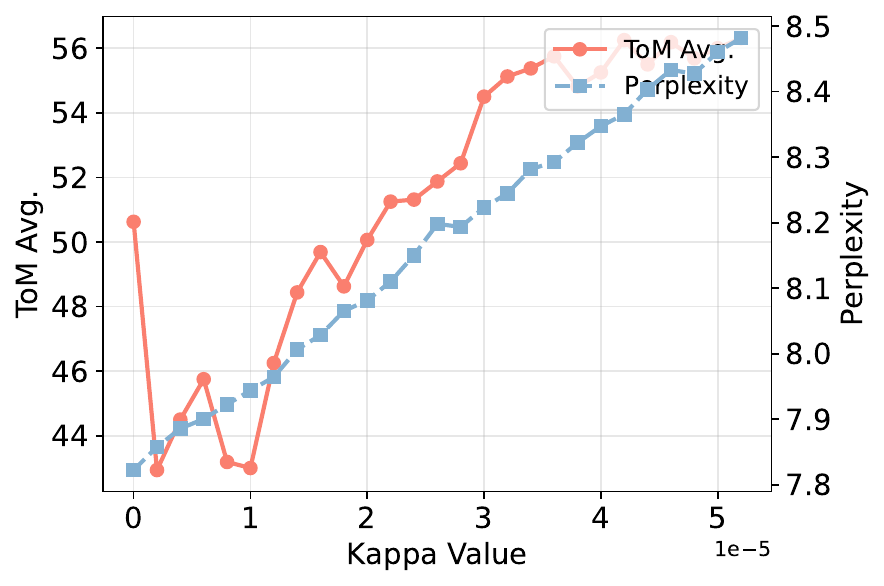}}
    \hfill
    \subfloat[Llama3.2-3B-Instruct]{%
        \includegraphics[width=0.24\textwidth]{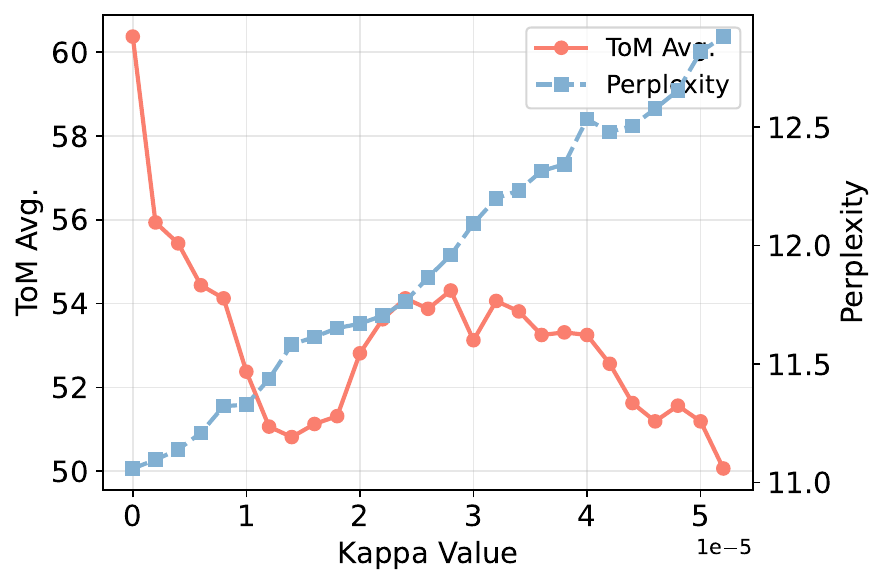}}

    \subfloat[Qwen2-7B]{%
        \includegraphics[width=0.24\textwidth]{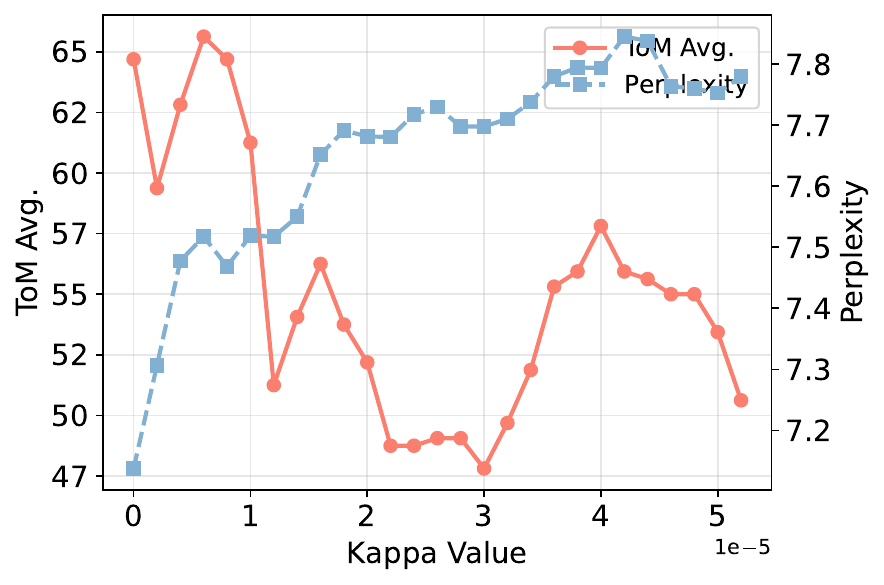}}
    \hfill
    \subfloat[Qwen2-7B-Instruct]{%
        \includegraphics[width=0.24\textwidth]{qwen2-7B_ToM_vs_Perplexity.pdf}}
    \hfill
    \subfloat[Qwen2.5-7B]{%
        \includegraphics[width=0.24\textwidth]{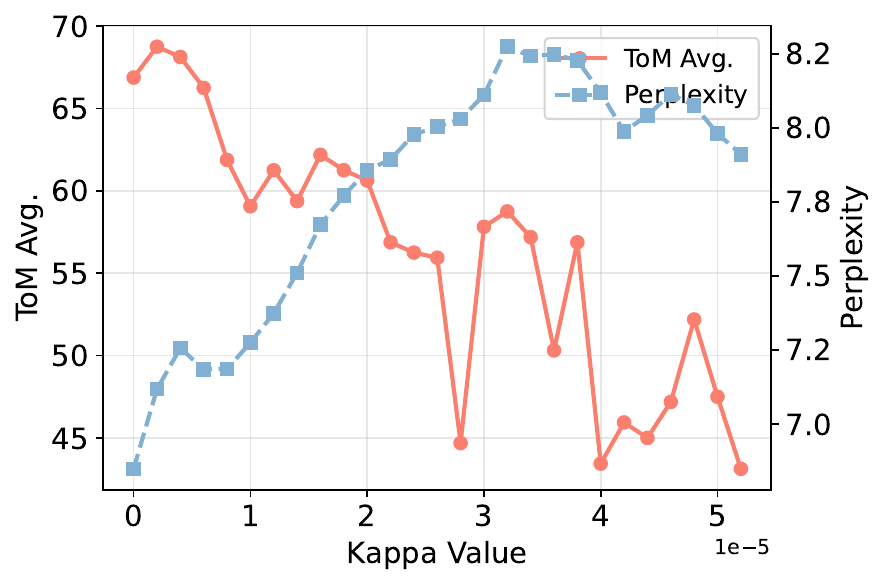}}
    \hfill
    \subfloat[Qwen2.5-7B-Instruct]{%
        \includegraphics[width=0.24\textwidth]{qwen2.5-7B_ToM_vs_Perplexity.pdf}}

    \subfloat[DeepSeek-Llama-8B]{%
        \includegraphics[width=0.24\textwidth]{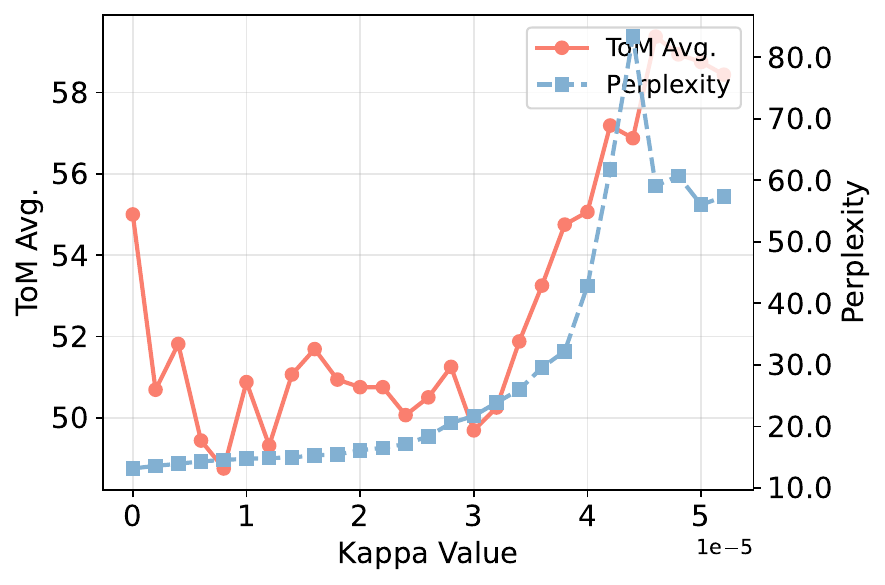}}
        \hspace{3pt}
    \subfloat[DeepSeek-Qwen-7B]{%
     \includegraphics[width=0.24\textwidth]{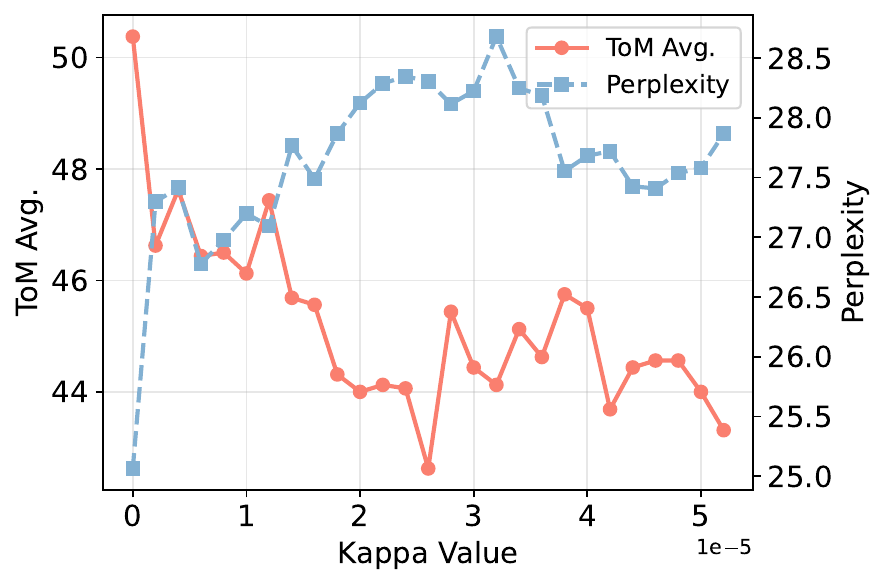}}
        \hspace{3pt}
    \subfloat[Llama3-8B-Random]{%
     \includegraphics[width=0.24\textwidth]{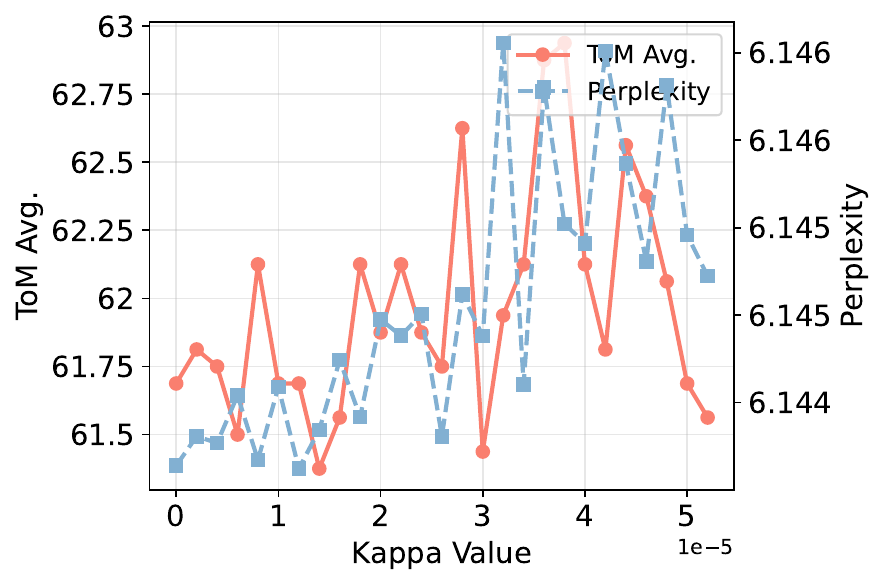}}
    \hfill
    \caption{Average ToM performance and perplexity of RoPE-based models across different values of \(\kappa\).}
    \label{fig:kapparatio}
\end{figure}

\paragraph{Searching for best \(\kappa\).}
We conduct a scan over \(\kappa\), setting \(\kappa\) to range from \(2 \times 10^{-6}\) to \(5 \times 10^{-5}\). The average ToM performance and perplexity of RoPE-based models across different values of \(\kappa\) are shown in \figurename~\ref{fig:kapparatio}. We consistently observe that within this extremely small range, a sensitive parameter pattern can be identified that significantly reduces ToM performance. In contrast, the increase in perplexity remains marginal.

\paragraph{Random perturbation of parameters does not affect model performance.}
In Figure~\ref{fig:kapparatio}(o), we report the ToM performance and perplexity results when randomly perturbing parameters with the same \(\kappa\) values. We observe that, compared to the ToM-sensitive parameter patterns, random perturbations have virtually no effect on either ToM ability or perplexity. This demonstrates that the models are indeed specifically sensitive to the structured patterns we identified.

\label{appendix:exp_tom}

\subsection{Contextual localization task additional results for RoPE-based models}
\begin{figure}[t]
    \centering
    \subfloat[Llama3-8B-Instruct]{%
        \includegraphics[width=0.24\textwidth]{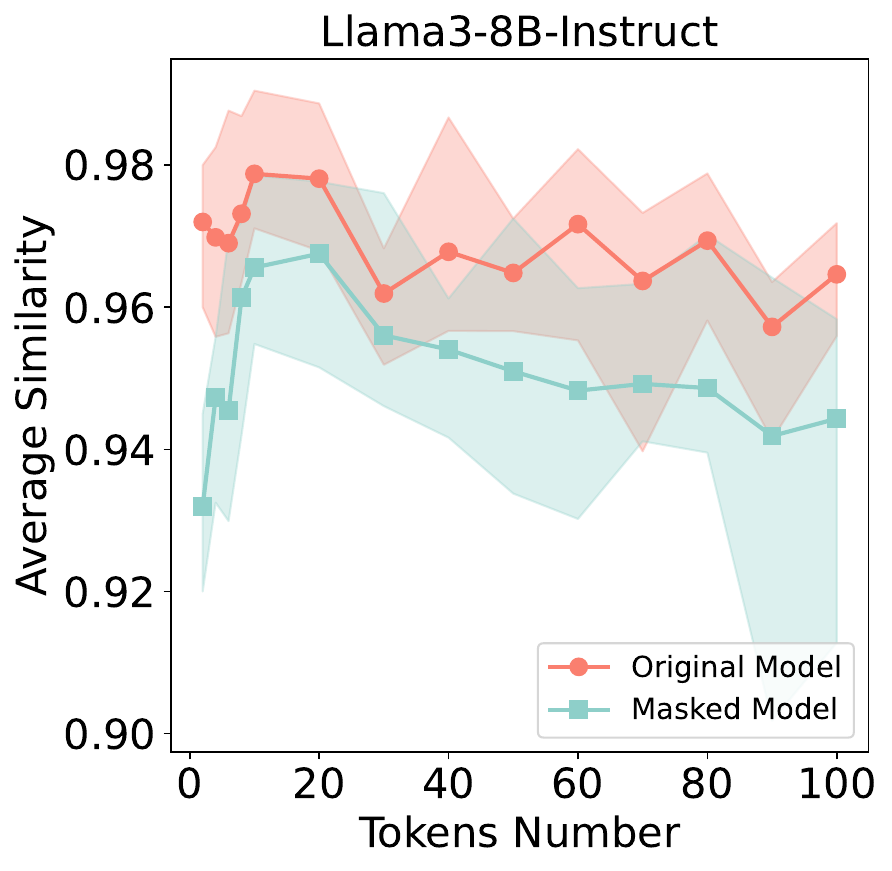}}
    \hfill
    \subfloat[Llama3.1-8B]{%
        \includegraphics[width=0.24\textwidth]{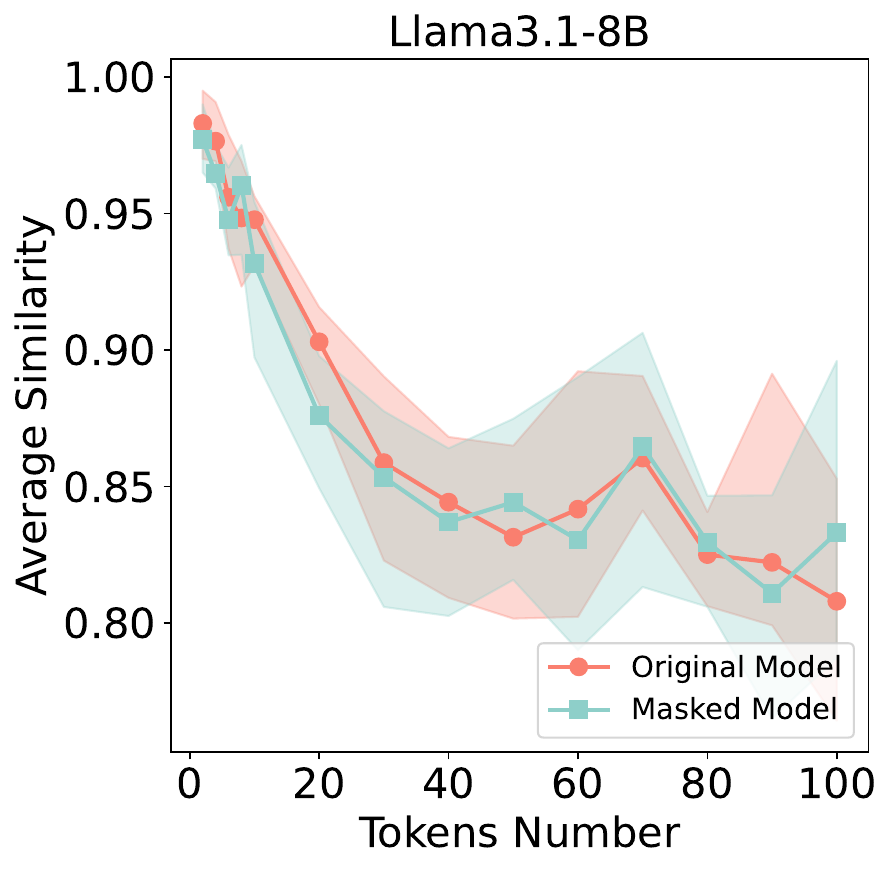}}
    \hfill
    \subfloat[Llama3.2-1B-Instruct]{%
        \includegraphics[width=0.24\textwidth]{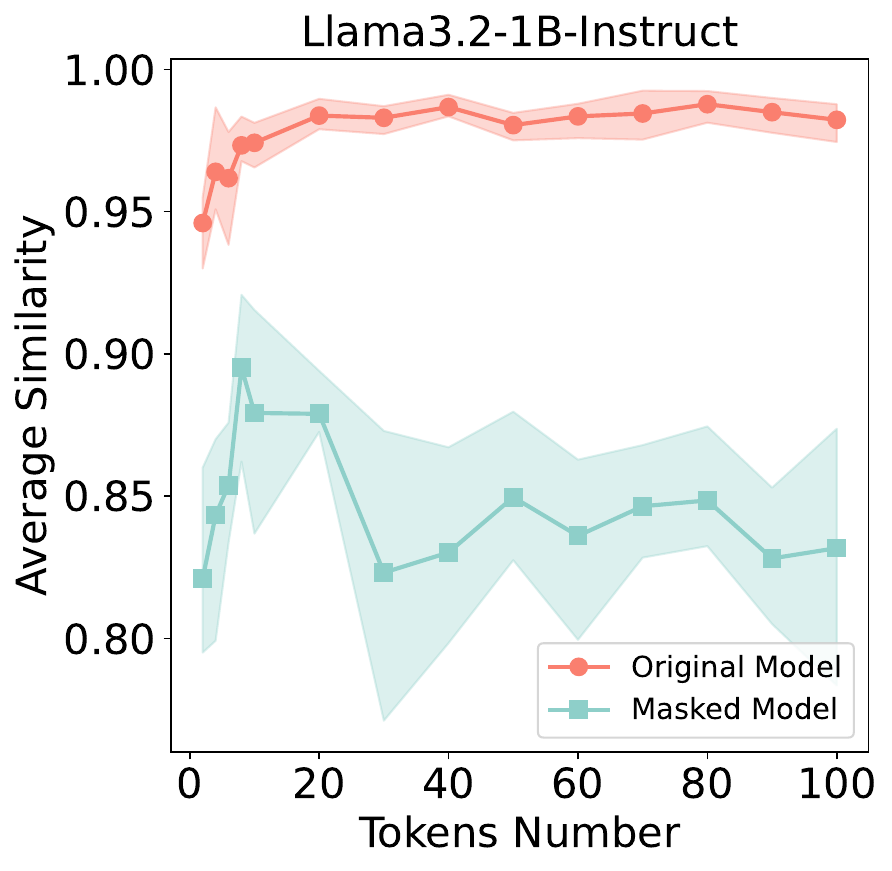}}
    \hfill
    \subfloat[Llama3.2-3B]{%
        \includegraphics[width=0.24\textwidth]{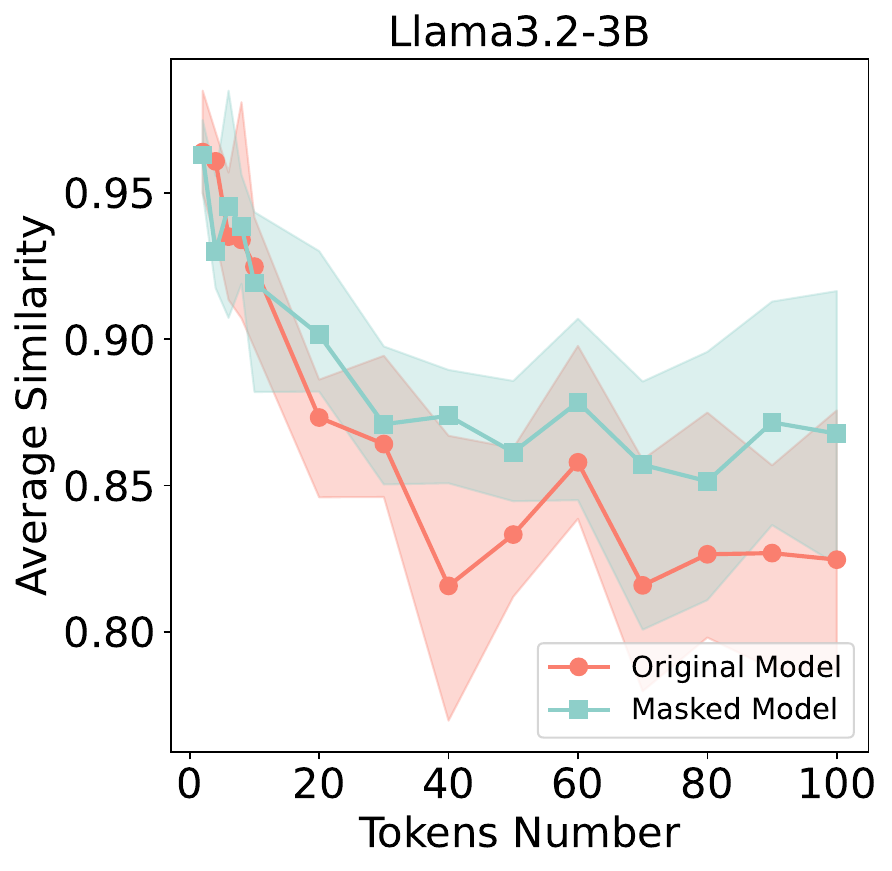}}

    \subfloat[Llama3.2-3B-Instruct]{%
        \includegraphics[width=0.24\textwidth]{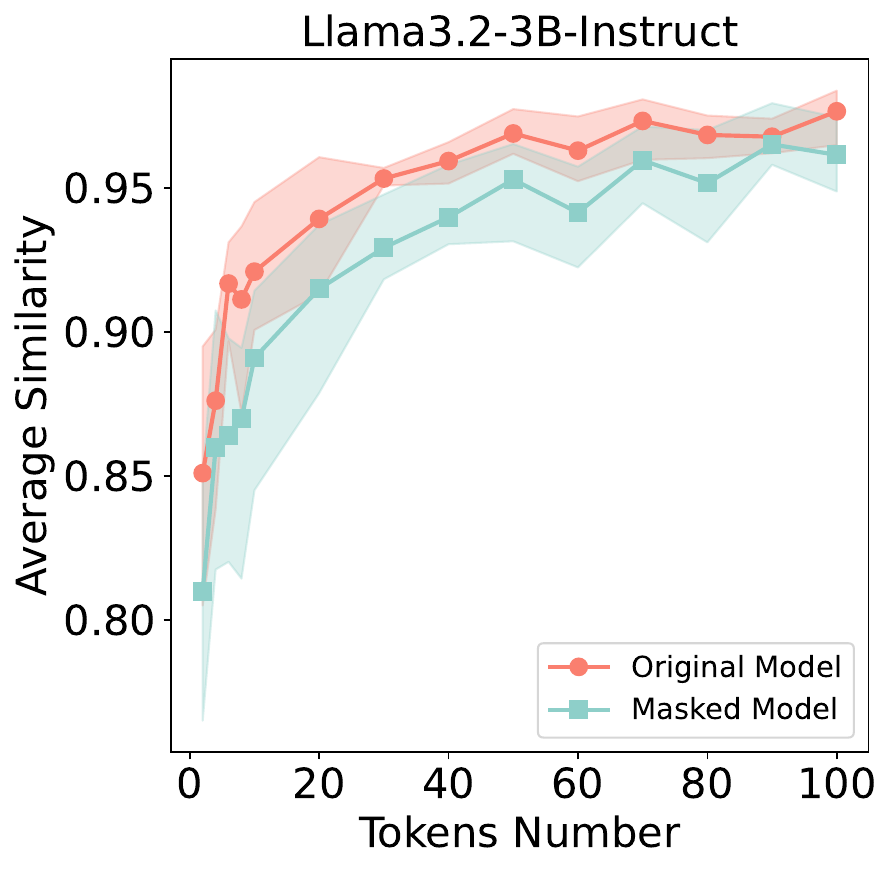}}
    \hfill
    \subfloat[Qwen2-7B]{%
        \includegraphics[width=0.234\textwidth]{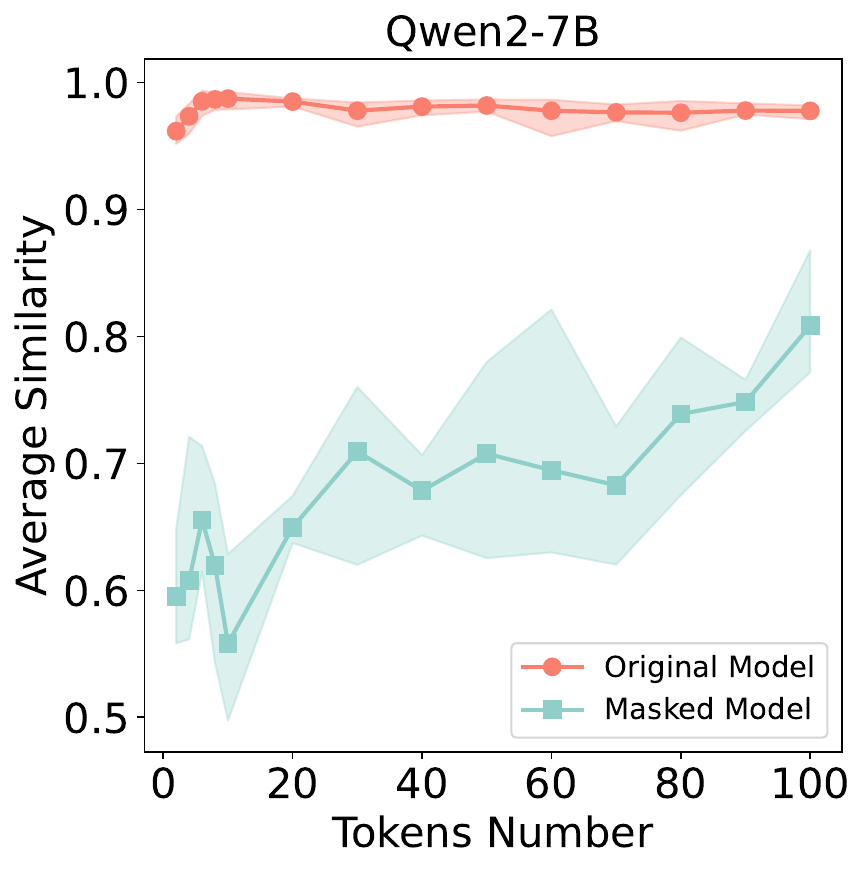}}
    \hfill
    \subfloat[Qwen2-7B-Instruct]{%
        \includegraphics[width=0.24\textwidth]{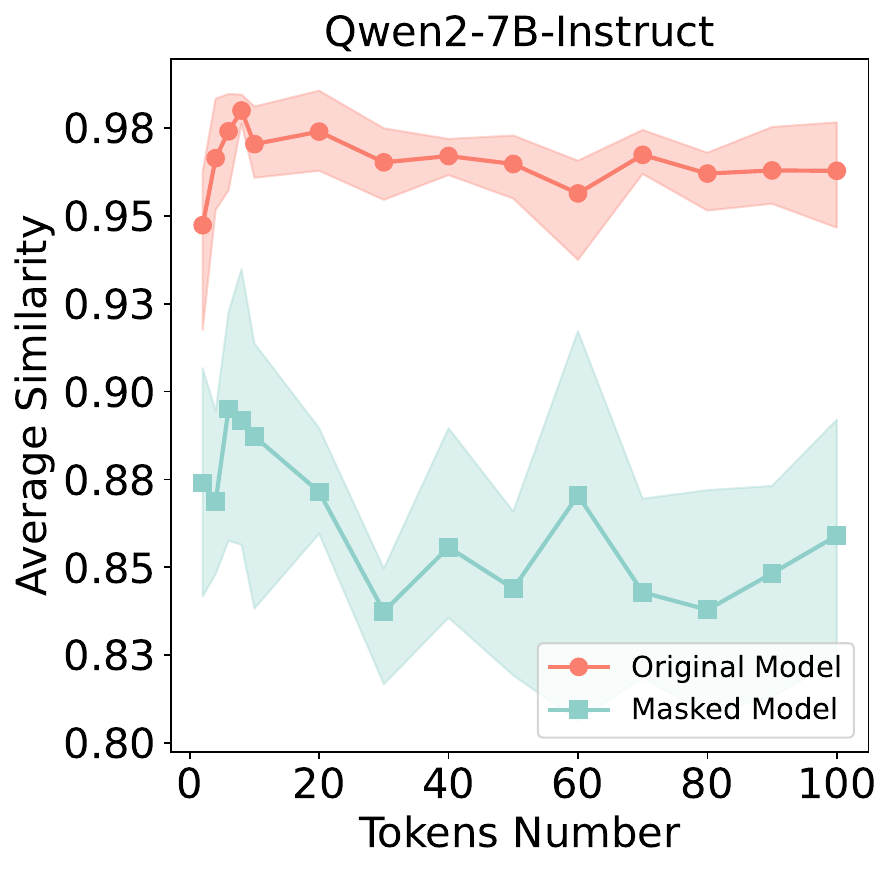}}
    \hfill
    \subfloat[Qwen2.5-7B]{%
        \includegraphics[width=0.24\textwidth]{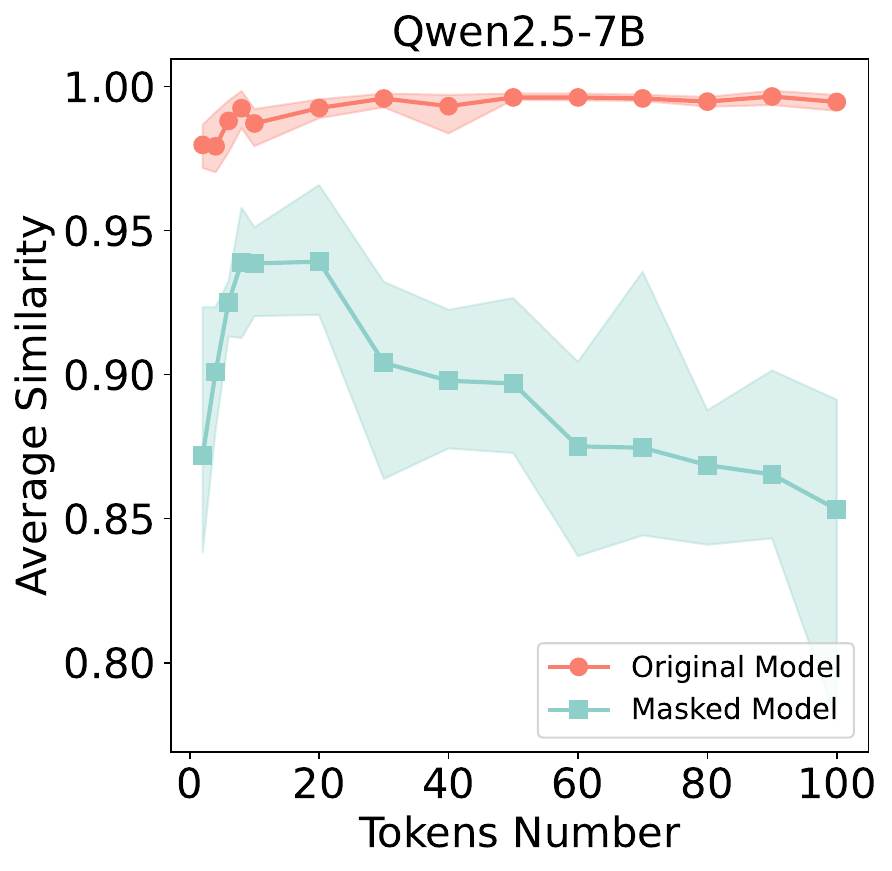}}

    \subfloat[DeepSeek-Llama-8B]{%
        \includegraphics[width=0.24\textwidth]{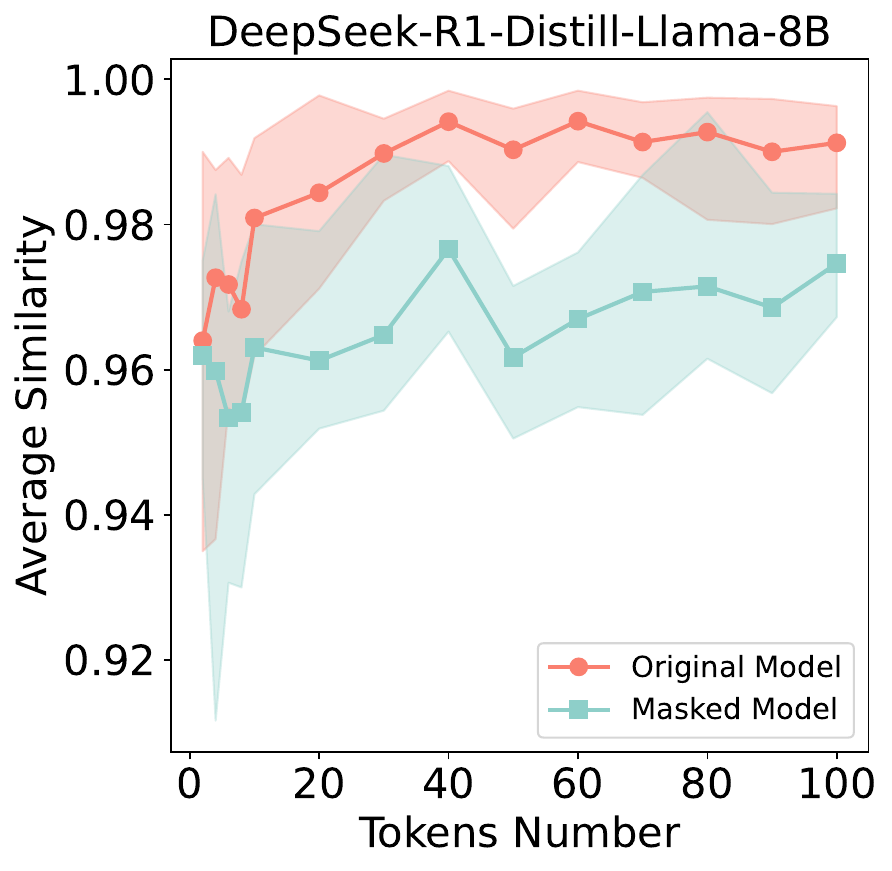}}
    \hspace{1.5pt}
    \subfloat[DeepSeek-Qwen-7B]{%
        \includegraphics[width=0.24\textwidth]{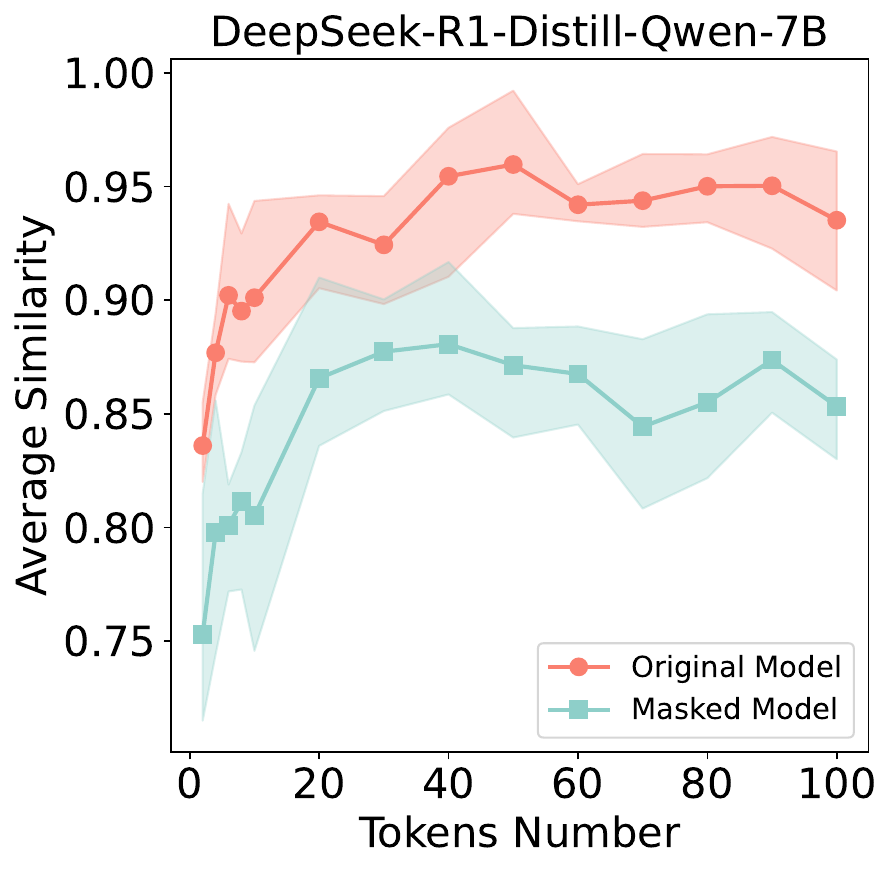}}
    \hfill
    \hfill

    \caption{Additional contextual localization ability evaluation across RoPE-based models.}
    \label{fig:comparison}
\end{figure}

\paragraph{RoPE-based models demonstrate consistent contextual localization capabilities.} As shown in Figure \ref{fig:comparison}, as the number of tokens to be repeated increases, most model performance either remains relatively stable or gradually declines. This trend aligns with our intuition: when a model exhibits poor contextual localization ability, it is likely to ``forget'' recently encountered tokens almost immediately, leading to diminished performance as the token count grows.

\paragraph{Sparse ToM sensitive parameter patterns influence the contextual localization capabilities.} In most cases, RoPE-based models with perturbed parameters exhibit significantly poorer positioning performance, especially when the repeated token length is large. Furthermore, masked models display higher output variance when experiments are repeated multiple times.
\label{appendix:exp_contextualresults}

\subsection{language understanding task additional results for RoPE-based models}

\label{appendix:exp_mmlu}

\begin{figure}[H]
    \centering
    \subfloat[Llama3-8B]{%
        \includegraphics[width=0.45\textwidth]{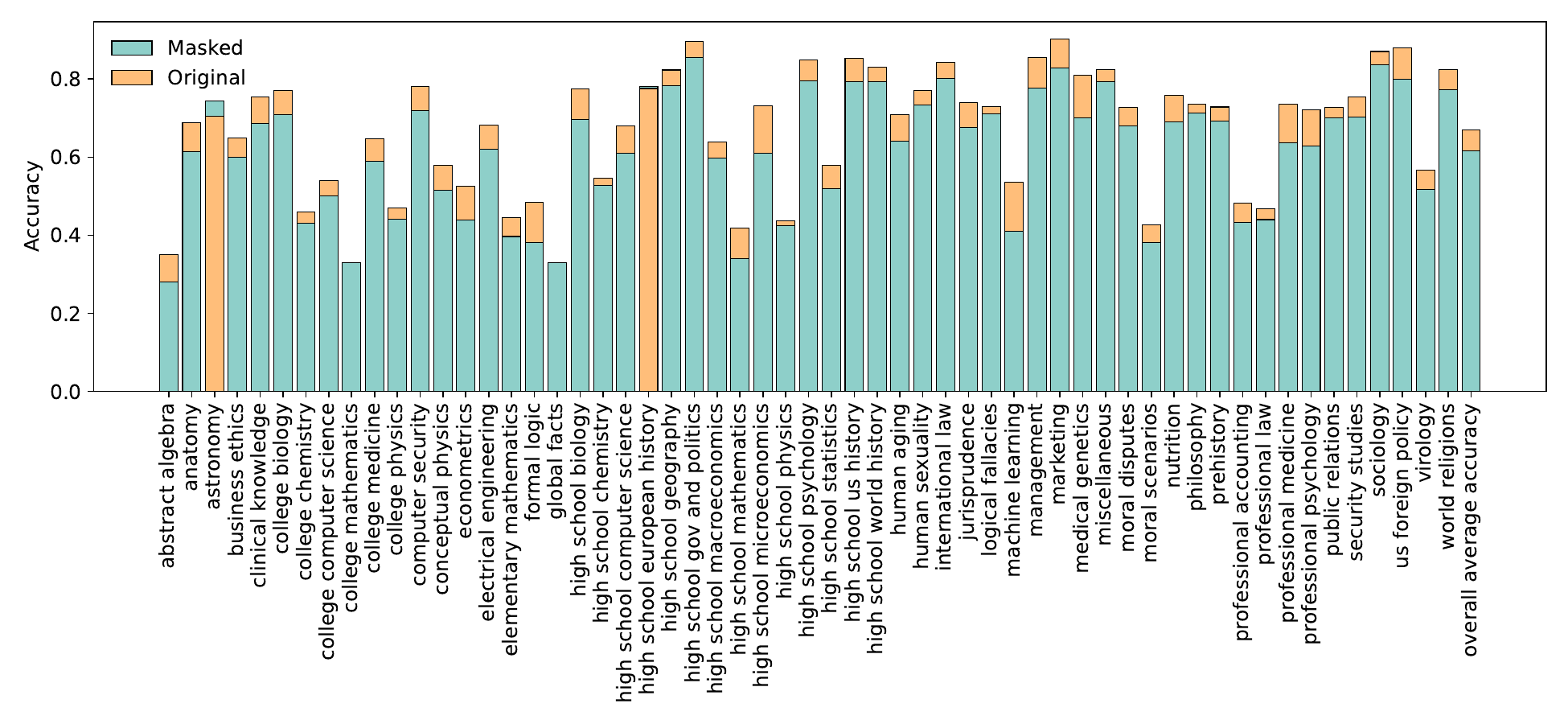}}
    \hfill
    \subfloat[Llama3-8B-Instruct]{%
        \includegraphics[width=0.455\textwidth]{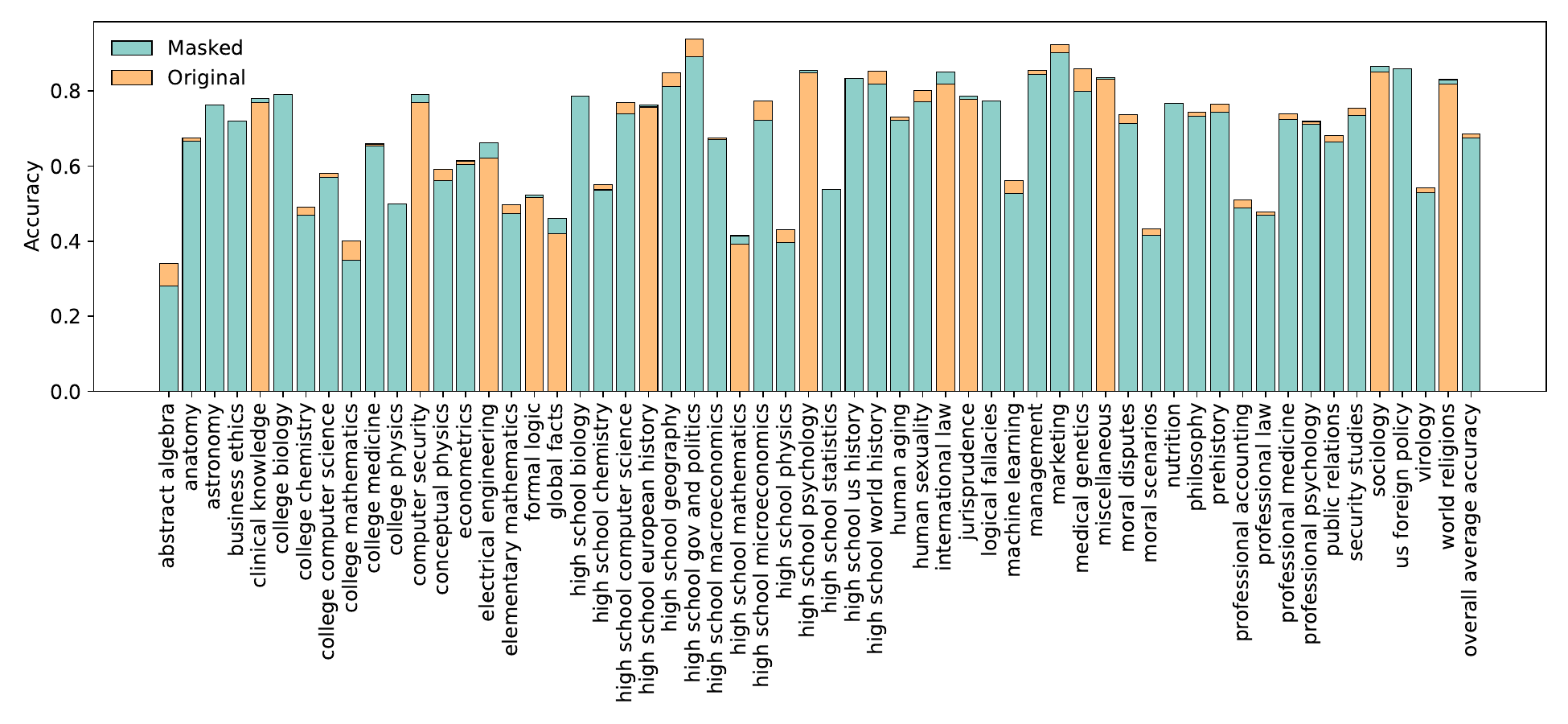}}

    \subfloat[Llama3.1-8B]{%
        \includegraphics[width=0.45\textwidth]{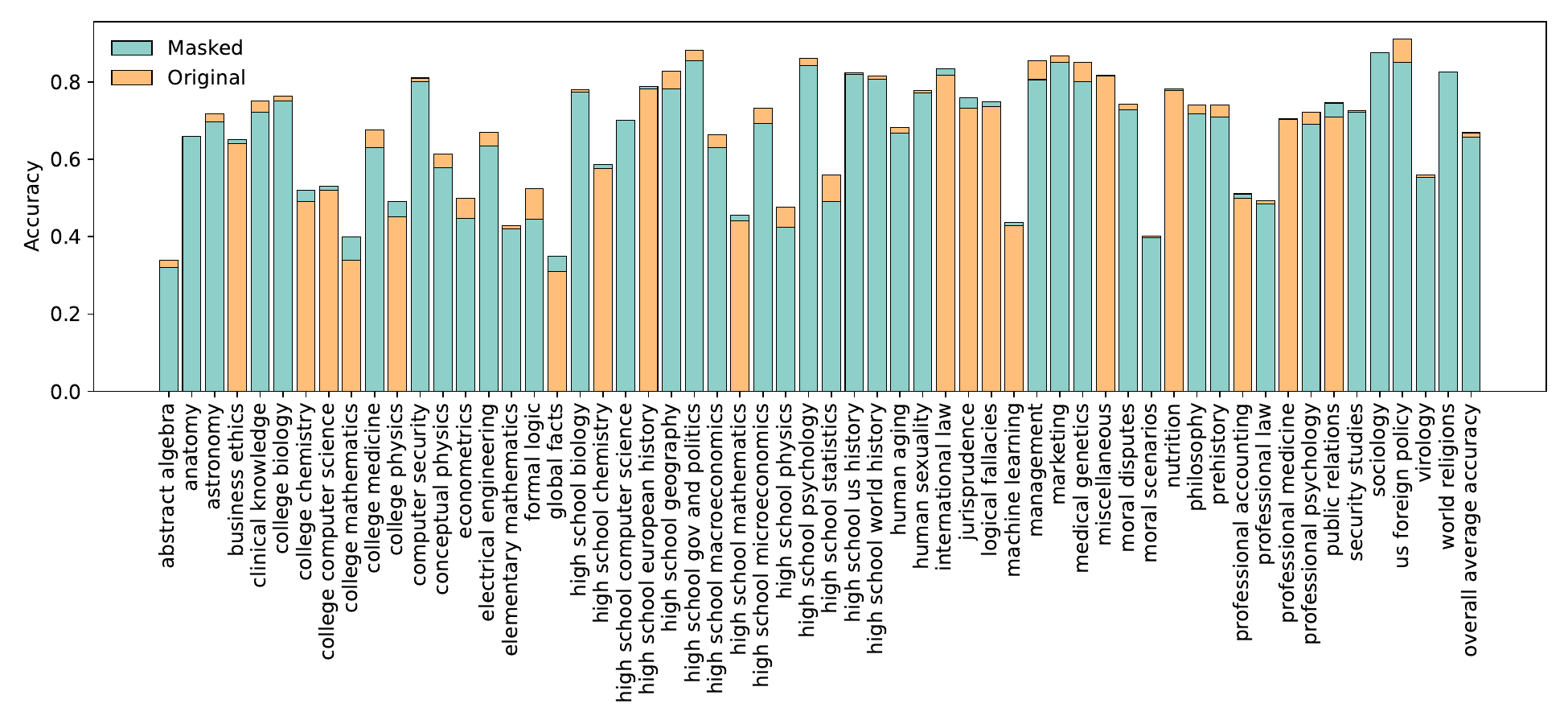}}
    \hfill
    \subfloat[Llama3.1-8B-Instruct]{%
        \includegraphics[width=0.45\textwidth]{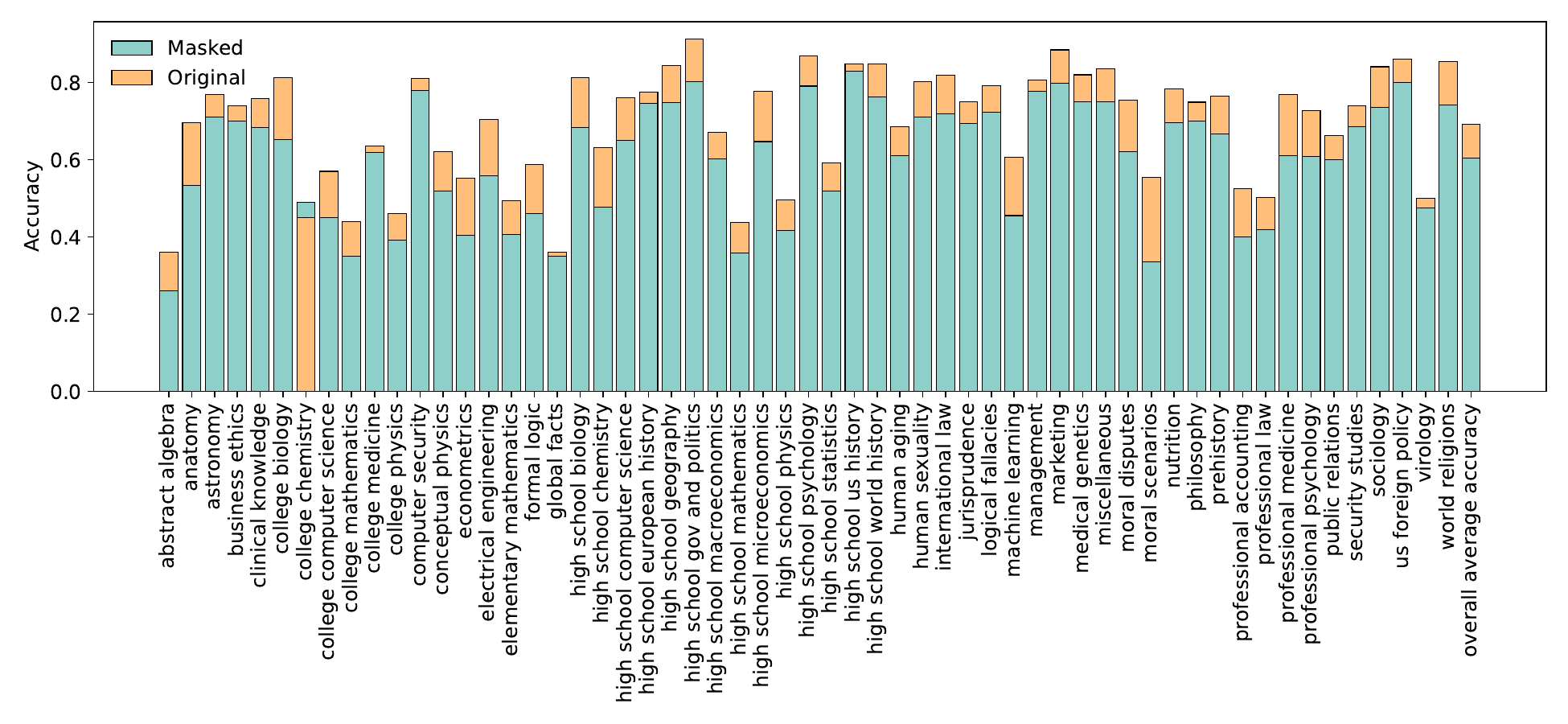}}

    \subfloat[Llama3.2-1B]{%
        \includegraphics[width=0.45\textwidth]{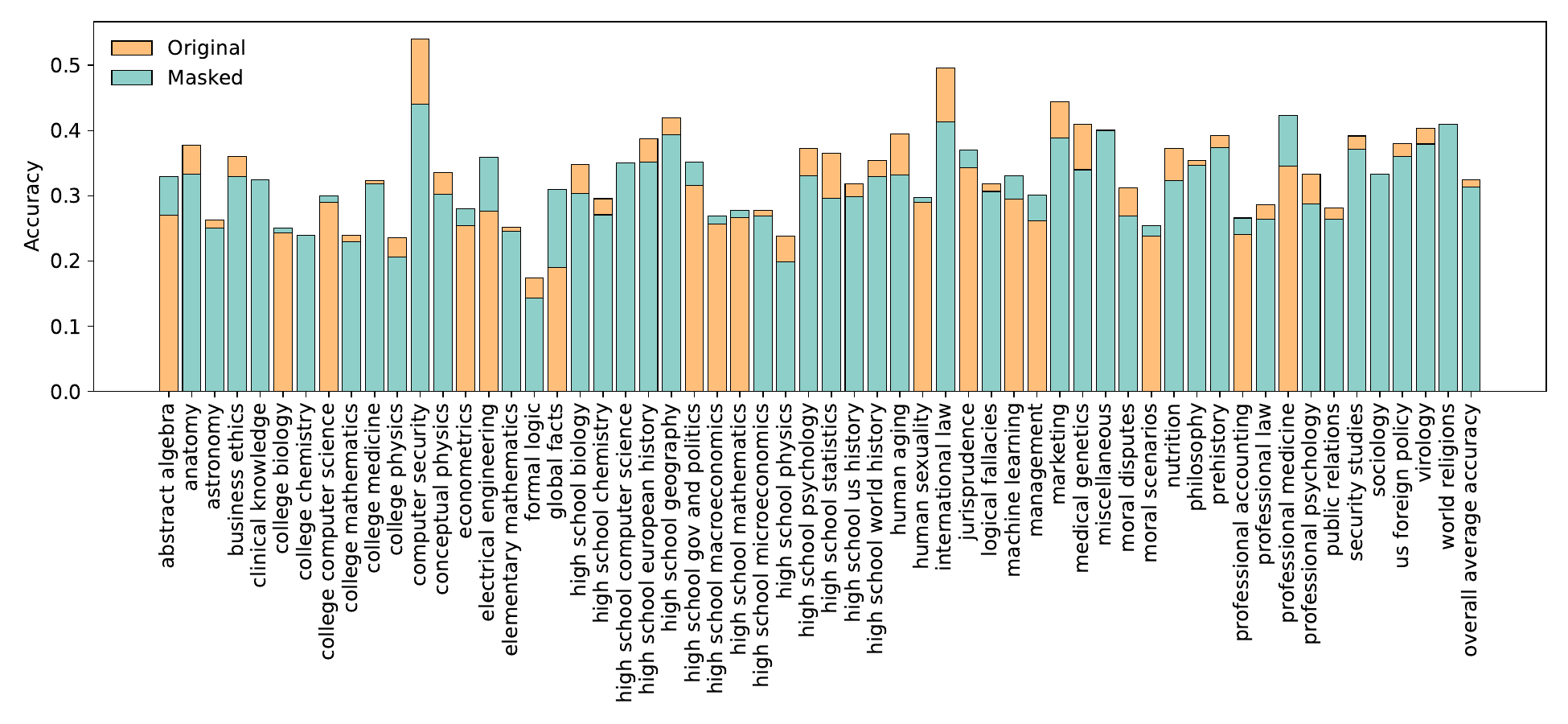}}
    \hfill
    \subfloat[Llama3.2-1B-Instruct]{%
        \includegraphics[width=0.45\textwidth]{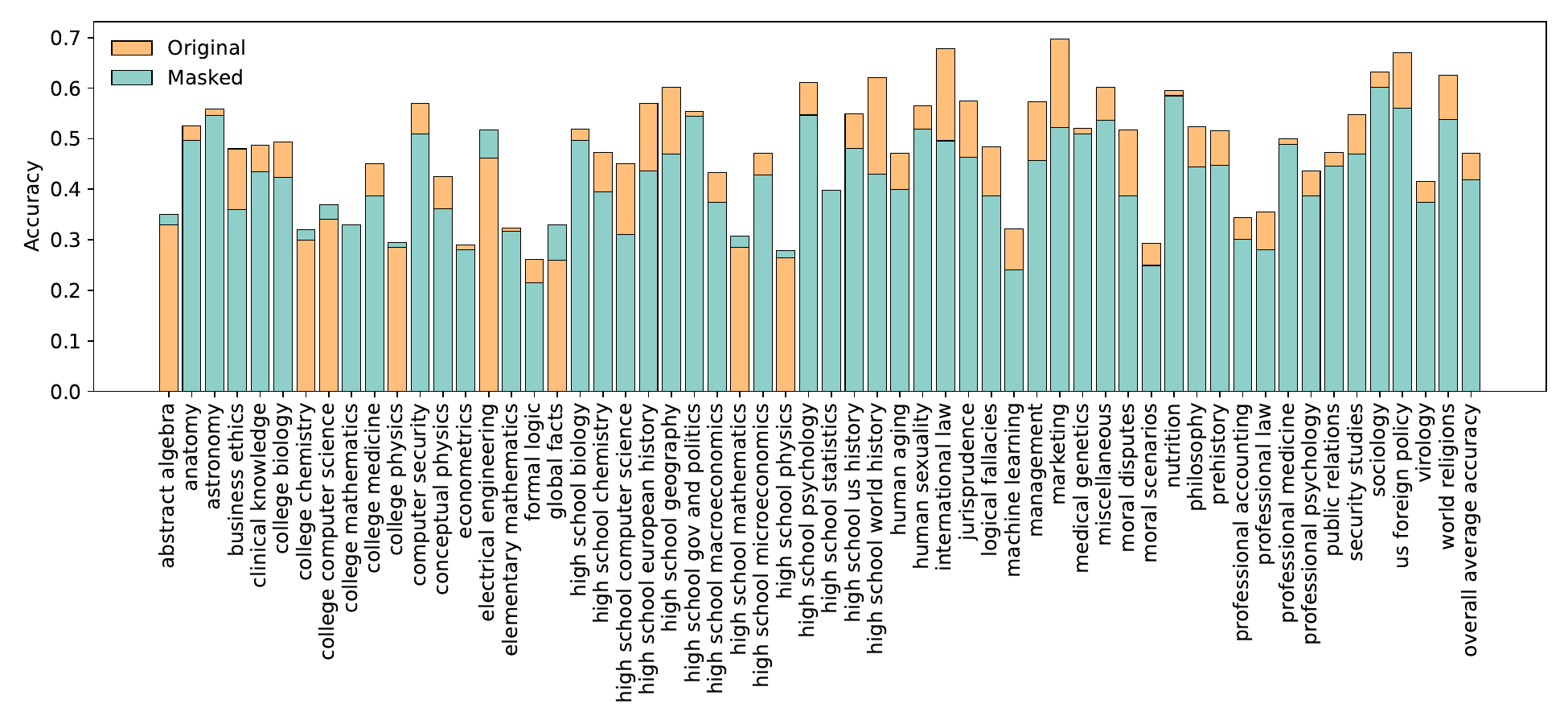}}

    \subfloat[Llama3.2-3B]{%
        \includegraphics[width=0.45\textwidth]{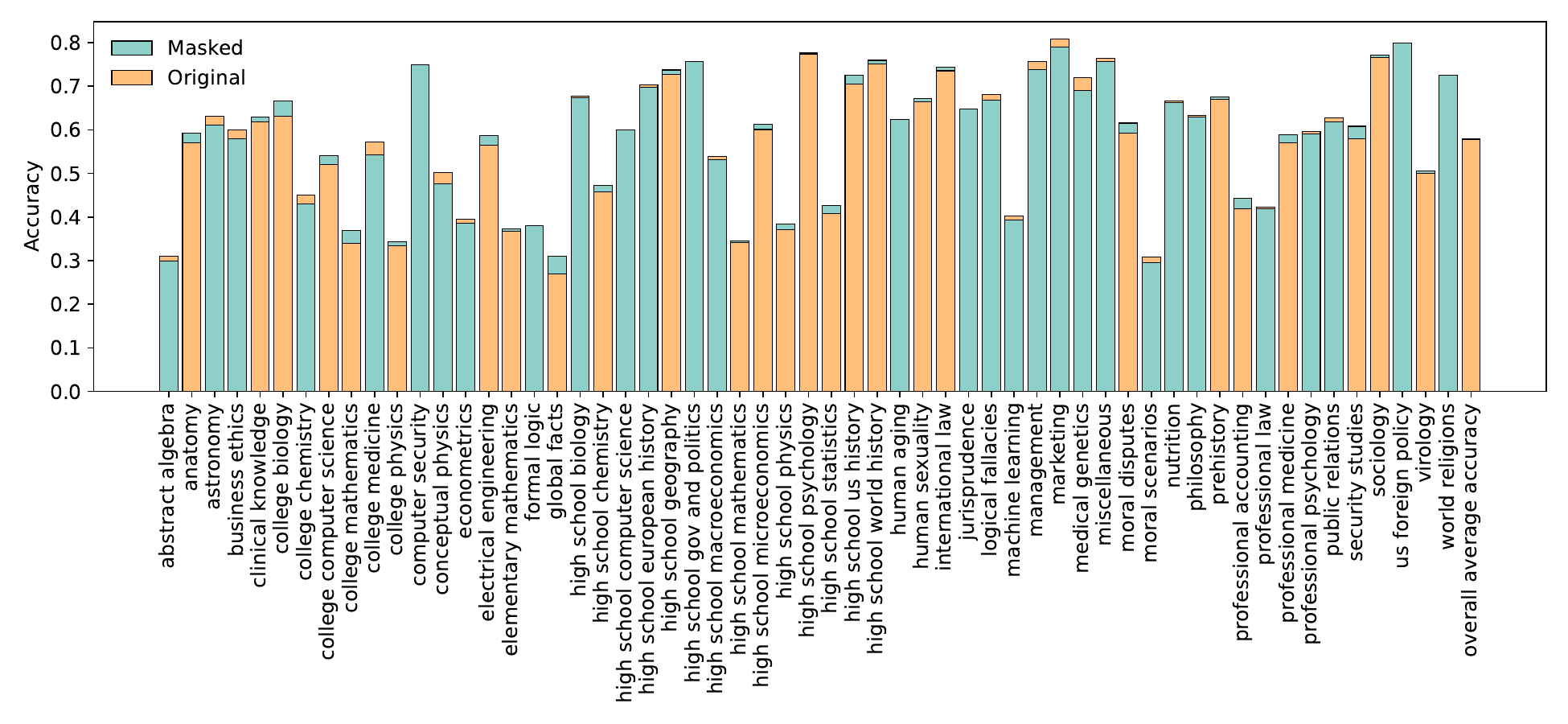}}
    \hfill
    \subfloat[Llama3.2-3B-Instruct]{%
        \includegraphics[width=0.45\textwidth]{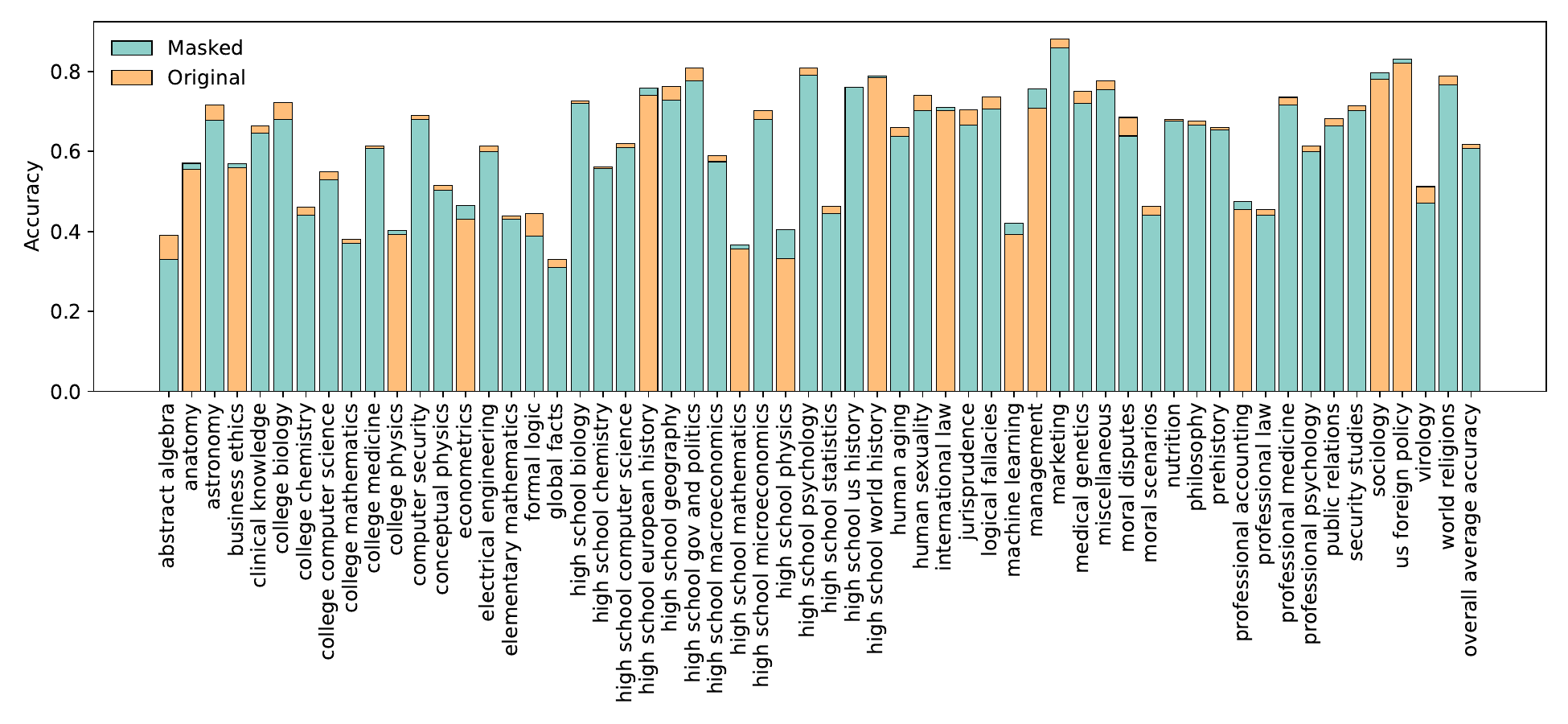}}

    \subfloat[Qwen2-7B]{%
        \includegraphics[width=0.45\textwidth]{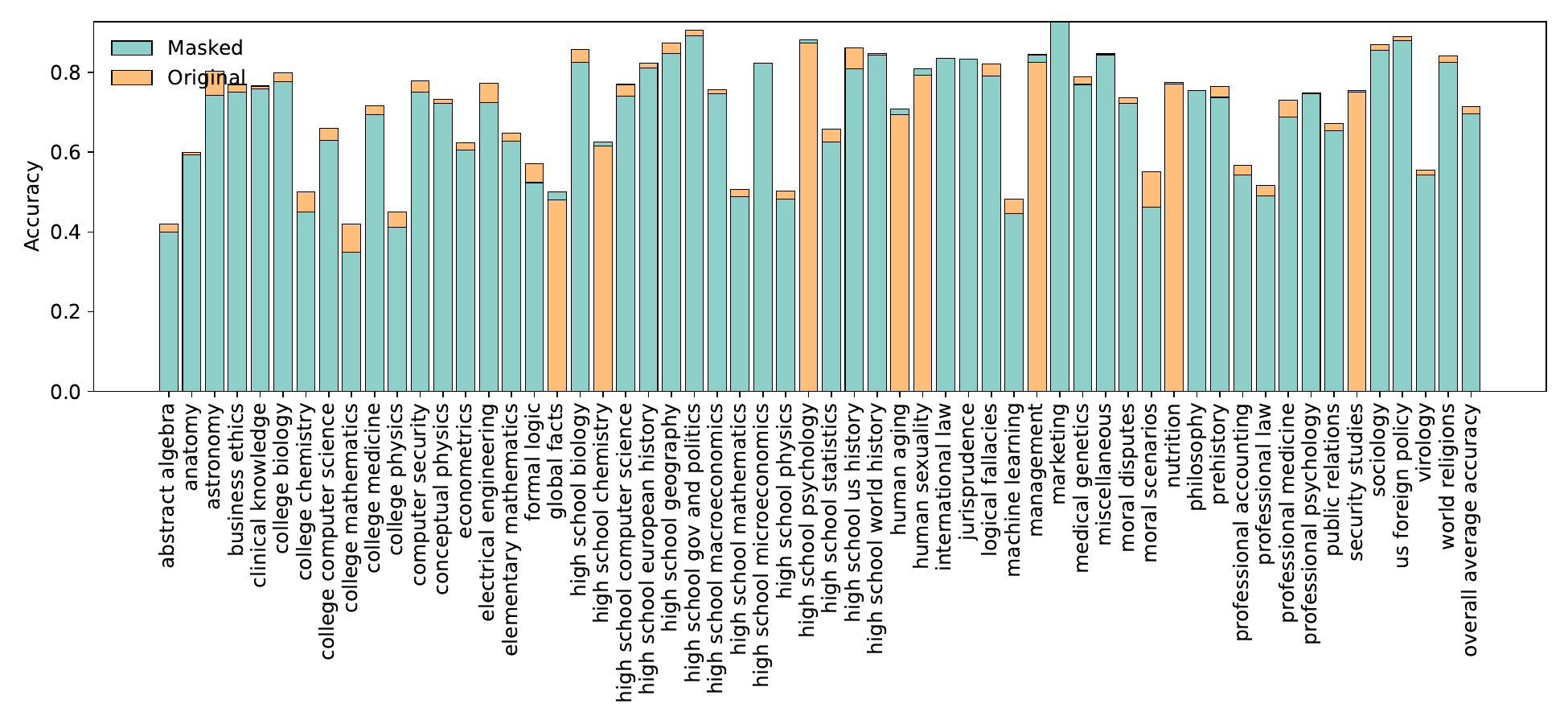}}
    \hfill
    \subfloat[Qwen2-7B-Instruct]{%
        \includegraphics[width=0.45\textwidth]{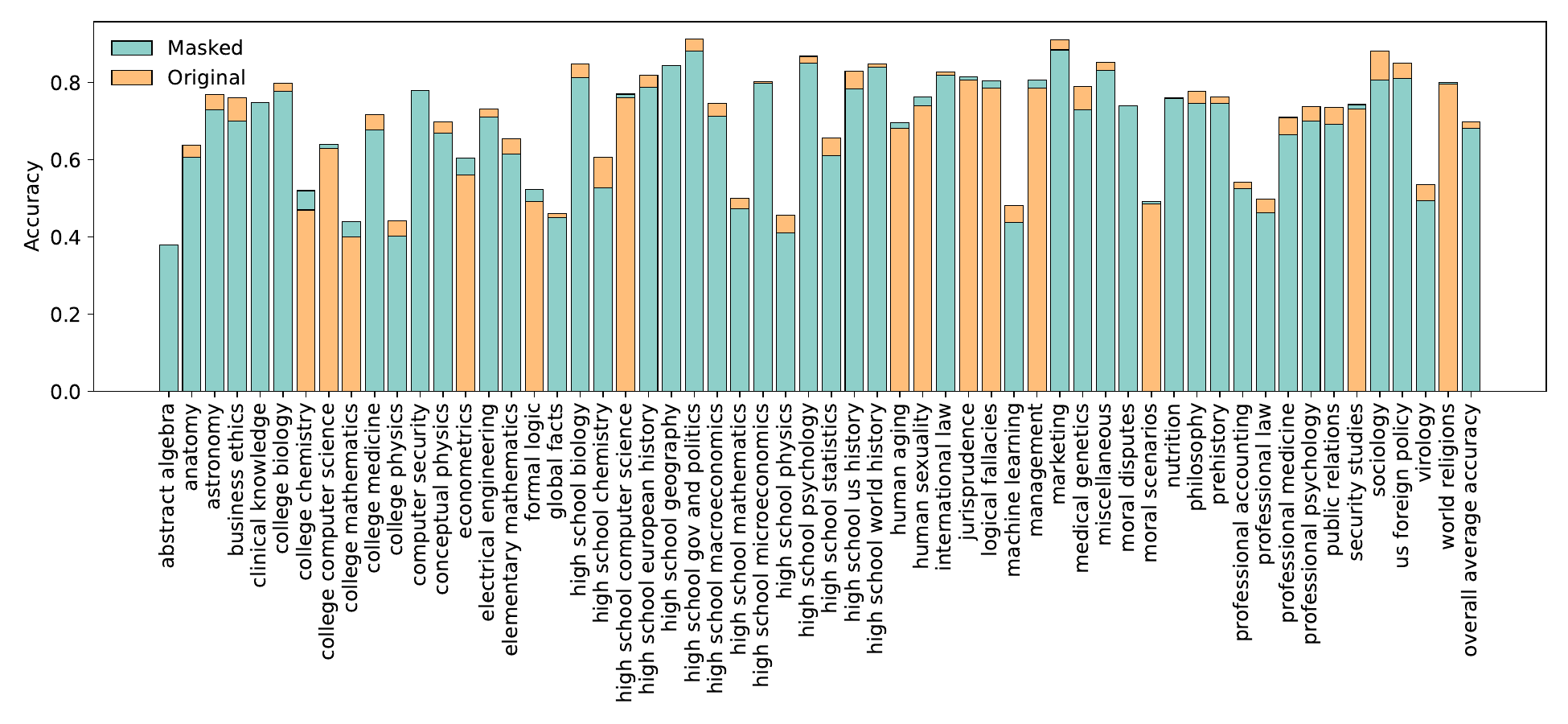}}
    \caption{Evaluating MMLU across RoPE-based models, part 1.}
    \label{fig:mmlu1}
\end{figure}

\begin{figure}[h]
    \centering

    \subfloat[Qwen2.5-7B]{%
        \includegraphics[width=0.49\textwidth]{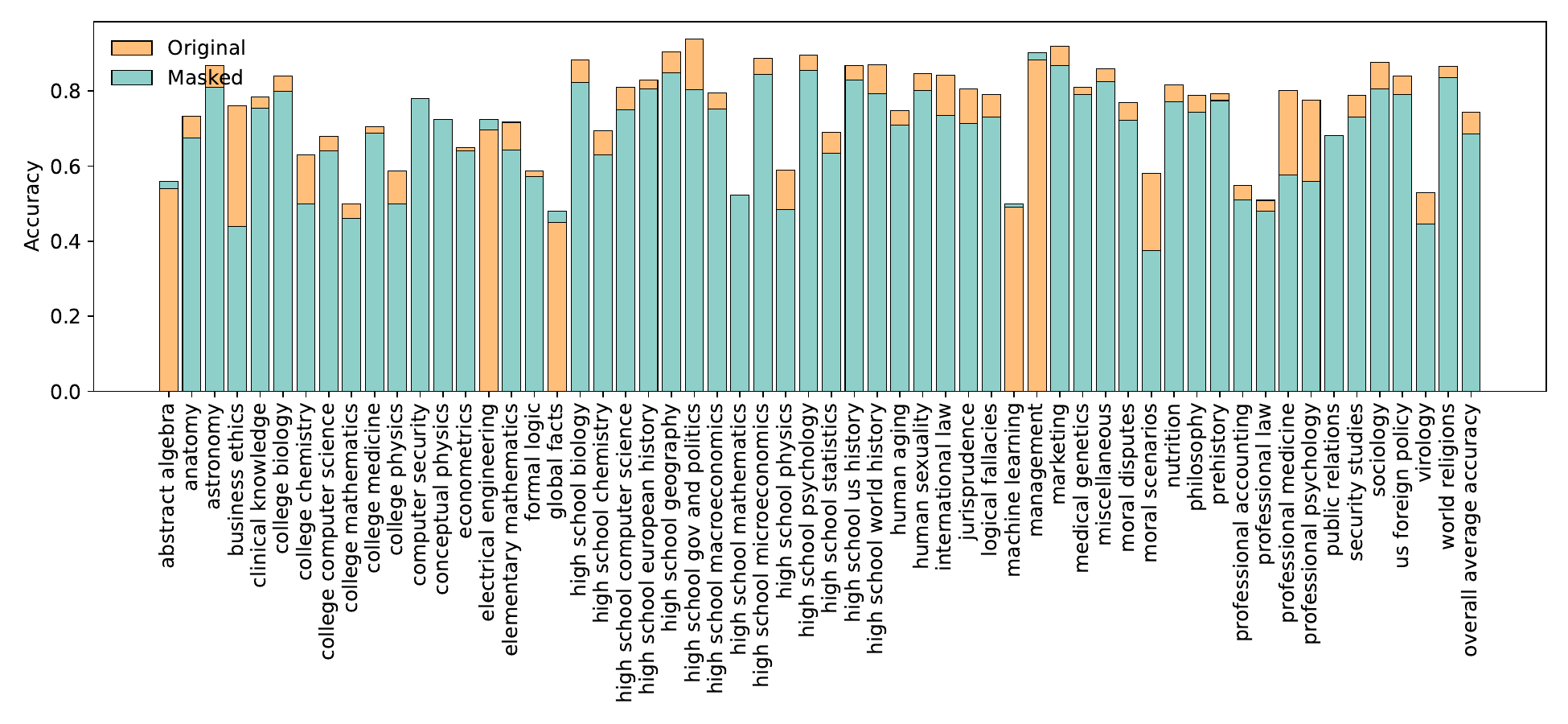}}
    \hfill
    \subfloat[Qwen2.5-7B-Instruct]{%
        \includegraphics[width=0.49\textwidth]{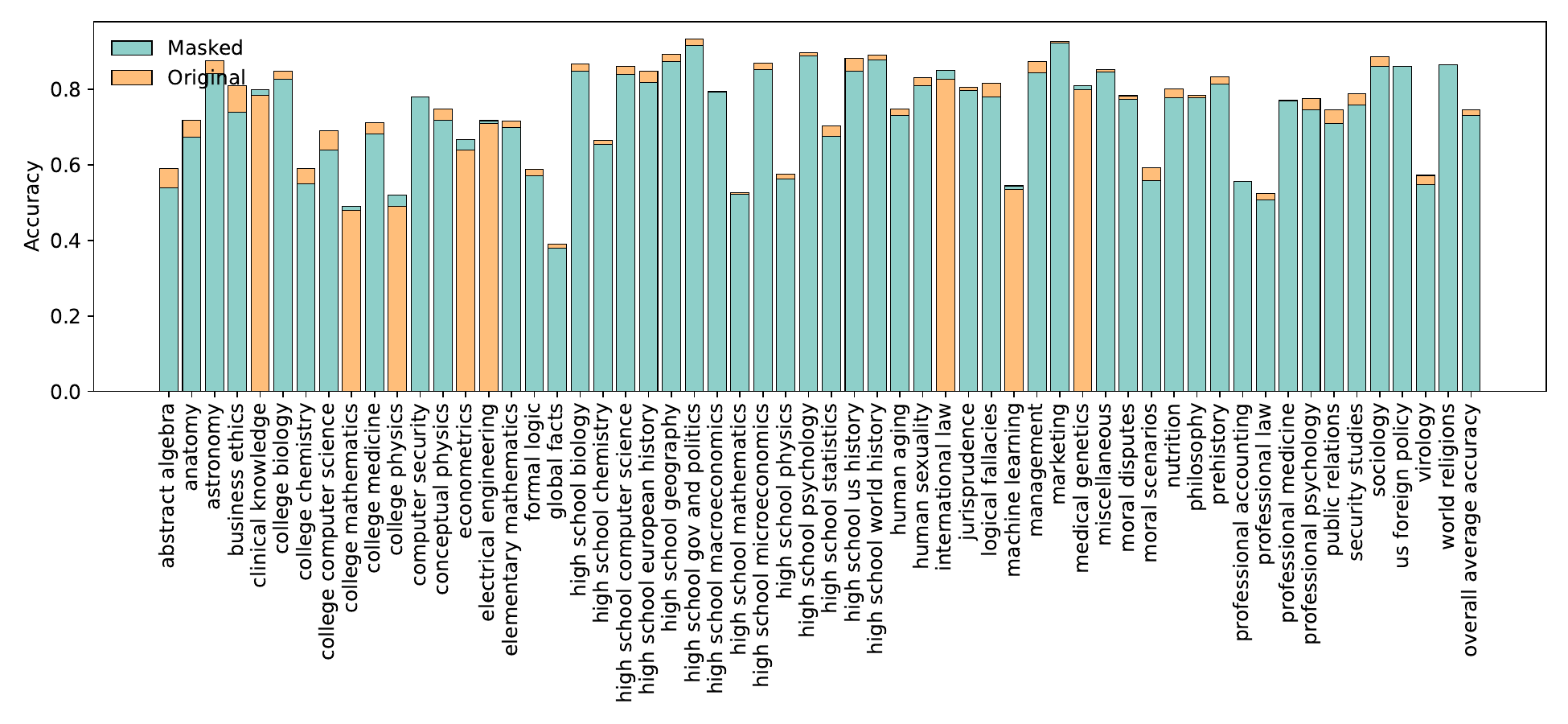}}

    \subfloat[DeepSeek-R1-Llama-8B]{%
        \includegraphics[width=0.49\textwidth]{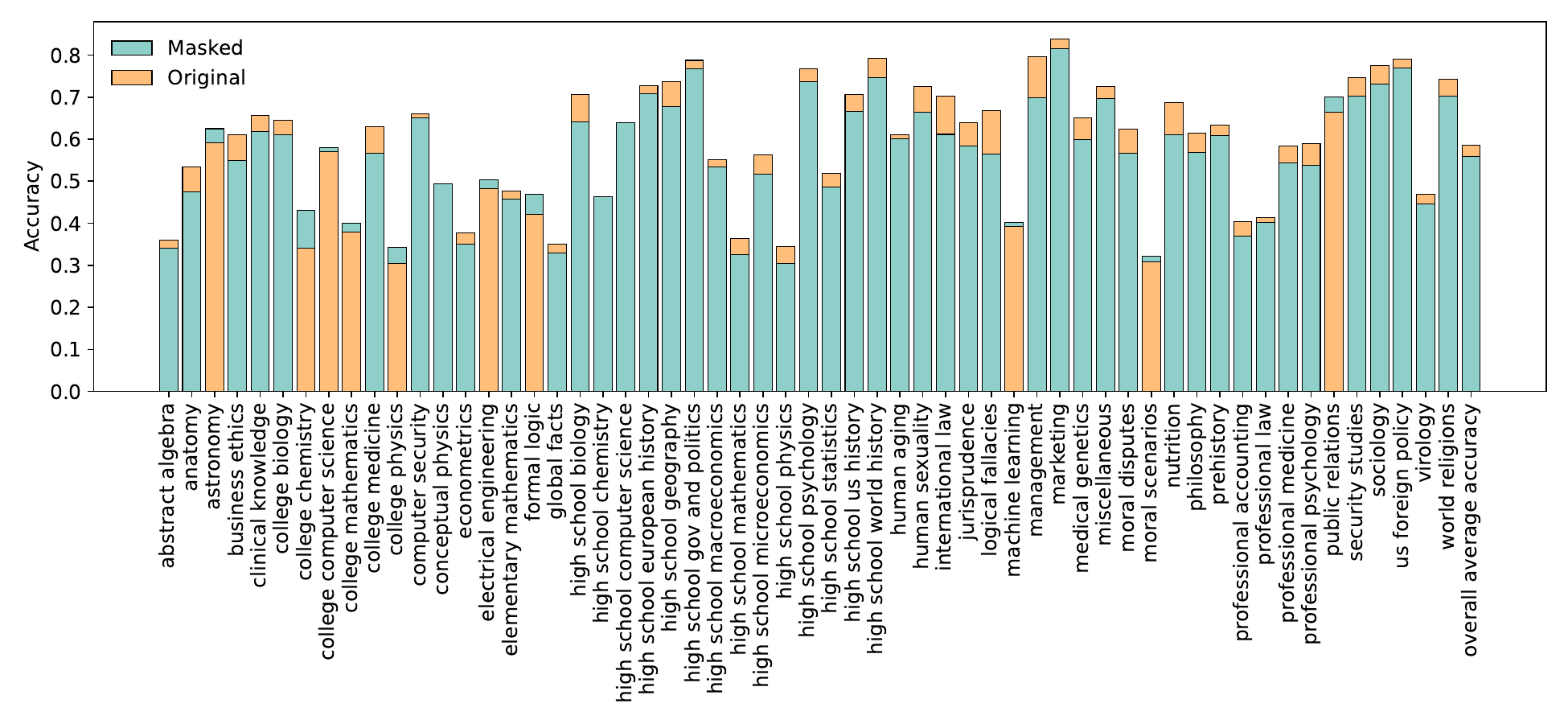}}
    \hfill
    \subfloat[DeepSeek-R1-Qwen-7B]{%
        \includegraphics[width=0.49\textwidth]{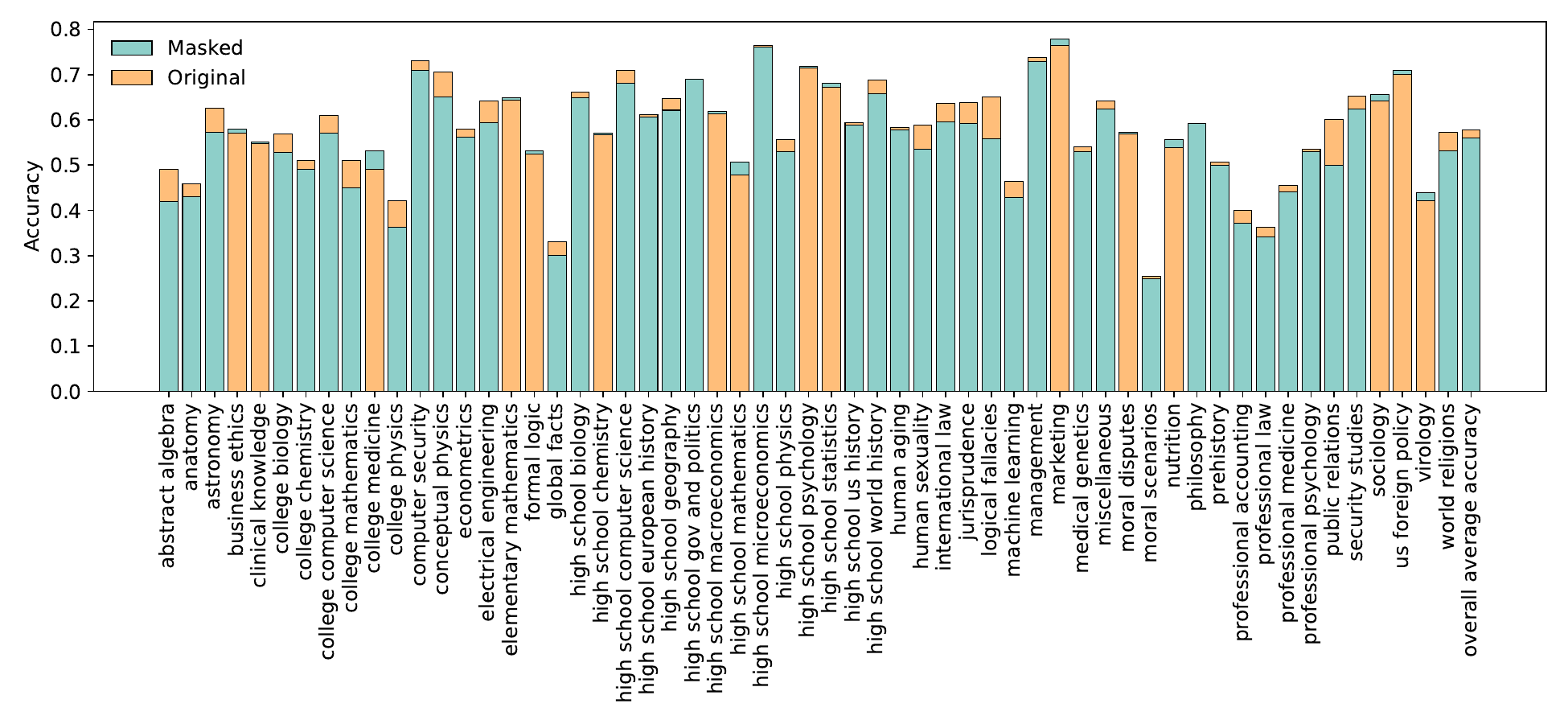}}
    \caption{Evaluating MMLU across RoPE-based models, part 2.}
    \label{fig:mmlu2}
\end{figure}

\textbf{Sparse ToM-sensitive parameter patterns impact the language understanding capabilities.} As shown in \figurename~\ref{fig:mmlu1} and \ref{fig:mmlu2}, perturbing these parameters leads to a performance decline in most tasks across the MMLU benchmark. 

\textbf{The extent of performance degradation varies with task types.} As illustrated in \figurename~\ref{fig:task_level}, tasks requiring memory and computation, such as global facts and high school mathematics, show smaller performance drops or even improvements. This suggests that the model's long-term memory remains largely intact and may even emerge more effectively after perturbing. Interestingly, this contrasts with the significant decline in the model's localization ability, indicating that ToM patterns may be more closely related to short-term memory. At the same time, tasks involving complex reasoning and emotional judgment, such as logical fallacies and moral scenarios, exhibit more pronounced performance drops. These tasks are more closely aligned with ToM-related abilities, further highlighting the importance of these patterns in higher-order cognitive functions.

\subsection{Additional results for Jamba model}
\begin{figure}[t]
  \centering
  \begin{minipage}[b]{0.48\textwidth}
    \centering
    \includegraphics[width=\textwidth]{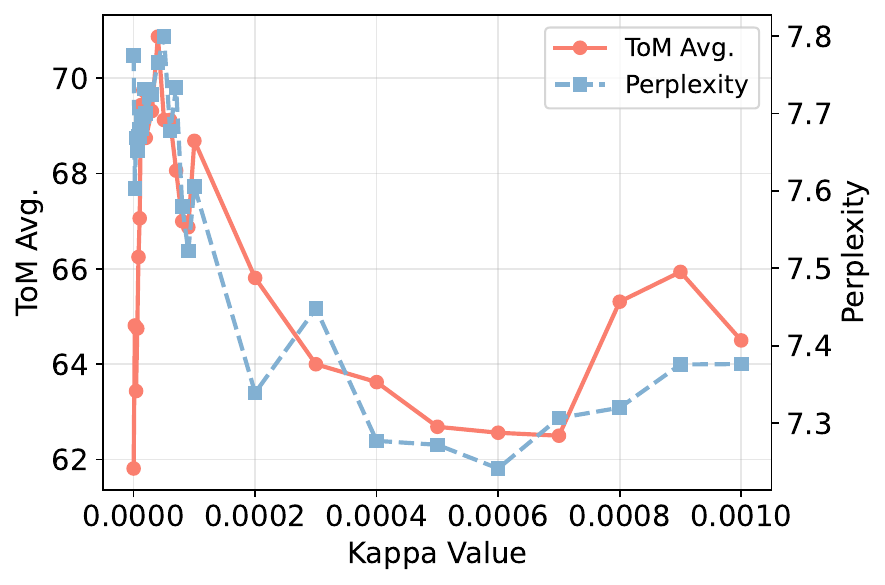}
    \caption{Average ToM performance and perplexity of Jamba across different values of \(\kappa\).}
    \label{fig:jambatomp}
  \end{minipage}
  \hfill 
  \begin{minipage}[b]{0.48\textwidth}
    \centering
    \includegraphics[width=\textwidth]{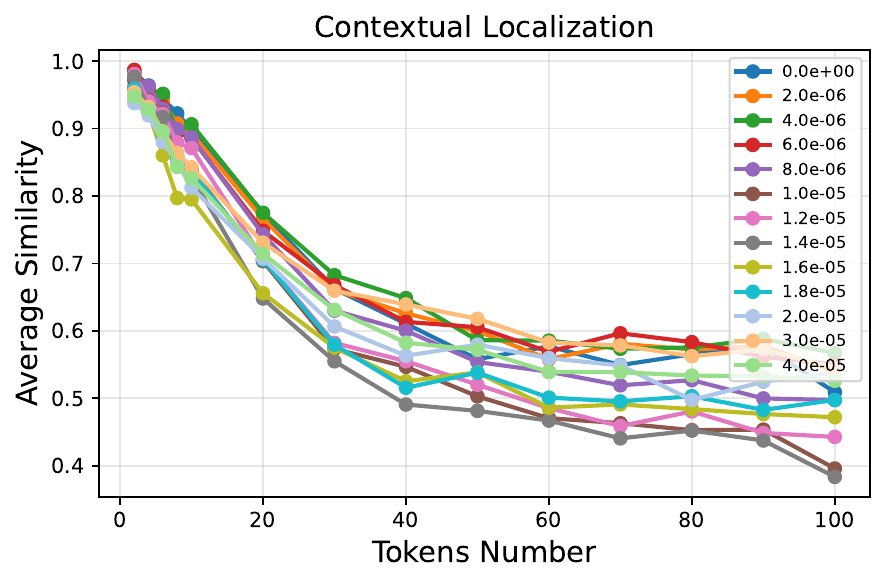}
    \caption{Contextual localization performance of Jamba across different values of \(\kappa\).}
    \label{fig:jambacontextual}
  \end{minipage}
\end{figure}

\paragraph{Lack of ToM-sensitive parameter pattern in Non-RoPE models.}
In Figure~\ref{fig:jambatomp}, we observe that no value of \(\kappa\) leads to a decline in ToM performance for the Jamba model. Instead, increasing \(\kappa\) consistently results in improved ToM performance. This suggests that the reasoning process underlying Jamba's ToM capabilities differs fundamentally from that of RoPE-based models.

\paragraph{Poor contextual localization performance in Mamba-based models.}
In Figure~\ref{fig:jambacontextual}, we observe that regardless of the value of \(\kappa\), the performance of Jamba deteriorates significantly as the token number increases. This suggests that contextual localization may represent a fundamental limitation of state-space-based models.

\label{appendix:exp_jambaresults}

\section{Additional results for Findings 2 and 3}

\subsection{Sensitive parameter mask rank analysis}
\label{appendix:C_rank}
\begin{table}[h]
\centering
\caption{Sensitive parameter mask rank analysis for LLaMA3-8B (\(\kappa = 0.000030\))}
\label{tab:average_rankllama}
\resizebox{0.85\textwidth}{!}{
\begin{tabular}{cccccccc}
\toprule
& \(W_\mathbf{Q}\) & \(W_\mathbf{K}\) & \(W_\mathbf{V}\) & \(W_\mathbf{O}\) & \(W_\mathbf{Gate}\) & \(W_\mathbf{Up}\) & \(W_\mathbf{Down}\) \\
\midrule
Original Rank & 4020.91 & 1022.72 & 1024.00 & 4083.50 & 4096.00 & 4096.00 & 4096.00 \\
Mask Rank & 21.69 & 10.50 & 6.88 & 16.06 & 29.66 & 26.09 & 17.56 \\
Normalized Mask Rank & 0.5774 & 0.6915 & 0.8512 & 0.5960 & 0.3235 & 0.3002 & 0.6484 \\
\bottomrule
\end{tabular}
}
\end{table}

\begin{table}[h]
\centering
\caption{Sensitive parameter mask rank analysis for Jamba-1.5-Mini (\(\kappa = 0.000030\))}
\label{tab:average_rankjamba}
\resizebox{0.85\textwidth}{!}{
\begin{tabular}{cccccccc}
\toprule
 & \(W_\mathbf{Q}\) & \(W_\mathbf{K}\) & \(W_\mathbf{V}\) & \(W_\mathbf{O}\) & \(W_\mathbf{Gate}\) & \(W_\mathbf{Up}\) & \(W_\mathbf{Down}\) \\
\midrule
Original Rank & 3994.25 & 1024.00 & 1024.00 & 4073.00 & 4096.00 & 4096.00 & 4096.00 \\
Mask Rank & 11.00 & 7.75 & 3.50 & 14.00 & 15.75 & 13.50 & 14.25 \\
Normalized Mask Rank & 0.7433 & 0.7423 & 1.0000 & 0.4107 & 0.5641 & 0.6643 & 0.5541 \\
\bottomrule
\end{tabular}
}
\end{table}

\paragraph{The ToM-sensitive parameter pattern is sparse and extremely low-rank.} 
Given that our \(\kappa\) is on the order of $10^{-5}$, the resulting masks are naturally sparse, which in turn induces a low-rank structure. However, we argue that it is not just low-rank but \emph{extremely} low-rank; namely, these few parameters are concentrated in a limited number of rows (or columns). To quantify this, we introduce the \emph{normalized mask rank}, defined as the ratio of the mask rank to the mask's non-zero rows (or columns) number. We compute this metric for all layers of LLaMA3-8B and report the average values in \tablename~\ref{tab:average_rankllama}. The results clearly indicate that the generated masks are extremely low-rank. 

\paragraph{Non-RoPE-based model also exhibits an extremely sparse pattern.}
We also conduct a rank analysis of the Jamba model and present the results in \tablename~\ref{tab:average_rankjamba}. Similar to RoPE-based models, we observe that the sensitive parameter pattern in Jamba is also extremely sparse. However, we find that the normalized mask ranks of \(W_\mathbf{Q}\), \(W_\mathbf{K}\), and \(W_\mathbf{V}\) in Jamba are consistently higher than those in LLaMA3-8B, suggesting that the sensitive patterns in state-space-based models might exhibit different structural characteristics.

\subsection{Perturbed weights value}
\label{appendix:C_weightvalue}

\begin{figure}[H]
    \centering
    \subfloat[Llama3-8B]{%
        \includegraphics[width=0.33\textwidth]{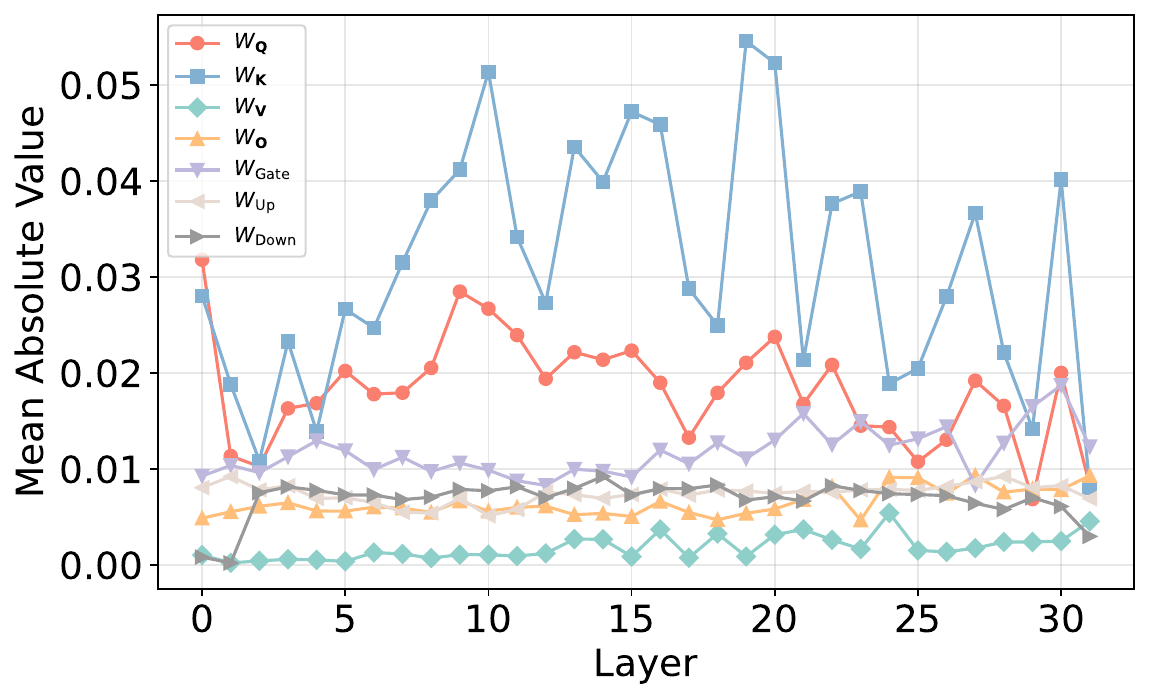}}
    \hfill
    \subfloat[Llama3-8B-Instruct]{%
        \includegraphics[width=0.33\textwidth]{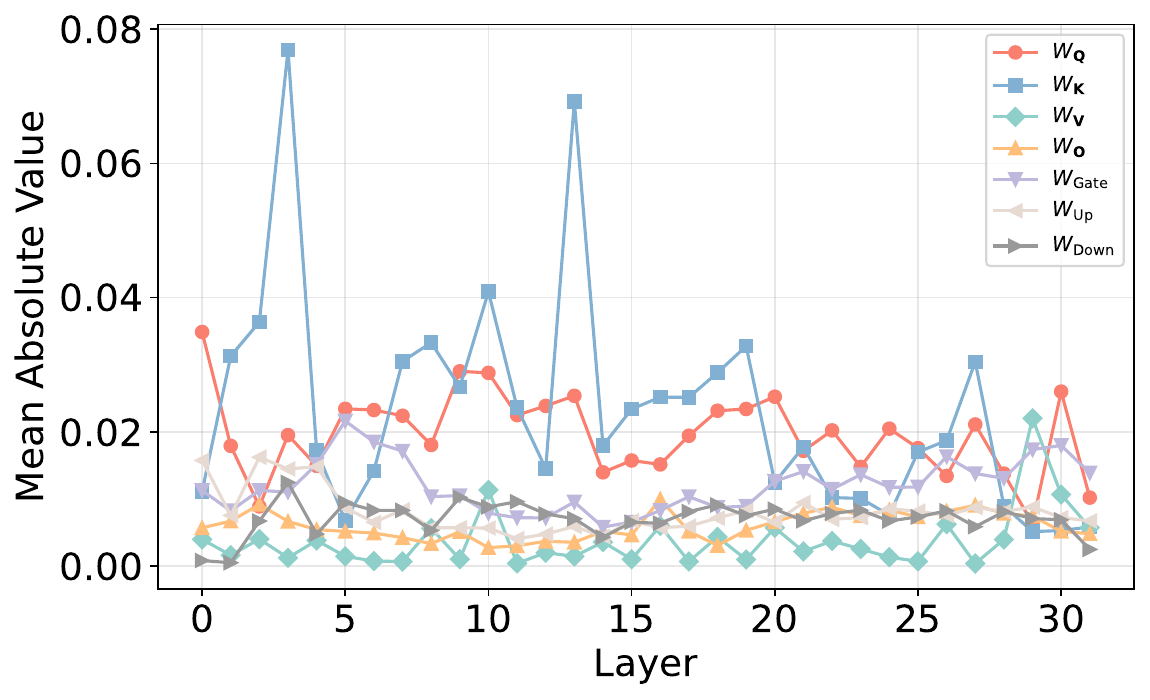}}
    \hfill
    \subfloat[Llama3.1-8B]{%
        \includegraphics[width=0.326\textwidth]{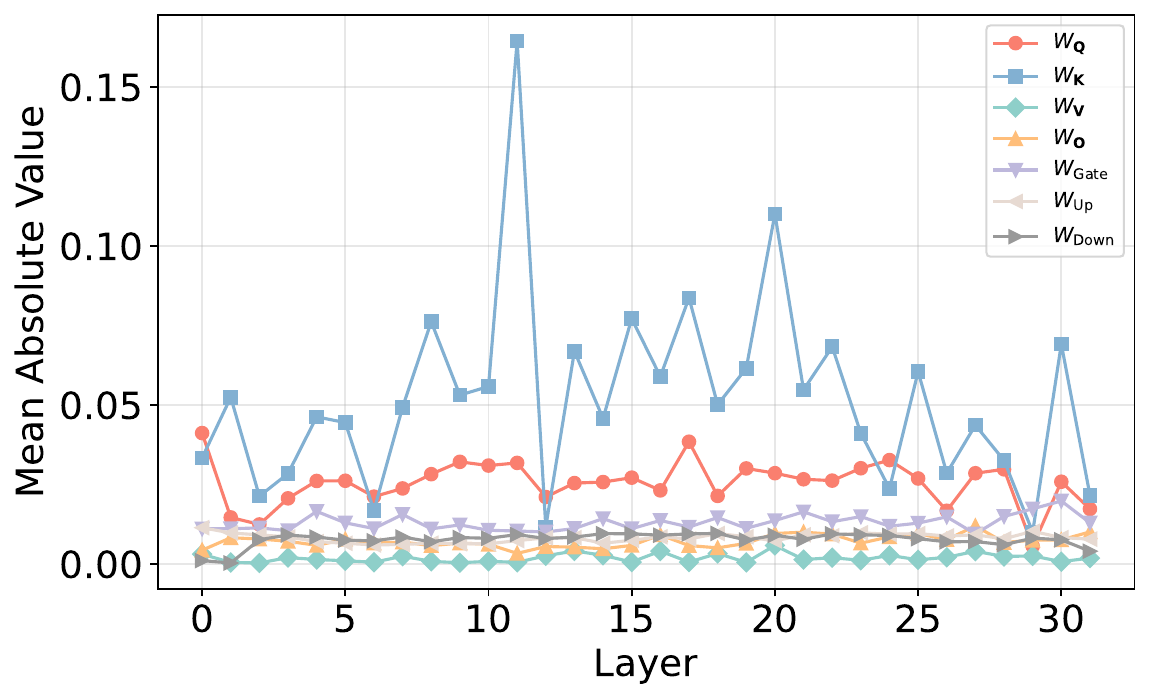}}

    \subfloat[Llama3.1-8B-Instruct]{%
        \includegraphics[width=0.33\textwidth]{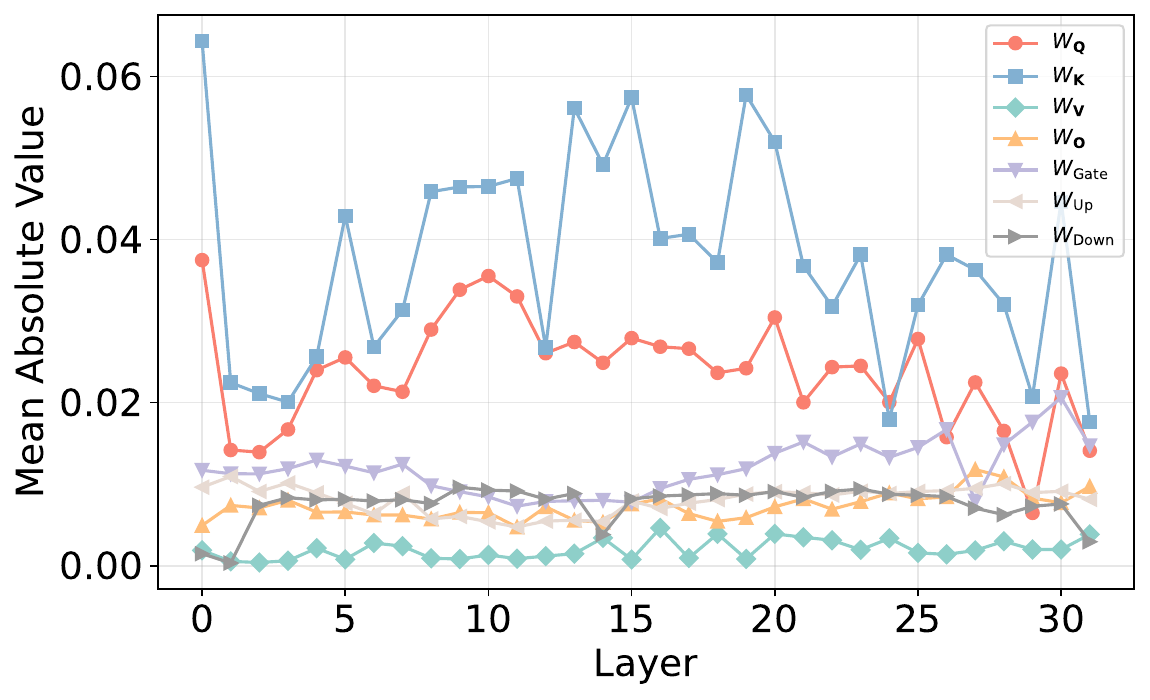}}
    \hfill
    \subfloat[Llama3.2-1B]{%
        \includegraphics[width=0.33\textwidth]{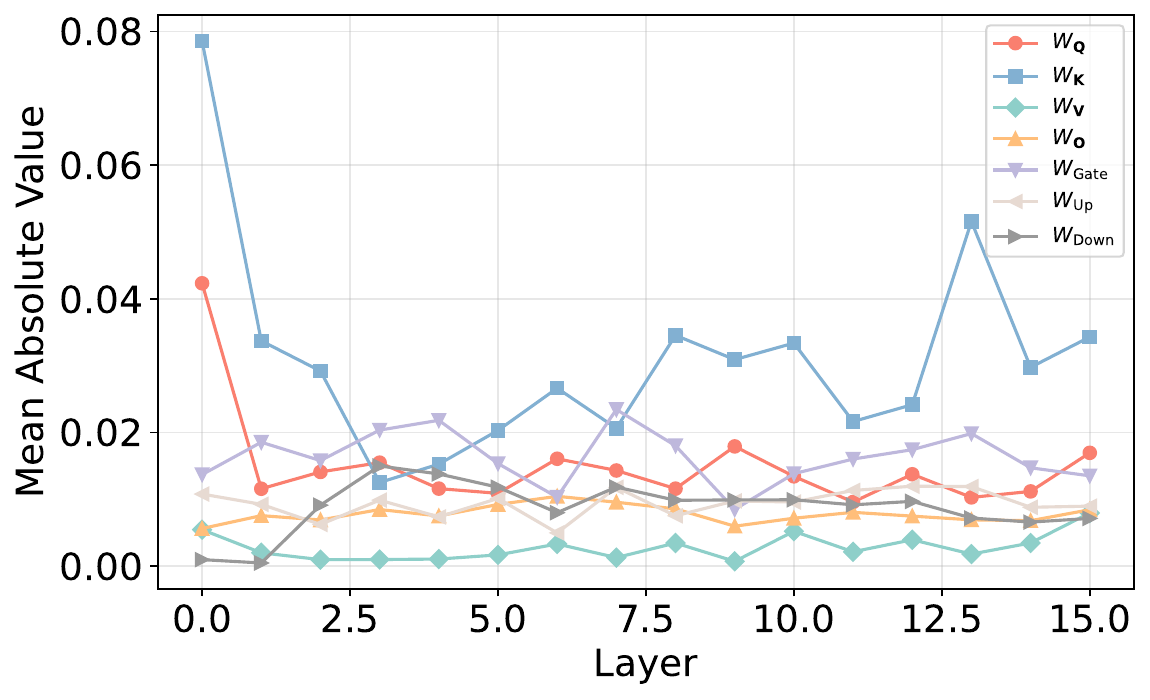}}
    \hfill
    \subfloat[Llama3.2-1B-Instruct]{%
        \includegraphics[width=0.33\textwidth]{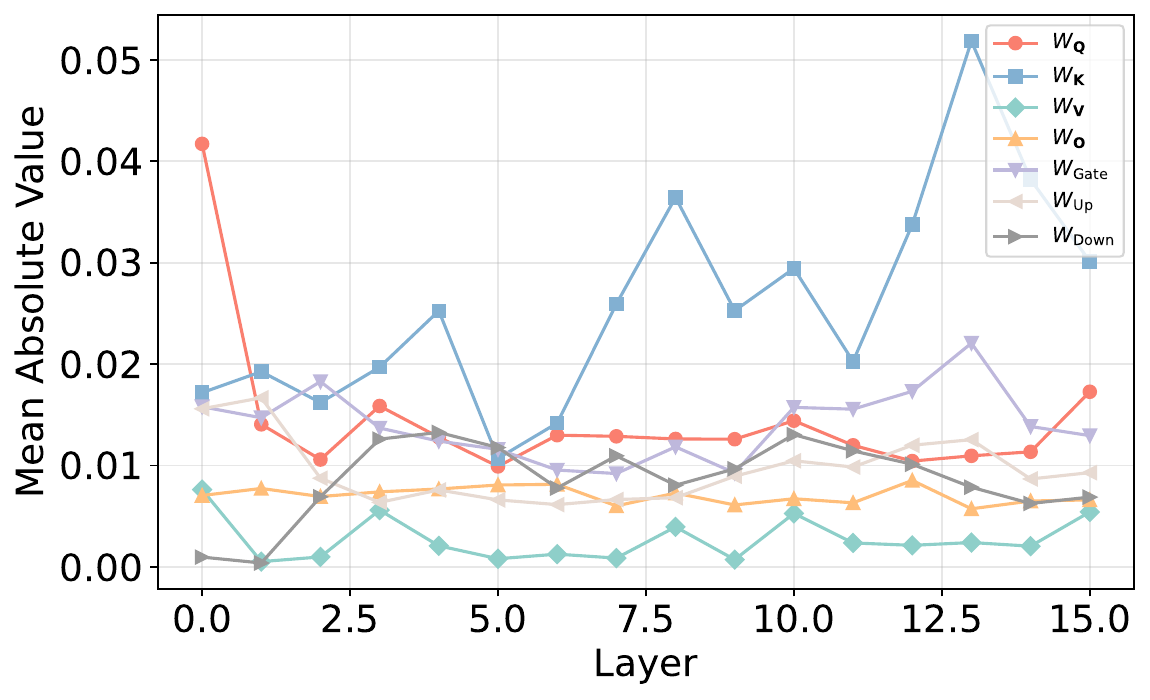}}

    \subfloat[Llama3.2-3B]{%
        \includegraphics[width=0.33\textwidth]{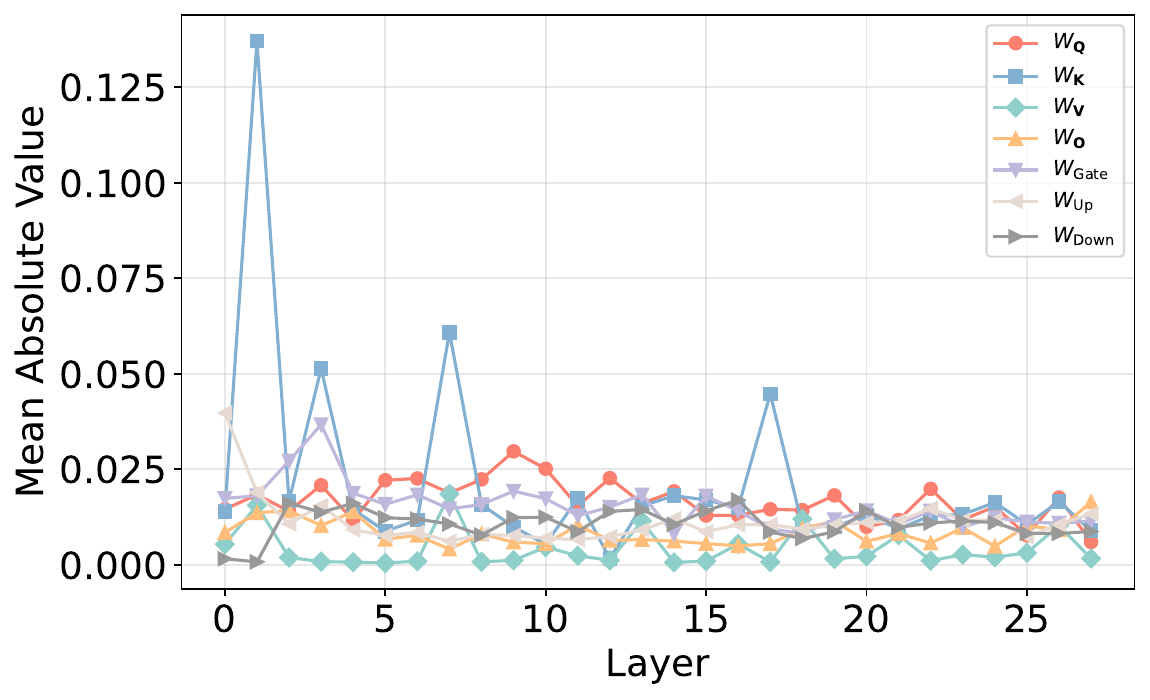}}
    \hfill
    \subfloat[Llama3.2-3B-Instruct]{%
        \includegraphics[width=0.33\textwidth]{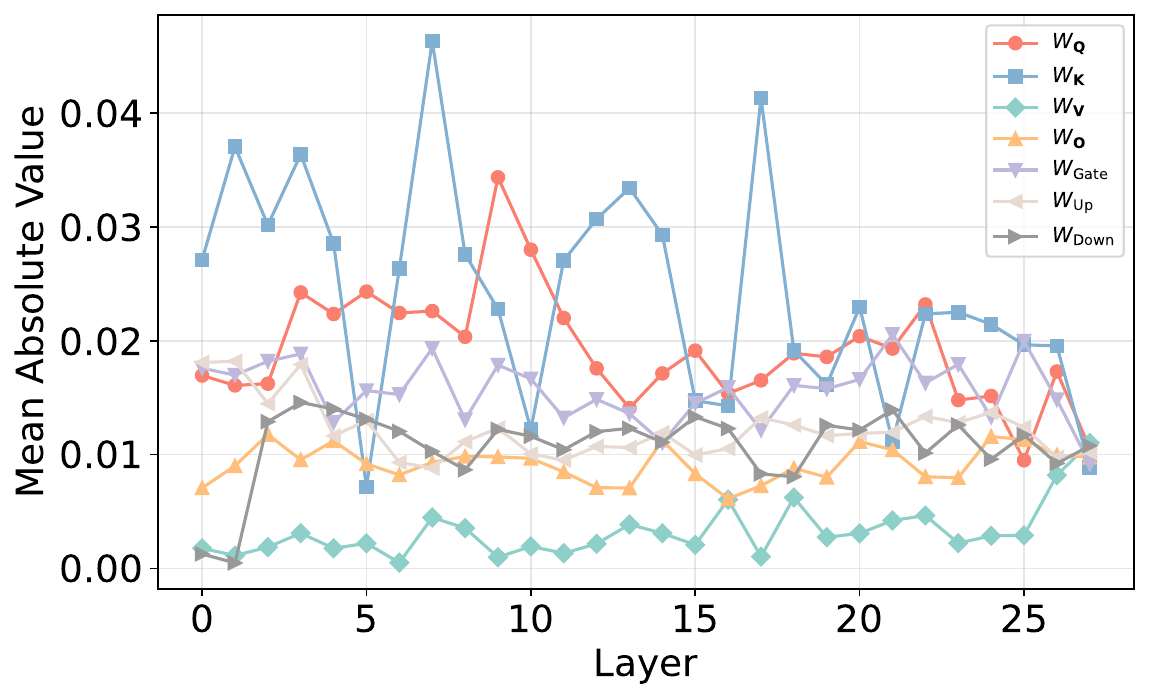}}
    \hfill
    \subfloat[Qwen2-7B]{%
        \includegraphics[width=0.33\textwidth]{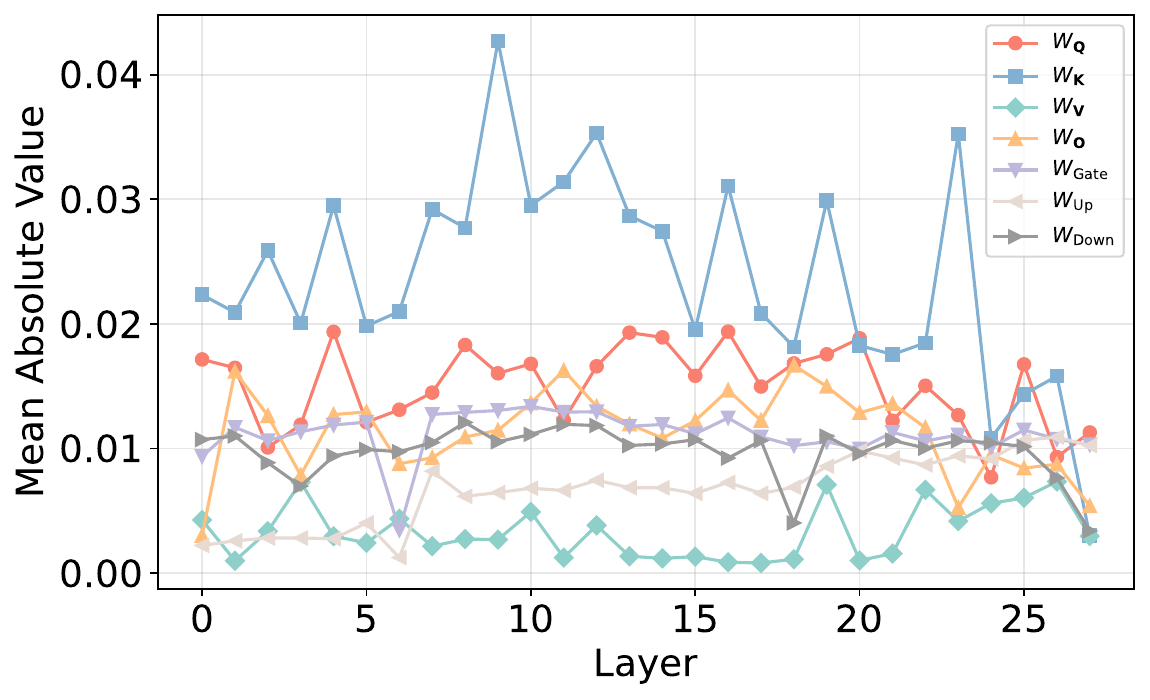}}

    \subfloat[Qwen2-7B-Instruct]{%
        \includegraphics[width=0.33\textwidth]{qwen2-7B.pdf}}
    \hfill
    \subfloat[Qwen2.5-7B]{%
        \includegraphics[width=0.33\textwidth]{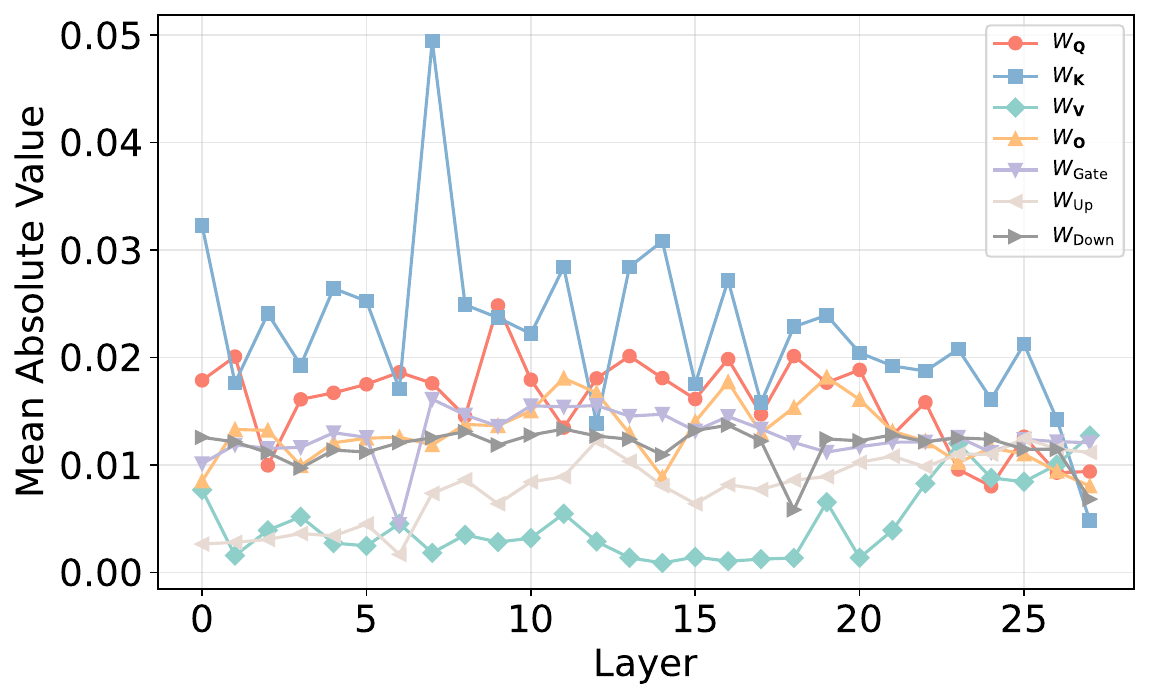}}
    \hfill
    \subfloat[Qwen2.5-7B-Instruct]{%
        \includegraphics[width=0.33\textwidth]{qwen2.5-7B.pdf}}

    \subfloat[DeepSeek-Llama-8B]{%
        \includegraphics[width=0.33\textwidth]{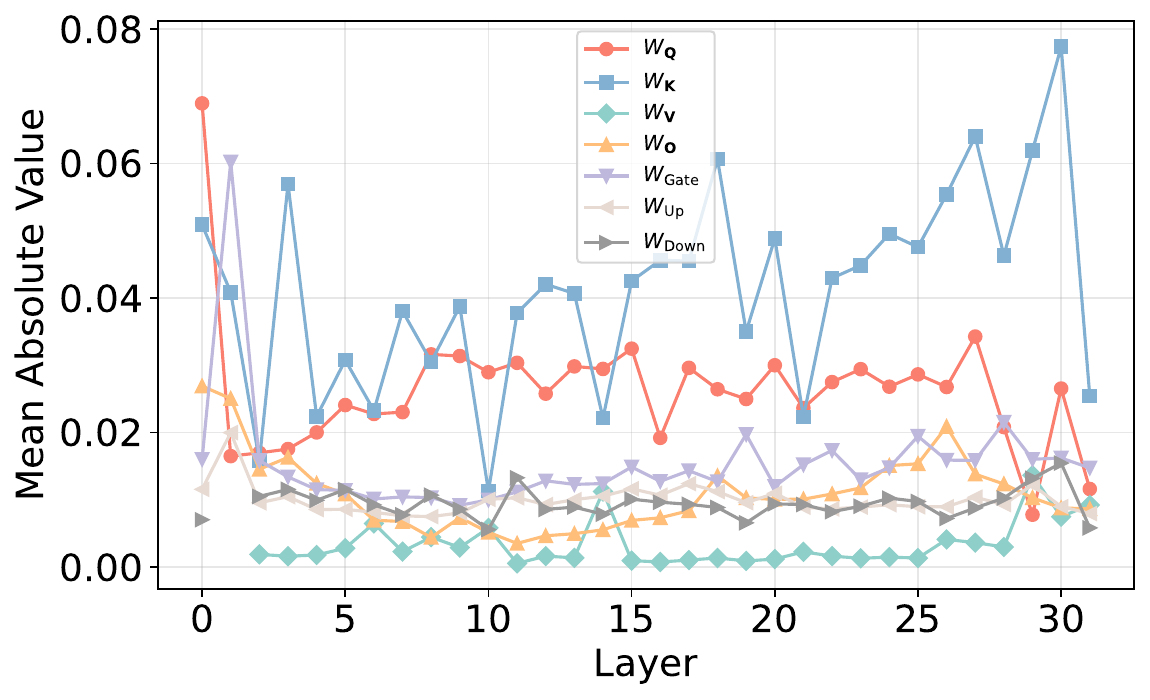}}
    \hfill
    \subfloat[DeepSeek-Qwen-7B]{%
     \includegraphics[width=0.33\textwidth]{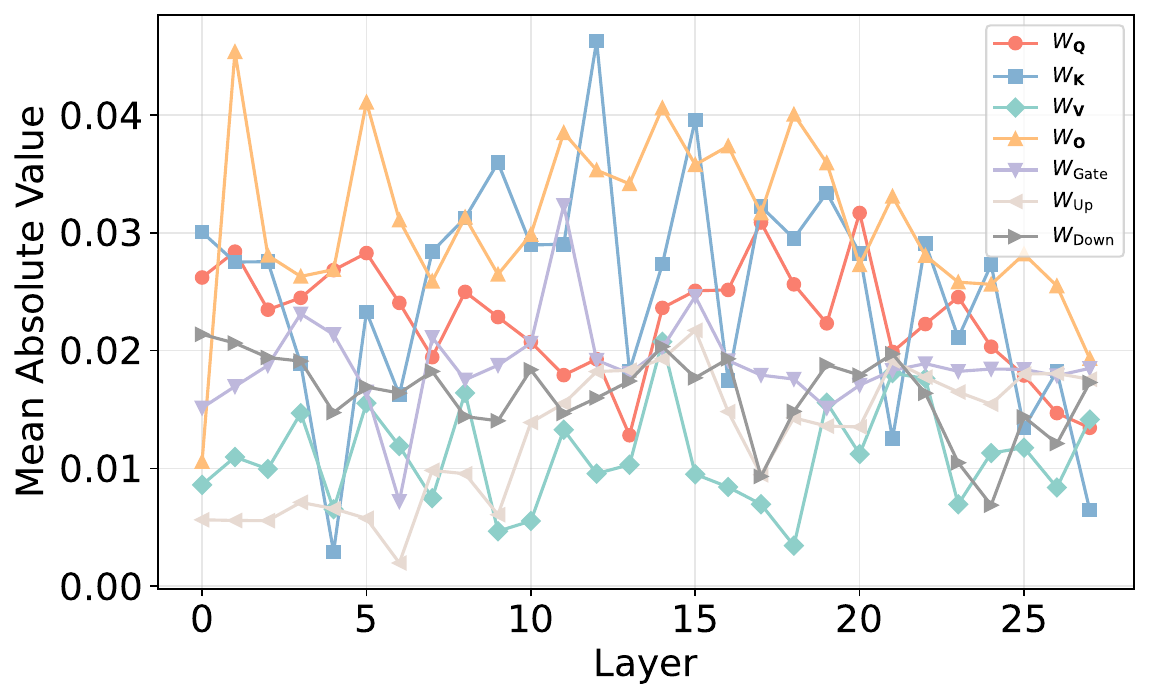}}
     \hfill
    \subfloat[Jamba-1.5-Mini]{%
     \includegraphics[width=0.33\textwidth]{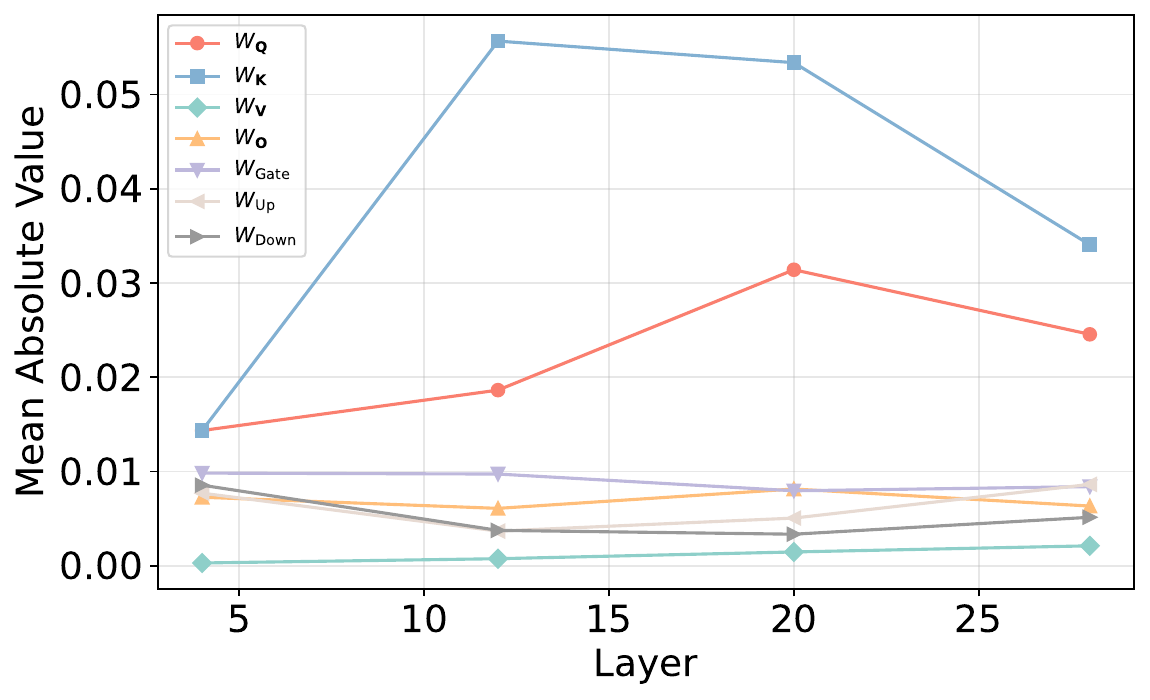}}
    \hfill
    
    \caption{Distribution of absolute values of perturbed weights across different layers and matrices.}
    \label{fig:WEIGHT}
\end{figure}

\paragraph{Perturbed values in $W_{\mathbf{Q}}$ and $W_{\mathbf{K}}$ matrices are significantly larger.} We visualize the mean absolute values of the perturbed weights across different layers and matrices. As shown in \figurename~\ref{fig:WEIGHT}, the weights in the $W_{\mathbf{Q}}$ and $W_{\mathbf{K}}$ matrices exhibit notably larger perturbations than those in other matrices. This suggests that changes in model performance may be closely tied to the attention mechanism.

\subsection{Visualization of weight distributions and activations}
\label{appendix:C_activations}

\begin{figure}[h]
    \centering
    \subfloat[ToM-sensitive parameter distribution]{%
\includegraphics[width=0.46\textwidth]{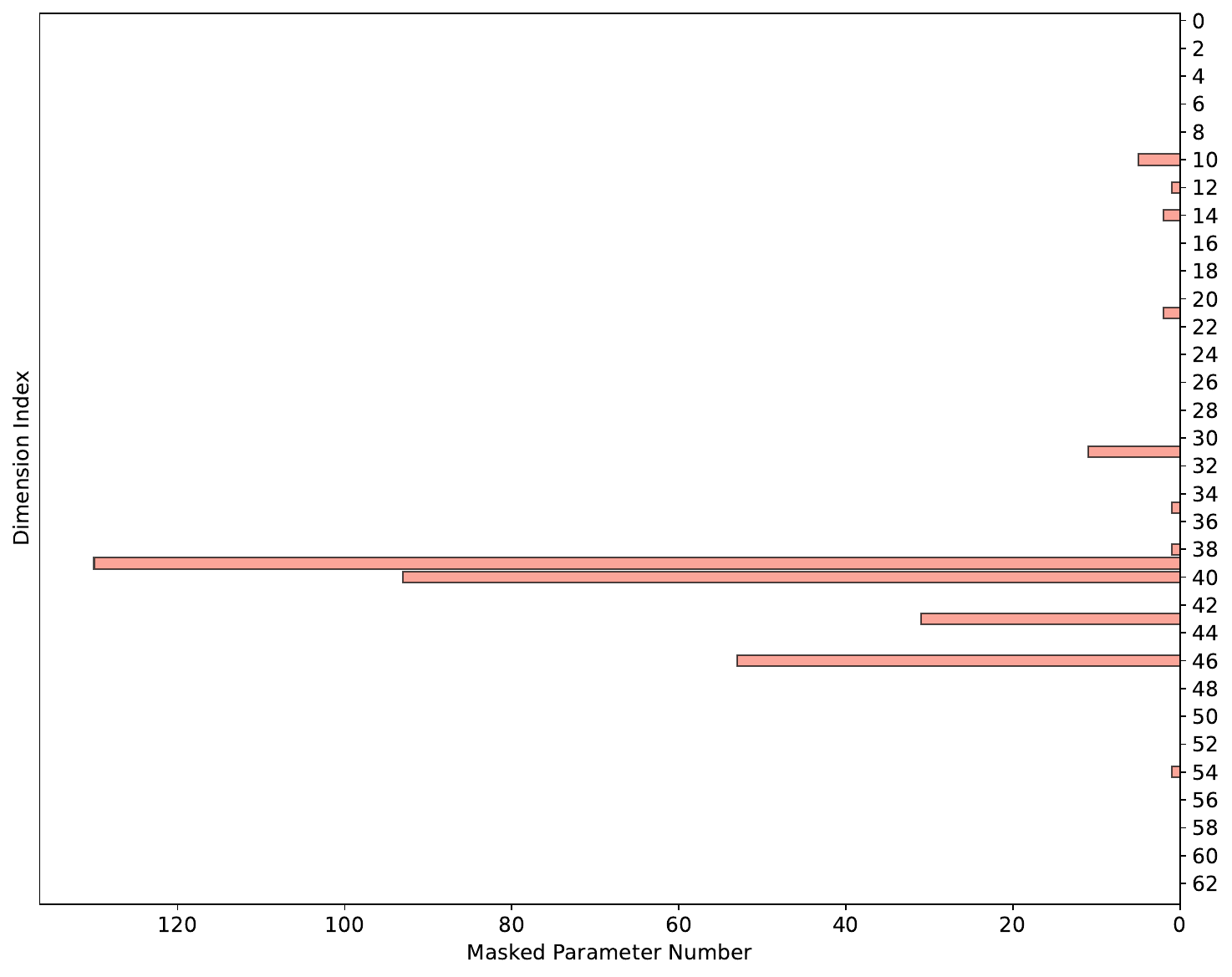}}
    \hfill
    \subfloat[Activation map]{%
        \includegraphics[width=0.5\textwidth, height=6.08cm]{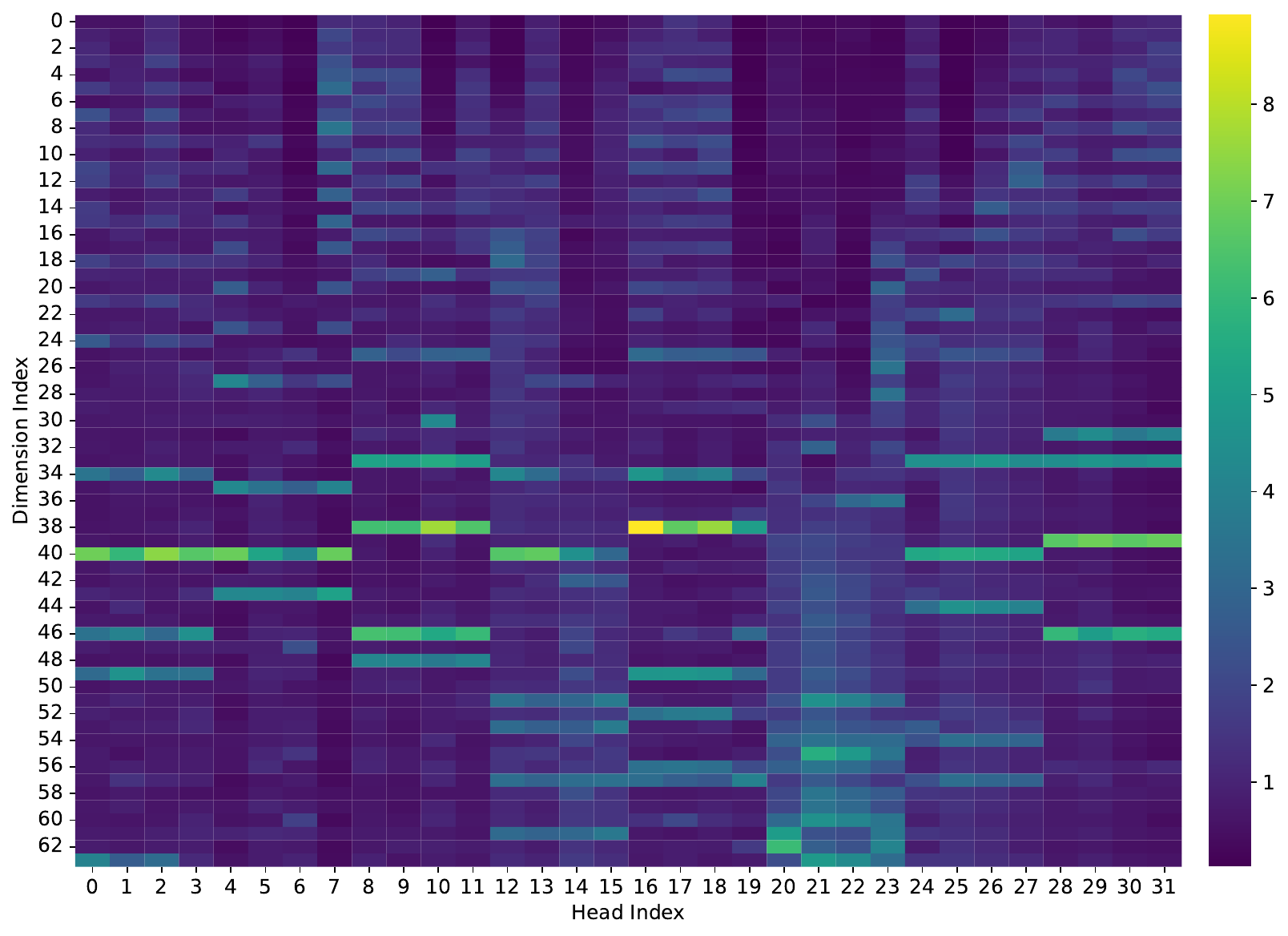}}
    
    \caption{ToM-sensitive parameter distribution and activation map for \(W_\mathbf{Q}\) matrix in Llama3-8B layer 2.}
    \label{fig:mask_and_activation_llama}
\end{figure}

Here, we visualize the distribution of ToM-sensitive parameters across different frequency positions within the same layer, along with the corresponding activation map. All results are averaged over attention heads. We make the following observations:

\paragraph{ToM-sensitive parameter patterns impair dominant-frequency activations in RoPE-based models.}
As shown in Figure \ref{fig:mask_and_activation_llama}, we first observe the presence of dominant-frequency activations in LLaMA3-8B, which is a common characteristic across RoPE-based models. Furthermore, we find that the sensitive parameters are concentrated precisely around these dominant frequencies. For instance, in LLaMA3-8B layer 2, the dominant frequency appears around the range of 39–40, and the ToM-sensitive parameters are densely distributed in this region.

\begin{figure}[h]
    \centering
    \subfloat[ToM-sensitive parameter distribution]{%
\includegraphics[width=0.46\textwidth, height=5.95cm]{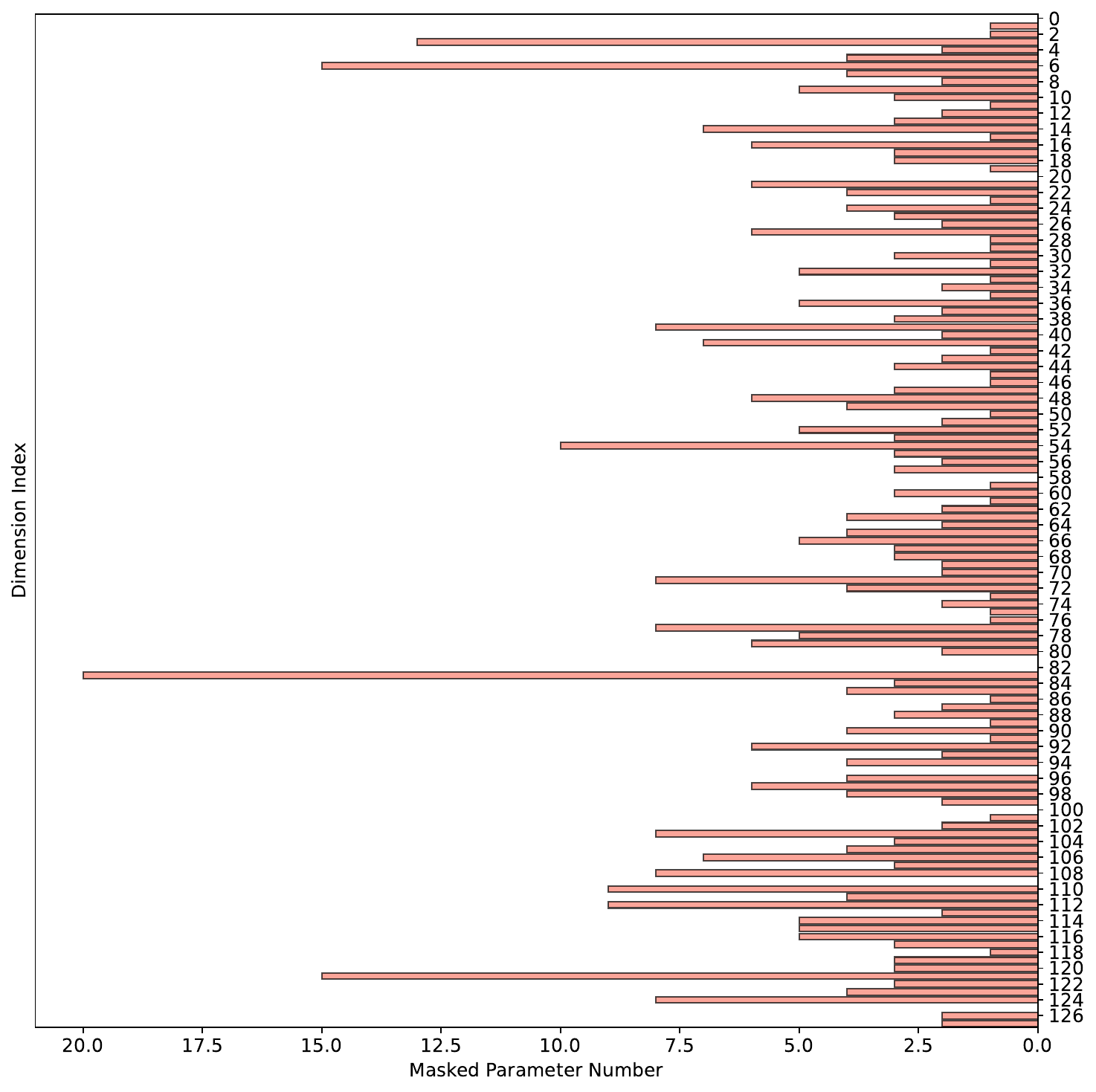}}
    \hfill
    \subfloat[Activation map]{%
        \includegraphics[width=0.5\textwidth]{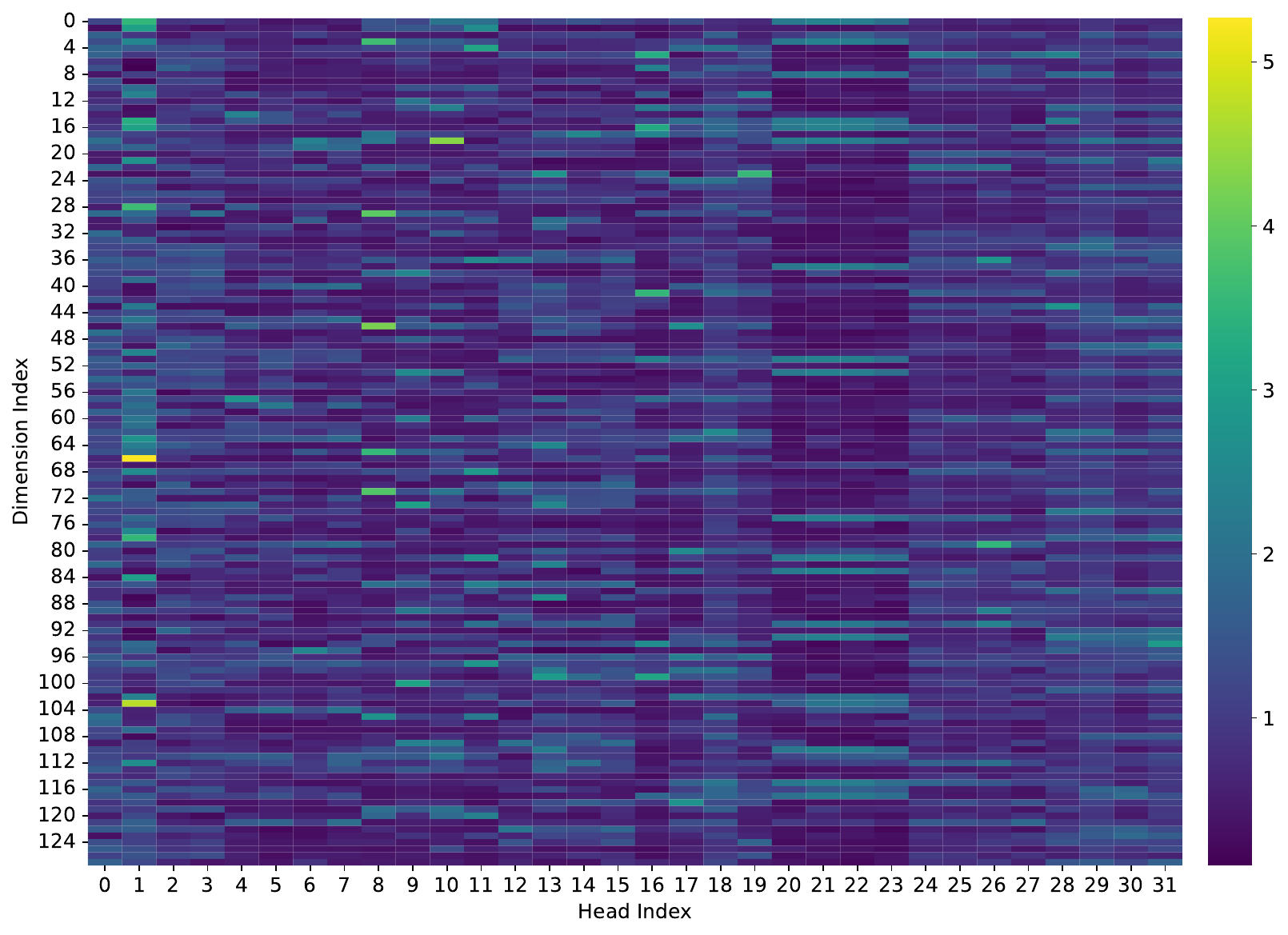}}
    
    \caption{ToM-sensitive parameter distribution and activation map for \(W_\mathbf{Q}\) matrix in Jamba-1.5-Mini layer 4.}
    \label{fig:mask_and_activation_jamba}
\end{figure}

\paragraph{ToM-sensitive parameter patterns do not affect Non-RoPE-based models.}
As shown in Figure \ref{fig:mask_and_activation_jamba}, we observe that the activation map of the Jamba model exhibits no dominant-frequency activation. Moreover, there is no apparent correlation between the parameter distribution and the activation map. This suggests that the mechanism underlying ToM reasoning in Non-RoPE-based models may fundamentally differ from that in RoPE-based models.

\subsection{Perturbing activations affects attention mechanism}
\paragraph{Perturbing Dominant-Frequency Activations.}
Let \(\mathbf{Q}, \mathbf{K}\) be the query and key activation matrices in a single attention head, and let \(\mathbf{Q}_{f_1}, \mathbf{K}_{f_2}\) denote the dominant-frequency components in \(\mathbf{Q}\) and \(\mathbf{K}\). The attention score matrix \(\mathbf{A}\) is given by:
\begin{align*}
\mathbf{A} &= \mathbf{Q}\,\mathbf{K}^\top.
\end{align*}
If we add small perturbations \(\Delta \mathbf{Q}_{f_1}\) and \(\Delta \mathbf{K}_{f_2}\) to these dominant-frequency parts, we can write the updated attention score matrix \(\mathbf{A}'\) as:
\begin{align*}
\mathbf{A}' &= (\mathbf{Q} + \Delta \mathbf{Q}_{f_1})(\mathbf{K} + \Delta \mathbf{K}_{f_2})^\top \\
            &= \mathbf{Q}\,\mathbf{K}^\top 
               \;+\; \mathbf{Q}\,\Delta \mathbf{K}_{f_2}^\top
               \;+\; \Delta \mathbf{Q}_{f_1}\,\mathbf{K}^\top
               \;+\; \Delta \mathbf{Q}_{f_1}\,\Delta \mathbf{K}_{f_2}^\top. \label{eq:att_score_perturb}
\end{align*}
Subtracting the original score \(\mathbf{A}\) from \(\mathbf{A}'\) yields:
\begin{align*}
\Delta \mathbf{A} \;=\; \mathbf{A}' - \mathbf{A} 
             &= \underbrace{\mathbf{Q}\,\Delta \mathbf{K}_{f_2}^\top}_{\text{term 1}}
              \;+\;\underbrace{\Delta \mathbf{Q}_{f_1}\,\mathbf{K}^\top}_{\text{term 2}}
              \;+\;\underbrace{\Delta \mathbf{Q}_{f_1}\,\Delta \mathbf{K}_{f_2}^\top}_{\text{term 3}}. 
\end{align*}

Terms 1 and 2 describe how perturbations in dominant-frequency components modulate the original query and key activations, while term 3 represents second-order effects. We often observe that \( f_1 \approx f_2 \), meaning the dominant frequencies in the original query and key activations are selectively involved in the perturbation. As a result, the perturbations in attention scores tend to be large. Since these dominant-frequency activations are crucial for attention computation, their disruption can distort attention distributions, ultimately impairing the model’s ability to encode accurate attention relationships.

\section{Related Works}\label{sec:related}

\paragraph{ToM in LLMs.}
The emergence of ToM capabilities in LLMs has been a subject of significant debate. Recent studies by \citep{kosinski_2024_evaluating} and \citep{wastrachan_2024_testing} suggest that LLMs exhibit emergent ToM abilities, demonstrating an understanding of false beliefs, intentions, and mental states. However, \citep{ullman_2023_large} argues that these abilities may not be genuine, as models often fail to correctly answer ToM questions when even minor changes are introduced. This controversy has spurred extensive research into the development of comprehensive, fair, and more complex ToM benchmarks \citep{soubki_2024_views, street_2024_llms, xu_2024_opentom}. For instance, \citep{wu_2023_hitom} has introduced Hi-ToM, a benchmark designed to test models' ability to infer higher-order mental states. Additionally, beyond evaluating the ToM capabilities of individual models, researchers have also investigated the role of ToM in multi-agent interactions \citep{li_2023_theory}. Furthermore, researchers have explored the use of ToM-related questions as prompts to elicit deeper reasoning from models \citep{wagner_2024_mind, wilf_2024_think}. 

\paragraph{Localizing Important Neurons in Networks.}
The problem of localizing important neurons in networks has garnered significant attention since the inception of neural networks. A common approach involves identifying critical neurons based on gradient information. For instance, \citep{lee_2018_snip} utilized first-order gradient information to pinpoint influential neurons, while \cite{lecun_1989_optimal} assumed that first-order gradients tend to vanish as the model converges and instead employed second-order gradient information for this purpose. This technique has been widely applied in various domains, including network pruning \cite{sun_2023_a}, quantization \citep{kim_2023_squeezellm}, and AI safety \citep{wei_2024_assessing}.

\end{document}